\documentclass{article} % For LaTeX2e
\usepackage{iclr2026_conference,times}

\usepackage[utf8]{inputenc} % allow utf-8 input
\usepackage[T1]{fontenc}    % use 8-bit T1 fonts
\usepackage{hyperref}       % hyperlinks
\usepackage{url}            % simple URL typesetting
\usepackage{booktabs}       % professional-quality tables
\usepackage{amsfonts}       % blackboard math symbols
\usepackage{nicefrac}       % compact symbols for 1/2, etc.
\usepackage{microtype}      % microtypography
\usepackage{xcolor}         % colors

\usepackage{multicol}
\usepackage{multirow}
\usepackage{tabularx}
\usepackage{enumitem}
\usepackage{longtable}
\usepackage{graphicx}
\usepackage{array}
\usepackage{amssymb}
\usepackage{multirow}
\usepackage{xcolor,colortbl}
\usepackage{xspace}
\usepackage{makecell}
\usepackage{bbding}
\usepackage{boldline}
\usepackage{pifont}
\usepackage{tabularx}
\usepackage{graphicx}
\usepackage{wrapfig}
\usepackage[export]{adjustbox}
\usepackage{fvextra}
\usepackage[normalem]{ulem}
\usepackage[most,breakable]{tcolorbox}
\usepackage[hypcap=false]{caption}
\usepackage{pgfplots}
\usepackage{arydshln}
\usepackage{subcaption}
\pgfplotsset{compat=newest}
\usepgfplotslibrary{units}
\usepackage{siunitx}

\usepackage{amsmath,amsfonts,bm}

\def\eqref#1{equation~\ref{#1}}
\def\1{\bm{1}}

\DeclareMathAlphabet{\mathsfit}{\encodingdefault}{\sfdefault}{m}{sl}
\SetMathAlphabet{\mathsfit}{bold}{\encodingdefault}{\sfdefault}{bx}{n}

\usepackage{xspace}
\usepackage{fontawesome}

\definecolor{darkblue}{rgb}{0, 0, 0.5}
\definecolor{veronica-red}{RGB}{196,30,58}
\hypersetup{colorlinks=true, citecolor=darkblue, linkcolor=darkblue, urlcolor=darkblue}
\definecolor{ForestGreen}{RGB}{34,139,34}
\definecolor{BrickRed}{rgb}{.72,0,0}
\definecolor{LakeBlue}{RGB}{0,61,153}
\definecolor{boxgreen}{rgb}{0.9,1.0,0.9}
\definecolor{boxblue}{rgb}{0.9,0.9,1.0}
\definecolor{boxred}{rgb}{1.0, 0.85, 0.95}

\newcommand{\cmark}{\color{ForestGreen}{\Checkmark}}
\newcommand{\xmark}{\color{BrickRed}{\XSolidBrush}}

\newcommand{\fstar}{\textsuperscript{\fontsize{6pt}{6pt}\selectfont \faStarO}}
\newcommand{\fmoon}{\textsuperscript{\fontsize{6pt}{6pt}\selectfont \faMoonO}}
\newcommand{\fleaf}{\textsuperscript{\fontsize{6pt}{6pt}\selectfont \faLeaf}}

\newcommand{\ours}{\textsc{ScienceBoard}\xspace}
\newcommand{\cua}{computer-using agent\xspace}
\newcommand{\cuas}{computer-using agents\xspace}

\newcommand{\numtasks}[0]{169\xspace}

\newcommand{\atree}{\texttt{a11ytree}\xspace}
\newcommand{\gpt}{\texttt{GPT-4o}\xspace}
\newcommand{\gptfive}{\texttt{GPT-5}\xspace}
\newcommand{\tembd}{\texttt{text-embedding-3-small}\xspace}
\newcommand{\othree}{\texttt{o3-mini}\xspace}
\newcommand{\claude}{\texttt{Claude-3.7-Sonnet}\xspace}

\newcommand{\gemini}{\texttt{Gemini-2.0-Flash}\xspace}
\newcommand{\geminitwofivepro}{\texttt{Gemini-2.5-Pro}\xspace}
\newcommand{\atlas}{OS-Atlas-Pro-7B\xspace}

\newcounter{instruction}

\title{\vspace{-0.75em}\raisebox{0.075em}{\includegraphics[scale=.05, valign=c]{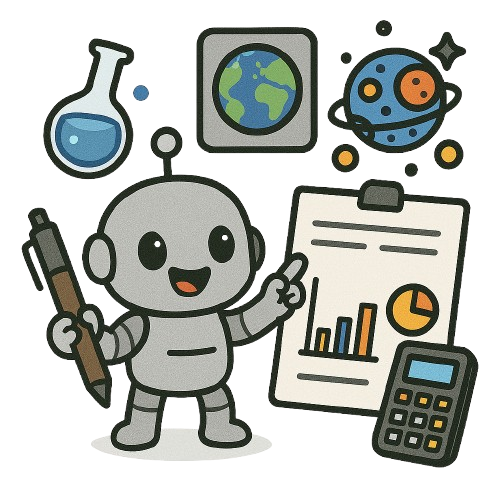}} ScienceBoard: \\ Evaluating Multimodal Autonomous Agents in Realistic Scientific Workflows}
\author{\textbf{Qiushi Sun}$^{\heartsuit}$  
        \textbf{Zhoumianze Liu}$^{\diamondsuit}$\fmoon\
        \textbf{Chang Ma}$^\heartsuit$
        \textbf{Zichen Ding}$^\diamondsuit$ 
        \textbf{Fangzhi Xu}$^\diamondsuit$ 
        \textbf{Zhangyue Yin}\fmoon\  \\
        \textbf{Haiteng Zhao}\fstar\  
        \textbf{Zhenyu Wu}$^\diamondsuit$ 
        \textbf{Kanzhi Cheng}$^\clubsuit$ 
        \textbf{Zhaoyang Liu}$^\diamondsuit$ 
        \textbf{Qintong Li}$^\heartsuit$ \\
        \textbf{Jianing Wang}$^\spadesuit$ 
        \textbf{Xiangru Tang}\fleaf\
        \textbf{Tianbao Xie}$^\heartsuit$ 
         \textbf{Xiachong Feng}$^\heartsuit$ 
        \textbf{Xiang Li}$^\spadesuit$ \\
        \textbf{Ben Kao}$^\heartsuit$ 
        \textbf{Wenhai Wang}$^\diamondsuit$ 
        \textbf{Biqing Qi}$^\diamondsuit$ 
        \textbf{Lingpeng Kong}$^\heartsuit$
        \textbf{Zhiyong Wu}$^\diamondsuit$ 
       \\ 
  $^\heartsuit$The University of Hong Kong 
  $^\diamondsuit$Shanghai AI Laboratory
    \fmoon Fudan University \\
  \fstar Peking University 
  $^\clubsuit$Nanjing University 
  $^\spadesuit$East China Normal University 
  \fleaf Yale University\\
  \texttt{\{qiushisun,changma\}@connect.hku.hk}, \texttt{\{kao,lpk\}@cs.hku.hk} \\ \texttt{\{liuzhoumianze,wangwenhai,qibiqing,wuzhiyong\}@pjlab.org.cn}
}

\iclrfinalcopy % Uncomment for camera-ready version, but NOT for submission.
\begin{document}

\maketitle

\vspace{-1em}
\begin{abstract}
Large Language Models (LLMs) have extended their impact beyond Natural Language Processing, 
substantially fostering the development of interdisciplinary research.
Recently, 
various LLM-based agents have been developed to assist scientific discovery progress across multiple aspects and domains.
Among these, 
computer-using agents, capable of interacting with operating systems as humans do,
are paving the way to automated scientific problem-solving and addressing routines in researchers’ workflows.
Recognizing the transformative potential of these agents,
we introduce \ours, which encompasses two complementary contributions: 
(i) a realistic, multi-domain environment featuring dynamic and visually rich scientific workflows with integrated professional software, where agents can autonomously interact via different interfaces to accelerate complex research tasks and experiments;
% (i) a realistic, multi-domain environment providing authentic scientific discovery workflows with integrated professional software, where agents can autonomously interact via different interfaces to accelerate complex research tasks and experiments;
and 
(ii) a challenging benchmark of \numtasks high‑quality, rigorously validated real‑world tasks curated by humans, spanning scientific‑discovery workflows in domains such as biochemistry, astronomy, and geoinformatics.
% Extensive evaluations demonstrate that, despite some promising results, current agents still fall short of reliably assisting scientists with complex workflows (15\% success rate).
Extensive evaluations of agents with state-of-the-art backbones (\textit{e.g.}, GPT-5, Gemini-2.5-Po, UI-TARS) show that, despite some promising results, they still fall short of reliably assisting scientists in complex workflows, achieving only a 20\% overall success rate.
In‑depth analysis further provides valuable insights for addressing current agent limitations and more effective design principles, paving the way to build more capable agents for scientific discovery.
Our code, benchmark, and leaderboard are available at \href{https://qiushisun.github.io/ScienceBoard-Home/}{Scienceboard Homepage}.
\end{abstract} 

\section{Introduction}

% \sqs{todo: @chang, any suggestions?}

% 1. Worried about outdated baselines, perhaps run claude 4 sonnet and latest UI tars?

% 2. I still feel the link between first 3 paragraphs are weak. Experimental observations -> need autonomous science agents -> use science softwares, perhaps merge 2 and 3 for more direct intro to the topic.\\ 

% \chang{3.You are mentioning discovery multiple times, but it's more about copilot than automatic science discovery, I think maybe give 1/2 examples how success on that problem is actually making some discovery?}

% 4.Conclusion paragraph in the intro is very weak, any insights on why these LLM fail as coscientists? fixed

% 5.Some of the appendix results are not quoted in main paper. Just give a very high level list would be ok, e.g. you could find difficulty, error analysis in xxx, otherwise reviewers will ignore them.

% \chang{remove these adjs, change to expertise}
In the pursuit of scientific advances, researchers combine ingenuity and expertise to perform novel research grounded in experimental explorations. In the modern era, scientific discovery is increasingly driven by specialized software and tools that empower scientists to engage deeply with the experimental world~\citep{hacking1983representing}. Tools like simulation engines~\citep{hollingsworth2018molecular}, data analysis software~\citep{matworks}, and visualization platforms~\citep{goddard2018chimerax} are essential for formulating hypotheses, validating results, and advancing scientific understanding. 
% \chang{For this paragraph, change tools to software, emphasize appropriately using these tools are essential for formulating ....}

% \chang{Given the increasing number of scientific tools and the rising demand for streamlined scientific workflows, there is a growing expectation that autonomous agents will progressively automate research pipelines, evolving into AI-powered research assistants or co-scientists. These agents should be capable of proficiently operating scientific tools in alignment with research objectives, and exhibit human-like planning and reasoning. Ultimately, this would enable fully autonomous workflows—from utilizing scientific tools to generating new scientific discoveries.}

\begin{figure}[th]
    \centering
        \includegraphics[width=0.925\linewidth]{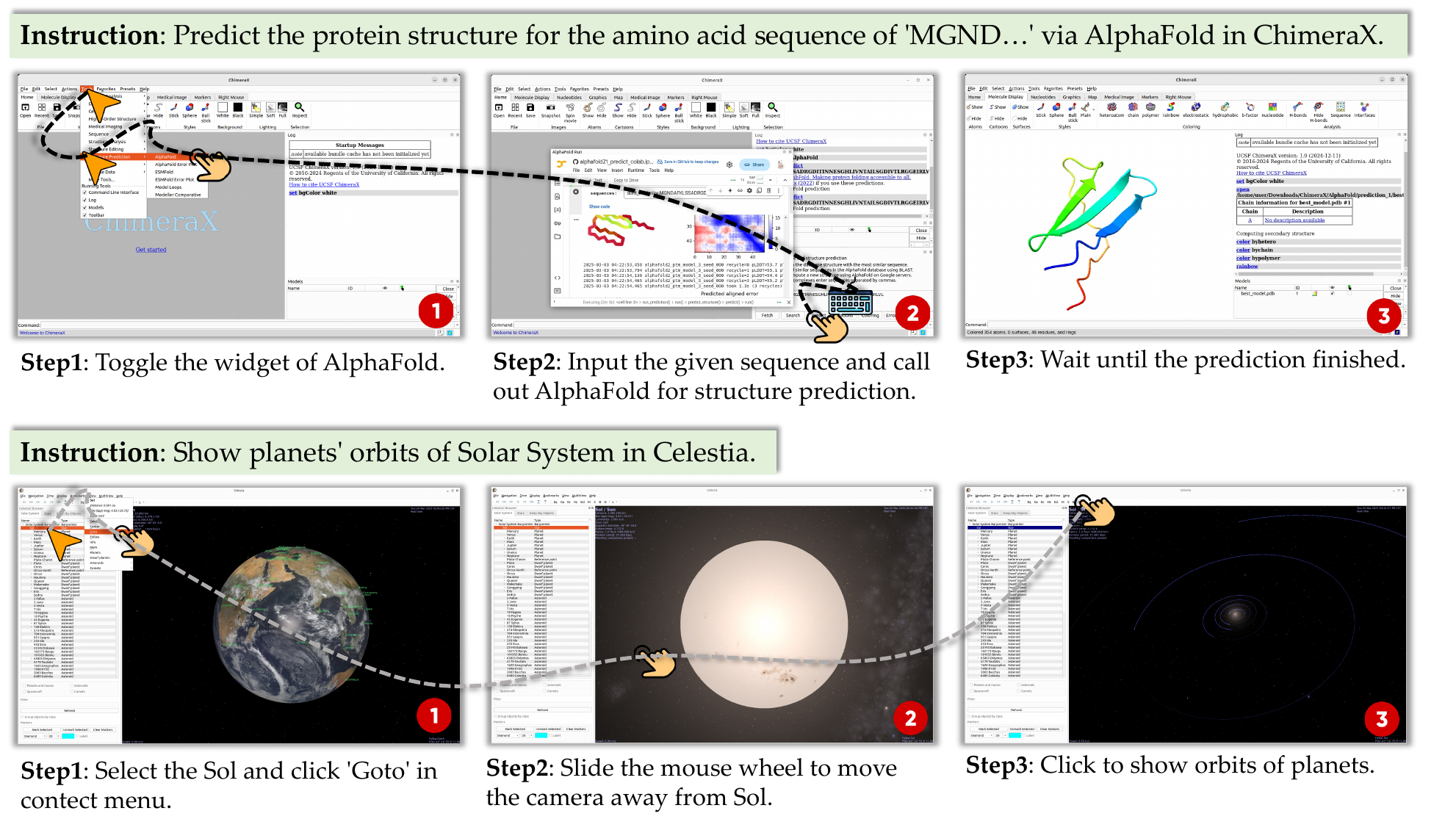}
        % \vspace{-0.5em}
        \caption{
    \ours is a pioneering computer environment for scientific discovery agents, integrated with professional software. It enables agents to autonomously follow instructions and complete realistic scientific tasks by interacting with the system via GUI or CLI.
    }
    \label{fig:task_demonstrate}
    \vspace{-1em}
\end{figure}

However, as scientific software grows more sophisticated and workflows become more demanding, the learning curve and operational burden on human researchers intensify~\citep{S_nger_2024}.
These challenges motivate the vision of autonomous agents to play a central role in automating research pipelines and assisting human researchers as {``AI co-scientists''}~\citep{luo2025llm4sr,schmidgall2025agentlab,gottweis2025aicoscientist}. For example, while a human scientist may take weeks to master a protein analysis tool~\citep{meng2023chimerax}
% like ChimeraX\citep{meng2023chimerax} 
and spend hours making sufficient observations, an autonomous agent could perform the same tasks within minutes. By enabling fully autonomous workflows—from tool usage to making novel discoveries~\citep{lu2024ai}—such agents promise to accelerate science and empower researchers with unprecedented capabilities.

% there is a rising expectation that autonomous agents will 

% With the increasing complexity of scientific tools and the growing demand for more streamlined scientific workflows, there is a rising expectation that autonomous agents will play a central role in automating research pipelines and assisting human researchers as {``AI co-scientists''}~\citep{luo2025llm4sr,schmidgall2025agentlab,gottweis2025aicoscientist}. For example, while a human scientist may take weeks to master a protein analysis tool~\citep{meng2023chimerax}
% like ChimeraX\citep{meng2023chimerax} 
% and spend hours making sufficient observations, an autonomous agent could perform the same tasks within minutes. By enabling fully autonomous workflows—from tool usage to making novel discoveries~\citep{lu2024ai}—such agents promise to accelerate science and empower researchers with unprecedented capabilities.
% \chang{You need to start mentioning softwares in this paragraph, the general logic is ok, but the tools are a bit vague and not sure why you need computer using.}

% \ours is a pioneering computer environment for scientific discovery agents, integrated with professional software. It enables agents to autonomously follow instructions and complete realistic scientific tasks by interacting with the system via GUI or CLI.

% Recently emerging computer-using agents (CUAs)\citep{wu2024oscopilot,openai2025cua} offer a concrete step toward realizing this vision.

Recently emerging computer-using agents \citep{wu2024oscopilot,openai2025cua}, capable of operating digital devices in a human-like manner, present a promising approach toward achieving these visions. These agents can interact with operating systems through Command-Line Interfaces (CLI; \citealp{sun2024survey,wang2024what}) or perform mouse and keyboard actions 
via Graphical User Interfaces~(GUI; \citealp{cheng2024seeclick,wu2025osatlas}), mimicking the user experience to flexibly automate complex workflows~\citep{OSWorld,rawles2024androidworld,hu2024dawn}.
% . By closely mimicking the user experience when interacting with tools~\citep{OSWorld,rawles2024androidworld,hu2024dawn}, 
% these agents enable a unified paradigm where software can be leveraged to automate complex scientific workflows with maximum flexibility.
As illustrated in Figure~\ref{fig:task_demonstrate}, to predict the protein structure of an amino acid sequence, the agent launches ChimeraX, selects the AlphaFold widget, and inputs the sequence for prediction. In this way,
scientific tasks could be performed through
step-by-step 
autonomous interaction with software.

% Recently emerging computer-using agents (CUAs)\citep{wu2024oscopilot,openai2025cua} offer a concrete step toward realizing this vision. Unlike conventional AI systems limited to text-based reasoning, CUAs interact with digital devices in a human-like manner. They can operate operating systems via Command-Line Interfaces (CLI; \citealp{sun2024survey,wang2024what}) or through mouse and keyboard actions on Graphical User Interfaces (GUI; \citealp{cheng2024seeclick,wu2025osatlas}), thereby closely mimicking human interaction to automate complex workflows\citep{OSWorld,rawles2024androidworld,hu2024dawn}.

% Recently emerging computer-using agents~\citep{wu2024oscopilot,openai2025cua} offer a promising paradigm for scientific automation by operating digital devices in a human-like manner. These agents can interact with software via Command-Line Interfaces (CLI; \citealp{sun2024survey,wang2024what}) or Graphical User Interfaces (GUI; \citealp{cheng2024seeclick,wu2025osatlas}), mimicking the user experience to flexibly automate complex workflows~\citep{OSWorld,rawles2024androidworld,hu2024dawn}. For example, to predict a protein's structure, an agent can autonomously launch ChimeraX, operate the AlphaFold widget, and input a sequence, thus completing the task through step-by-step software interaction (Figure~\ref{fig:task_demonstrate}).

% \vspace{0.5em}

To initiate the use of \cuas to assist human scientists with daily tasks,
we introduce \ours,
a novel realistic environment designed for developing AI-powered research assistants.
Our infrastructure comprises a scalable framework for scientific exploration that integrates: 
(1) a flexible ecosystem comprising scientific software across multiple domains,
and 
(2) standardized evaluation pipelines for rigorous assessment.
It supports dual-mode interaction, 
allowing LLM/VLM-based computer agents to operate through either CLI or GUI.

Building upon \ours, we curate a benchmark comprising \numtasks tasks that encompass scientific experiment workflows drawn from six scientific domains, including algebra, biochemistry, theorem proving, geographic information systems, astronomy, and scientific documentation.
% as exemplified in Figure~\ref{fig:task_demonstrate}
% \chang{cite figure 1 twice, maybe just list it out} \sqs{ok}. 
These high-quality and challenging tasks are meticulously designed by annotators with disciplinary backgrounds, simulating the daily routines faced by human scientists. 
% Agents are required to complete these tasks through interactions with the system via CLI and GUI actions, leveraging visual or structured information (or both).
Task completion requires agents to interact with the system via CLI and GUI, exercising a wide range of capabilities—including visual and textual reasoning, tool manipulation, coding, mathematics, spatial understanding, and deep domain-specific knowledge.
Unlike widely used desktop applications, scientific software exhibits considerable complexity in I/O formats. Consequently, we reconfigure all software involved to ensure the accuracy and reliability of execution-based evaluation. We design a suite of evaluation functions that verify task completion by retrieving the internal states of the system.

We evaluate widely used LLMs and VLMs as agents on \ours, incorporating both proprietary models and their open-source counterparts. 
Across different observation settings, the average success rate of agents ranges between 0\% to 15\%, 
with performance peaking at 20\% in the most favorable subcategories. 
This demonstrates that current \cuas, while promising, remain far from capable of serving as scientific assistants,
largely due to their limited action capability and domain knowledge.
% and brittle performance in complex multi-step workflows.
% \chang{why? add more insights?}
Our analysis further reveals their inherent limitations and explores design principles for developing more agents for science.

% This demonstrates that current \cuas, while promising, remain far from capable of serving as scientific assistants, largely due to their weak grounding ability, limited domain knowledge, and brittle performance in complex multi-step workflows. Our analysis further reveals these inherent limitations and explores design principles for developing more capable agents for science.

\section{Related Works}

% The interactive capabilities of language agents~\citep{sumers2024cognitive,liu2024agentbench} have led to the rapid development of Computer-Using Agents (CUAs) that automate digital tasks like a human~\citep{openai2025cua}. Research in this domain follows two primary approaches. The first is through the Command-Line Interface (CLI), where agents programmatically interact with systems by generating executable scripts or invoking APIs~\citep{wang2024codeact,wu2024oscopilot}. The second is through the Graphical User Interface (GUI), where VLM-powered agents translate instructions into human-like mouse and keyboard actions~\citep{cheng2024seeclick,wu2025osatlas,niu2024screenagent}. This GUI-based approach has already shown promise in automating desktop, mobile, and engineering workflows~\citep{OSWorld,rawles2024androidworld,cao2024spiderv}. Building on these advances, our work pioneers the application of these agents to scientific workflows, advancing the vision of an autonomous research assistant.

\paragraph{Computer-Using Agents.}
Language agents~\citep{sumers2024cognitive}
have recently garnered significant attention due to their interactive capabilities~\citep{li2023camel,sun2023corex,hong2024metagpt,liu2024agentbench}.
Recent studies indicate their potential to interact with operating systems and automate computer tasks as humans do, 
leading to the proliferation of \cuas~\citep{openai2025cua}.
One line of research utilizes Command Line Interface (CLI), 
where agents generate executable scripts (\textit{e.g.}, Python or Shell scripts) to interact with systems programmatically~\citep{wang2024codeact}. 
In this process, agents perform code synthesis~\citep{sun2024survey} or invoke APIs~\citep{wu2024oscopilot,zhang2024ufo}.
% to manipulate computers.  % according to user instructions.
Another line of research focuses on Graphical User Interface (GUI) agents~\citep{cheng2024seeclick,wu2025osatlas,lin2024showui} that interact with digital devices through human-like mouse and keyboard actions~\citep{niu2024screenagent,zheng2024seeact,gou2025navigating}. 
These agents transform user instructions into executable actions within the operating system (\textit{e.g.}, clicking an icon or scrolling through a page).
Powered by VLMs, 
GUI agents have been applied to automate desktop~\citep{OSWorld} and mobile~\citep{rawles2024androidworld} tasks, 
as well as specialized engineering workflows~\citep{cao2024spiderv}, 
showing promising paths toward digital automation. 
This work innovatively initiates the use of computer agents in scientific workflows,
taking a step closer to autonomous research assistants.

\vspace{-0.85em}
\paragraph{AI for Scientific Discovery.}
The rapid advancement of LLMs has reshaped the landscape of scientific discovery~\citep{ai4science2023impact},
boosting multiple stages of the research cycle~\citep{luo2025llm4sr}.
% providing support across multiple stages of the research cycle~\citep{luo2025llm4sr}.
With the rise of LLM/VLM-based agents, 
there is a growing demand for these game-changers with college-level knowledge~\citep{wang2024scibench} to transcend traditional tasks like question answering~\citep{lu2022learn,krithara2023bioasq,lu2024moleculeqa}.
Recent efforts have been directed towards harnessing such power to assist with diverse components of the research cycle, including idea and hypothesis generation~\citep{si2024can,liu2024aigs}, data analysis~\citep{chen2025scienceagentbench,gu2024blade,majumder2024discoverybench},
scientific programming~\citep{tian2024scicode,novikov2025alphaevolve},
paper writing~\citep{wang2024autosurvey}, and peer-reviewing~\citep{yu2024sea}.
Meanwhile, 
incorporating domain knowledge or even constructing foundation models~\citep{xia2025naturelm} can endow these agents with the capability to solve domain-specific problems, 
such as theorem proving~\citep{song2025towards}, chemical reasoning~\citep{ouyang2024structchem,tang2025chemagent} and biological discovery~\citep{wang2025spatialagent, zhao2025biomaze, wang2025spatialagent, frey2025lab}.
With the vision of constructing autonomous research assistants~\citep{schmidgall2025agentlab}, our work represents the first to support agents in executing end-to-end scientific exploration workflows, thereby laying a cornerstone for advancing AI-powered scientific discovery.
\section{\ours Environment}

In this part, 
we introduce \ours environment, 
which encompasses real-world science software that could be manipulated through GUI and CLI interfaces. The interface is developed based on an Ubuntu virtual machine (VM),  serving as the underlying infrastructure. The dynamic and visually intensive environments distinguish \ours from all previous works that evaluate the scientific capabilities of models or agents.

\subsection{Preliminaries and Task Definition\label{sec: preliminaries}}
A \cua receives task instructions, selects actions to manipulate software, and receives feedback reflecting changes in the environment (tabletop). This interaction is modeled as a Partially Observable Markov Decision Process (POMDP), defined by the tuple $\langle g, \mathcal{S}, \mathcal{A}, \mathcal{O}, \mathcal{T}\rangle$, where $g$ is the goal, $\mathcal{S}$ is the state space, $\mathcal{A}$ is the action space, $\mathcal{O}$ is the observation space (including environment feedback), and $\mathcal{T}: \mathcal{S} \times \mathcal{A} \rightarrow \mathcal{S}$ is the state transition function. Given a policy $\pi$, the agent predicts actions at each time step $t$ based on the goal $g$ and memory $m_t = {o_j, a_j, o_{j+1}, a_{j+1}, \dots, o_t}$ ($0 \leq j < t$), which records the sequence of past actions and observations. The trajectory $\tau = [s_0, a_0, s_1, a_1, \dots, s_t]$ is determined by the policy and environment dynamics:

\vspace{-7.5pt}
\begin{equation}
p_{\pi}(\tau) = p(s_0)\prod_{t=0}^T\pi(a_t|g, s_t, m_t)\mathcal{T}(s_{t+1}|s_t, a_t)
\end{equation}
\vspace{-18pt}

\begin{figure}[t]
    \centering
    \includegraphics[scale=0.475]{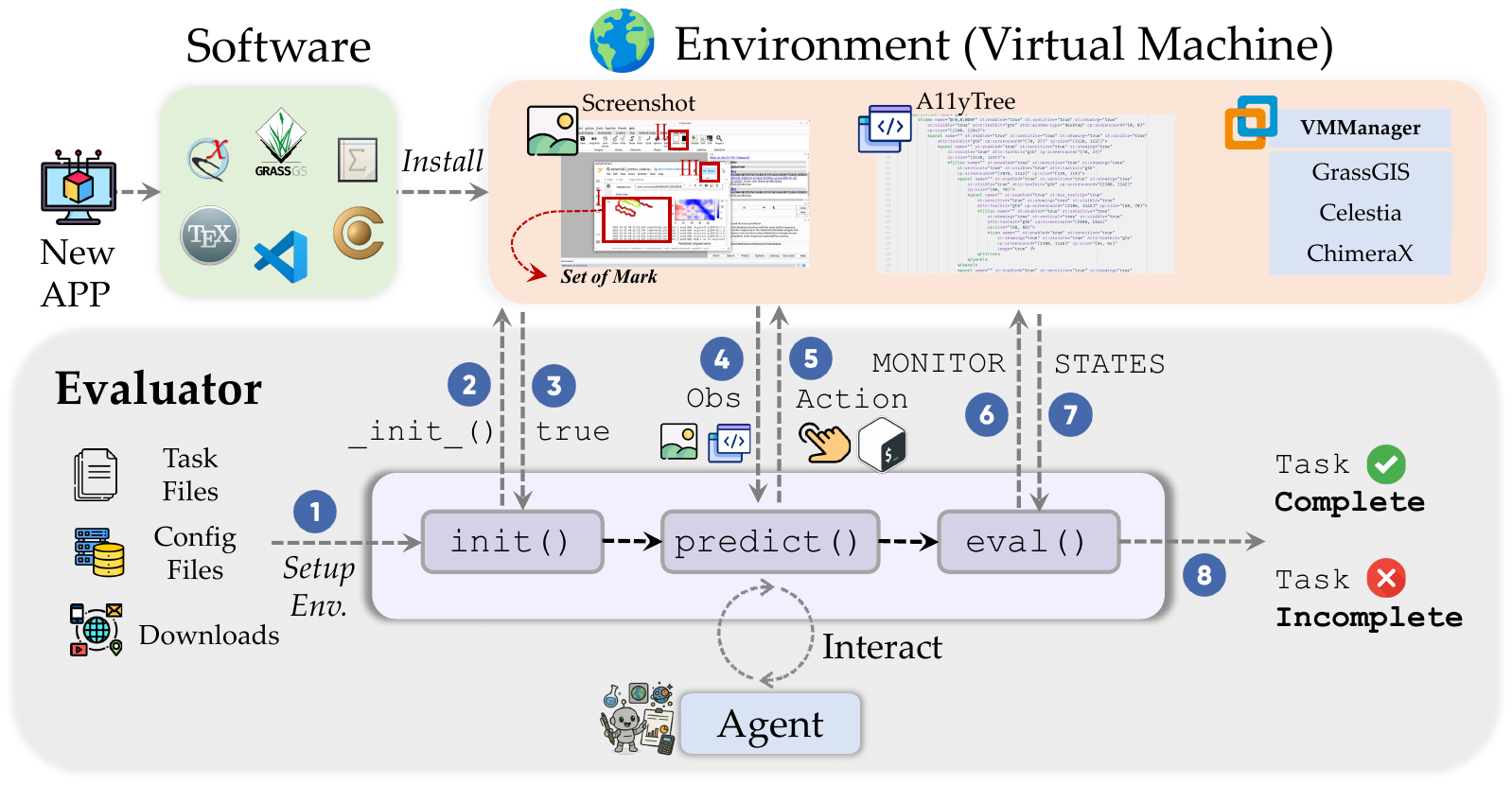}
    \caption{Overview of the \ours infrastructure. 
    The scalable environment is built upon a VM pre-installed with scientific discovery software. 
    It supports both CLI and GUI interfaces to enable autonomous agent interaction. 
    For each task designed to evaluate the agent’s capability as a research assistant, 
    an initialization script, configs, and related files are provided. 
    Agents perceive the environment through visual or textual modalities, and are expected to plan and act accordingly. 
    After the interaction, 
    an evaluation function determines completion based on the VM internal states.
    }
    \label{fig:scienceboard_sys}
    \vspace{-1em}
\end{figure}

\paragraph{Observation and Memory.}
We evaluate computer agents using three types of observation spaces: text-only, visual-only, and combined text-visual observations. For text-based observations, we use accessibility trees (\atree\footnote{\atree: Accessibility (a11y) trees are hierarchical structures representing UI elements on the screen.}) to generate structured textual representations of screenshots. For visual observations, we capture high-resolution screenshots directly. The specific observation combinations used in our experiments are detailed in Section~\ref{sec:exp_setup}, with further information in Appendix~\ref{app:obs_space}.
Our POMDP agent requires memory to retain historical information. Following previous work~\citep{yao2023react,ma2024agentboard}, we construct this memory by concatenating the agent’s most recent observations.

\vspace{-0.5em}
\paragraph{Goal and Unified Action Space.} 
Each task is specified by a natural language (NL) instruction, such as \texttt{Display atoms in sphere style}, describing the user’s intended goal. The policy model decomposes a complex goal instruction into a sequence of actions.
% The action space $\mathcal{A}$ in \ours includes both GUI and CLI actions.
We specially design a unified action space $\mathcal{A}$ in \ours,
integrating diverse interaction modalities crucial for scientific tasks.
For GUI actions, agents can perform the full range of human-computer interactions, including mouse movements, clicks, keystrokes, and other typical input behaviors as in prior work~\citep{OSWorld,zhou2024webarena}~(\textit{e.g.}, \texttt{CLICK[991, 019]}).
% For CLI actions, agents can invoke system-level commands in Ubuntu and commands within the scientific software, enabling them to write and execute code to interact with the environment.
For CLI actions, agents can interact at two levels: 
(a) invoking system-level commands within the Ubuntu terminal, and 
(b) utilizing application-specific CLI or scripting mechanisms.
Moreover, $\mathcal{A}$ comprises an \texttt{answer} action, enabling agents to provide specific answers for QA tasks, and a \texttt{call\_api} action, allowing agents to leverage predefined external APIs to broaden their capabilities.
A comprehensive list of supported action types is available in Appendix~\ref{app:action_space}.

\vspace{-0.5em}
\paragraph{LLM/VLM-based Policy Model.}
An LLM / VLM model acts as the policy model to drive the agent’s behavior. 
% At each step, 
The policy model receives the current observation and generates the next action accordingly. 
For pure-text observation, we adopt LLMs as the policy. 
Otherwise, we leverage VLMs. 
% For environments that involve visual inputs (e.g., GUI screenshots), we employ a VLM to process multimodal observations and determine the corresponding actions.

\subsection{Scientific Discovery Evaluation Framework}
\label{sec:env_details}

Unlike prior work that primarily focuses on static QA, coding, or single-step tasks,
we aim to provide agents with a realistic and visually grounded environment to support autonomous exploration, 
which in turn introduces greater challenges for planning and action.
In \ours environment, as shown in Figure~\ref{fig:scienceboard_sys}, 
we 
(1) simulate scenarios where scientific software is used to solve domain-specific problems, 
(2) enable agents to interact with the environment through diverse observations, 
and (3) ensure that agent behaviors can be rigorously evaluated.

% \vspace{-1em}
\paragraph{Scientific Software Installation and Adaptation.} For each domain, we select an open-source application that supports both visual and textual observations as the agent’s playground. 
To enable access to the internal state of each application within the VM, we adapt the software accordingly. 
Given the complexity and limited completeness of scientific applications, we inject a lightweight server that launches alongside the application’s main UI process to expose internal states via HTTP requests. 
This server is capable of querying the application’s runtime internal states, which serve as the basis for downstream evaluation.
For applications that do not natively support remote control via RESTful APIs, we modify and recompile their source code to ensure that both UI elements and internal states can be accessed.
In addition, the server supports partial state control of the software,
allowing us to initialize 
% the application 
with specific configurations to simulate contextualized task environments.
More about the software selected and further implementation details are provided in Appendix~\ref{app:benchmark_software}.
% zy{I understand that we spent lots of time solving those dirty engineering problems, but they'd better be put in the appendix rather than in the main bady. }

% For applications that do not natively support remote control via RESTful APIs, 
% 重新编译软件，在主进程外另加一个 http 服务器线程
% - 服务器可以获取 app 运行时 (runtime) 内部的状态，用于 eval 的根据
% - 部分服务器可以改变软件内部状态，用于初始化 (init) 从而为任务提供情境
% \chang{how do you modify and make sure that there is an UI}
% we modify and recompile their source code to ensure that both UI elements and internal states can be accessed.

\vspace{-1em}
\paragraph{Agent Interactions with the Environment.} The LLM/VLM agent interacts with the environment as described in Section~\ref{sec: preliminaries}, receiving observations and executing actions accordingly. Scientific software processes these actions and returns updated states. 
The agent operates autonomously, continuing this loop until it outputs a 
% termination 
signal (\verb|DONE| or \verb|FAIL|) or reaches the predefined attempt limit.

\begin{table}[htb]
\vspace{-5pt}
\caption{
Typical evaluation cases of \ours include exact matching, range-based assessment, and numerical tasks with tolerance. We have tailored appropriate evaluation methods for each task. Additional evaluation strategies are detailed in Appendix~\ref{app:benchmark_eval_cases}.
}
\resizebox{\textwidth}{!}{
\begin{tabular}{lp{.35\textwidth}l}
  \toprule
  \textbf{Initial State} &   \textbf{Instruction}  & \textbf{Evaluation Script (Simplified)}  \\
  \midrule
  \multirow{3}{*}{\raisebox{-3.15cm}{\includegraphics[width=5.5cm]{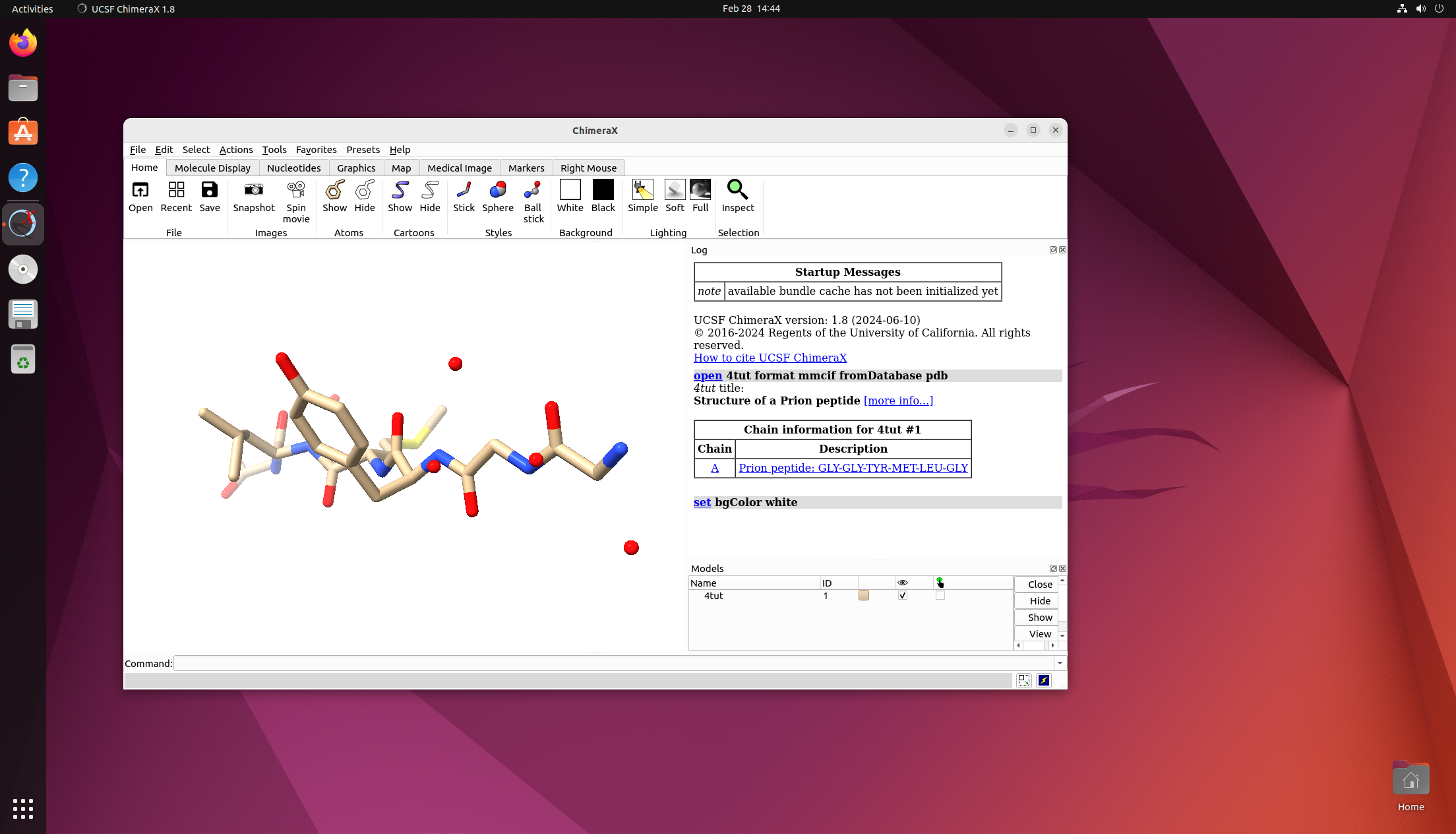}}}  & \multirow{3}{*}{\begin{minipage}{.35\textwidth}
  \textit{Select all water molecules and draw their centroids with radius of 1Å in ChimeraX.}
  \end{minipage}} 
&  \footnotesize\texttt{\ \ \{ }\\
% \footnotesize\texttt{[} \\
% &   & \footnotesize\texttt{\ \ \{ }\\
&   & \footnotesize\texttt{\ \ \ \ "type":"info","key":"sell",~}\\
% &   & \footnotesize\texttt{\ \ \ \ "key":"sel",~}\\
&   & \footnotesize\texttt{\ \ \ \ "value":["atom id \#!1/A:201@O idatm\_type O3"~}\\
% &   & \footnotesize\texttt{\ \ \ \ \ \ "atom id \#!1/A:201@O idatm\_type O3",~}\\
&   & \footnotesize\texttt{\ \ \ \ \ \ "...",]~}\\
% &   & \footnotesize\texttt{\ \ \ \ ]~}\\
&   & \footnotesize\texttt{\ \ \},\{~}\\
% &   & \footnotesize\texttt{\ \ \{ }\\
&   & \footnotesize\texttt{\ \ \ \ "type":"states",~}\\
&   & \footnotesize\texttt{\ \ \ \ "find":"lambda k,v:k.endswith('.\_name')",~}\\
&   & \footnotesize\texttt{\ \ \ \ "key":"lambda k:'...\_atoms\_drawing'",~}\\
&   & \footnotesize\texttt{\ \ \ \ "value":"[[13.0012 1.7766 21.3672 1.]]"~}\\
&   & \footnotesize\texttt{\ \ \} }\\
% &   & \footnotesize\texttt{]} \\
  \midrule
  \multirow{4}{*}{\raisebox{-3cm}{\includegraphics[width=5.5cm]{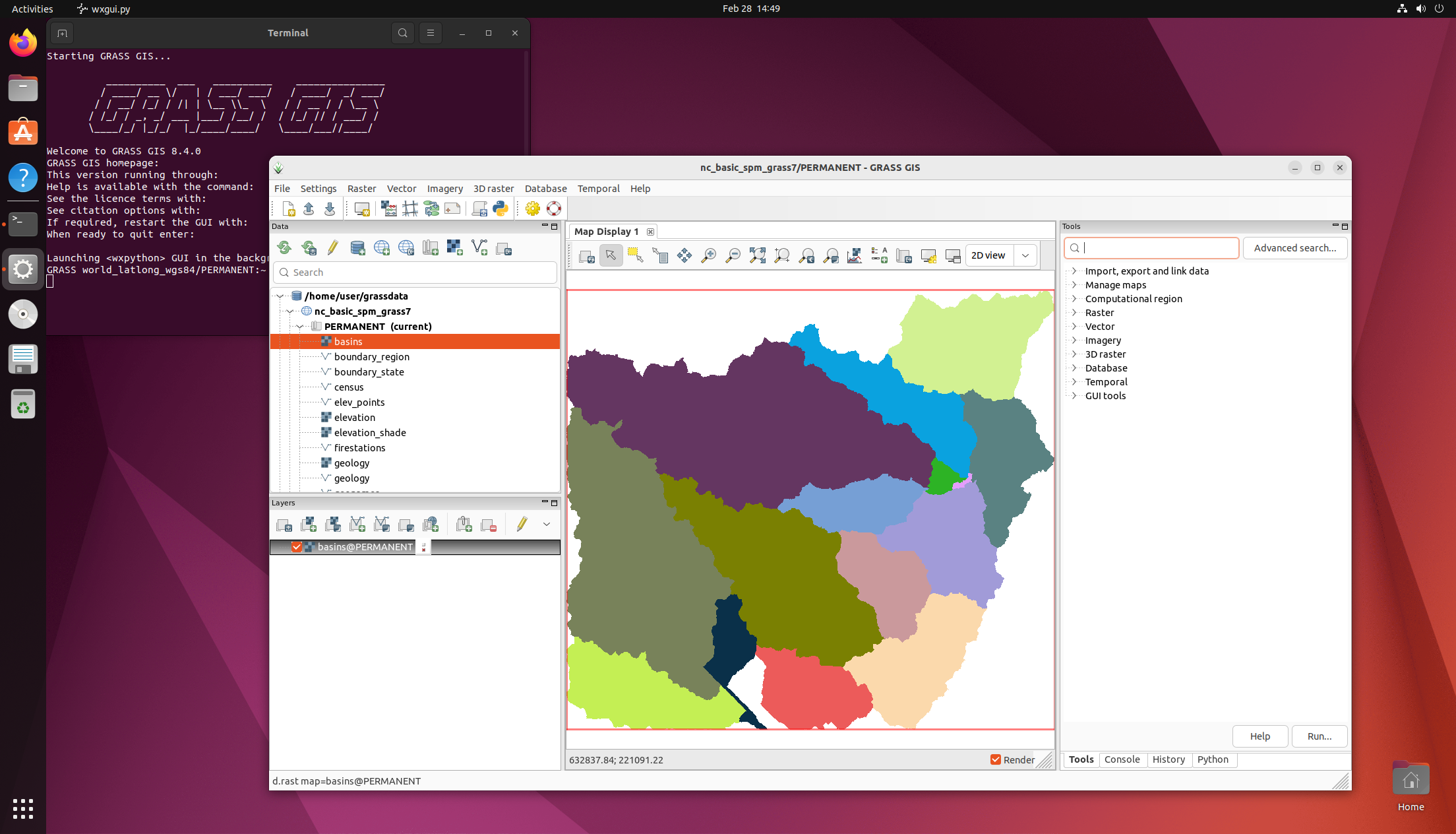}}} &  
   \multirow{4}{*}{
   \begin{minipage}{.35\textwidth}
   \textit{Display and ONLY display the layer of 'boundary\_region' in Grass GIS.}
   \end{minipage}}  & \footnotesize\texttt{\ \ \{ }\\
% \end{minipage}}  & \footnotesize\texttt{[} \\
% &   & \footnotesize\texttt{\ \ \{ }\\
&   & \footnotesize\texttt{\ \ \ \ "type":"info",~}\\
&   & \footnotesize\texttt{\ \ \ \ "key":"lambda dump:len(dump['layers'])",~}\\
&   & \footnotesize\texttt{\ \ \ \ "value":1~}\\
&   & \footnotesize\texttt{\ \ \},\{"type":"info"} \\
% &   & \footnotesize\texttt{\ \ \{ }\\
% &   & \footnotesize\texttt{\ \ \ \ "type":"info",~}\\
&   & \footnotesize\texttt{\ \ \ \ "key":"lambda dump:dump['layers'][0]['name']",~}\\
&   & \footnotesize\texttt{\ \ \ \ "value":"boundary\_region@PERMANENT"~}\\
&   & \footnotesize\texttt{\ \ \} }\\
    &   &   \\
% &   & \footnotesize\texttt{]} \\
    \midrule
    \multirow{4}{*}{\raisebox{-3cm}{\includegraphics[width=5.5cm]{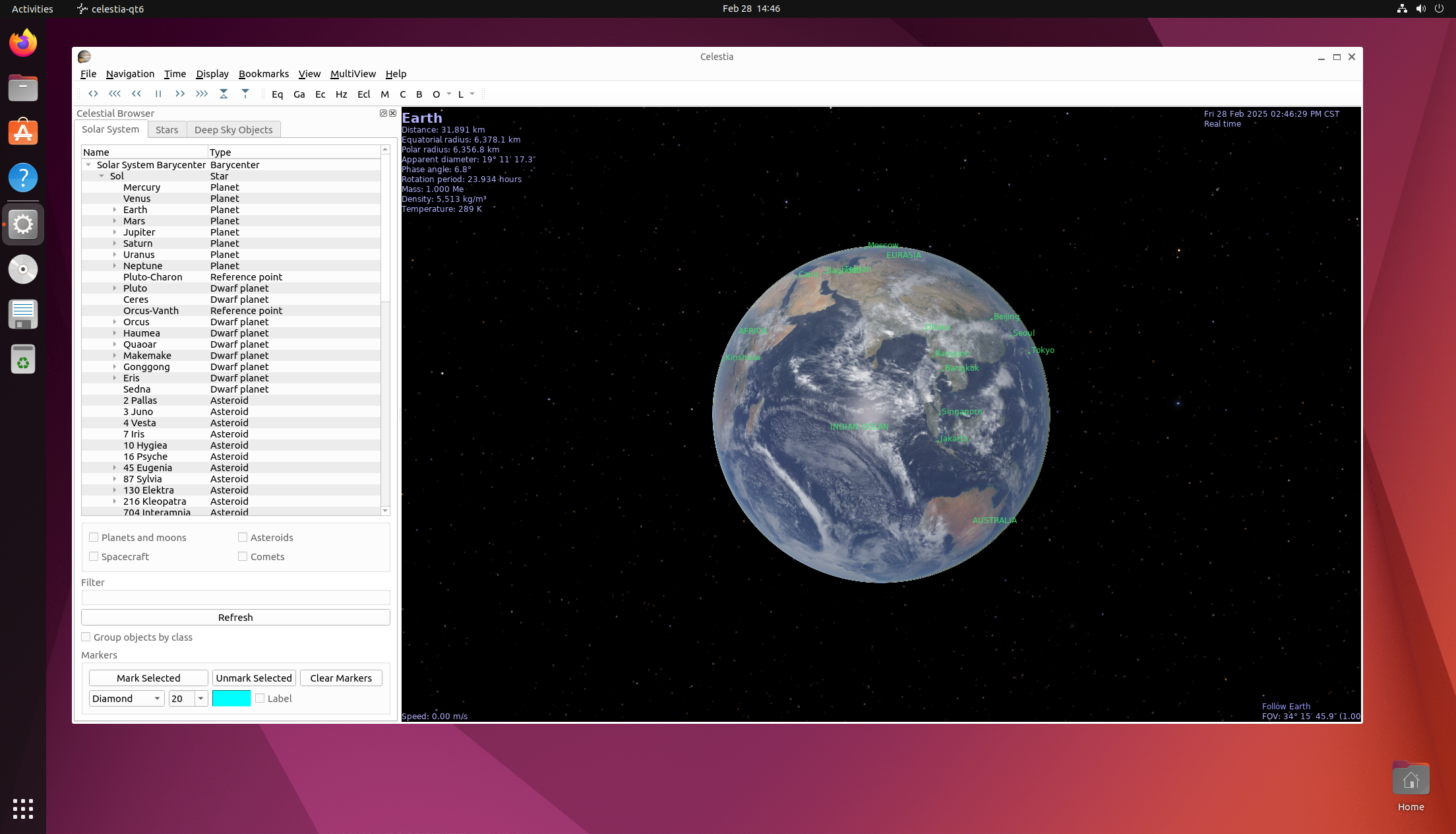}}} &  \multirow{4}{*}{
   \begin{minipage}{.35\textwidth}
   \textit{Set the Julian date to 2400000 in Celestia.}
   \end{minipage}} 
    & \footnotesize\texttt{\ \ \{ }\\ 
    &   & \footnotesize\texttt{\ \ \ \ "type":"info",~}\\
    &   & \footnotesize\texttt{\ \ \ \ "key":"simTime",~}\\
    &   & \footnotesize\texttt{\ \ \ \ "value":2400000,~}\\
    &   & \footnotesize\texttt{\ \ \ \ "pred":"lambda left, right:abs(left-right) < 1",~}\\
    &   & \footnotesize\texttt{\ \ \} }\\ 
    &   &   \\
    &   &   \\
    &   &   \\
  \bottomrule
\end{tabular}
}
% \vspace{-1em}
\vspace{-2pt}
\label{tab:evaluation_examples}
\end{table}

\vspace{-1em}
\paragraph{Evaluation Pipeline.} 
% Given the diversity and complexity of scientific tasks, 
% diversity and 
Given the complexity of scientific tasks, 
conventional answer‑matching metrics and even execution‑based evaluations~\citep{OSWorld,zhou2024webarena}, 
% such as those used in OSWorld~\citep{OSWorld} and WebArena~\citep{}, 
often lack the granularity required to assess workflows accurately.
For instance, as shown in Table~\ref{tab:evaluation_examples}, the rotation of a protein does not affect the correctness of visualization,
whereas computational tasks in astronomy are usually influenced by the current clock state.
Therefore, we propose a fine-grained evaluation based on both the correctness of key I/O during the workflow and the final state of the VM.

% This robust evaluation framework is specifically designed to assess agent performance on the diverse and complex scientific tasks that comprise our benchmark, which we detail below.

To handle the diverse criteria for determining task correctness (\textit{e.g.}, exact matching, range-based assessment, numerical tolerance, file comparison), we design a set of evaluation templates. For each specific task, the relevant template is then instantiated with the appropriate parameters and expected gold standard values.
% This ensures consistent validation across tasks and
This ensures both consistent validation and scalability for future extension.
More evaluation details are in Appendix~\ref{app:env_eval}.
% To handle the diverse criteria for determining task correctness (such as exact matching, range-based assessment, numerical tolerance, file comparison, etc.), we designed a set of evaluation templates, one for each criterion type. For each specific task, the relevant template is then instantiated with the appropriate parameters and expected gold standard values. This template-based approach ensures consistent validation across tasks and provides inherent extensibility for future additions of new evaluation methods or task types. We detail these templates and the full suite of evaluation methods in Appendix~\ref{app:env_eval}.

% To handle the diverse criteria for determining task correctness (such as exact matching, range-based assessment, numerical tolerance, file comparison, etc.), we designed a set of evaluation templates, one for each criterion type. This template-based approach ensures consistent validation across tasks and provides inherent extensibility for future additions of new evaluation methods or task types. We detail these templates and the full suite of evaluation methods in Appendix~\ref{app:env_eval}.

% 典型的涉及，以及用容错区间来判断的评测listed 在
% Table~\ref{tab:evaluation_examples}
% 比如，一个蛋白质旋转来旋转去
% 强调是可扩展的，新domain的软件可以被添加
% 为啥我们要这么好的评测，因为科学软件的IO空间很复杂，比如里面有3D的东西
% \vspace{-1em}
\section{\ours Benchmark}
In this section, we present the covered domains, the annotation pipeline, 
and statistics of the benchmark constructed based on the \ours environment.

% \zy{Can we add some sentences to describe our domain selection principle? i.e., why each app is important and selected.}
\vspace{-0.5em}
\subsection{Domain and Task Coverage}
As a pioneering benchmark for scientific exploration, 
\ours spans six domains selected for their relevance to key stages of the scientific workflow, such as simulation, modeling, prediction, and knowledge~\citep{ai4science2023impact}. 
% These choices are informed by efforts on LLMs for science~\citep{ai4science2023impact}.
In selecting software for each domain,
we consider not only its representativeness, but also practical criteria for evaluation: open-source availability, \atree compatibility, and no requirement for user authentication.
\begin{enumerate}[leftmargin=*,label=(\arabic*),itemsep=0pt,parsep=1pt,topsep=1pt]
    % \item {\bf Biochemistry.} We utilize \texttt{UCSF ChimeraX}~\citep{goddard2018chimerax,meng2023chimerax}, a tool for the interactive analysis of molecular structures, equipped with functionalities such as structure prediction of biomolecules and their complexes via AlphaFold~\citep{jumper2021alphafold}.
    \item {\bf Biochemistry.} We employ~\texttt{UCSF ChimeraX}~\citep{goddard2018chimerax,meng2023chimerax}, a molecular 
    % visualization and 
    analysis tool that supports structural modeling~(\textit{e.g.}, AlphaFold~\citep{jumper2021alphafold}).
    The tasks assess the agent’s ability to 
    % interpret and 
    manipulate biomolecular structures, as well as to reason over spatial conformations and biochem annotations.
    % It assesses the agent's ability in biocomputing, 
    % as well as interpret and manipulate complex biomolecular models.
    \item {\bf Algebra.} \texttt{KAlgebra} is employed to evaluate the agent's potential in symbolic mathematics.
    % requiring an understanding of algebraic operations and the ability to interpret graphical representations, 
    % thus demanding both numerical and visual reasoning ability.
    Tasks involve executing algebraic expressions, interpreting plots, and manipulating symbolic functions. These scenarios require the agent to exhibit strong mathematical symbolic reasoning and visual grounding capability.
    % \item {\bf Theorem Proving.} \texttt{Lean 4}~\citep{leonardo2021lean4} serves as a platform to evaluate the agent's capacity for formal mathematical reasoning and proof construction, demanding precision and logical coherence in abstract settings.
    \item {\bf Theorem Proving.} We use \texttt{Lean 4}~\citep{leonardo2021lean4} as a proof assistant to assess agents' abilities in formal logic and deductive reasoning. The ATP tasks in this category emphasize syntactic precision and logical coherence, evaluating the agent’s capability to generate semantically valid formal proofs.
    \item {\bf Geographic Information System.} \texttt{GrassGIS}, a computational engine for raster, vector, and geospatial processing, is included to examine the agent's skills in understanding terrain, hydrology, and handling spatio-temporal data, 
    with support for functions such as ecosystem modeling.
    \item {\bf Astronomy.} We integrate \texttt{Celestia},
    a planetarium software simulating real-world astronomical scenarios.
    % is utilized to challenge the agent's knowledge of the cosmos and celestial objects, as well as its ability to manage multi-object simulations and spatial perception. 
    Agents must demonstrate temporal-spatial awareness and knowledge of the cosmos and celestial objects by tracking planetary systems, simulating orbital events, and querying object metadata across time and space.
    \item {\bf Scientific Documentation.} 
    % To simulate research documentation workflows, we incorporate \texttt{TeXstudio} technical writing. 
    % In standalone tasks, agents are expected to generate well-structured research abstracts, plots, and reports. In cross-application scenarios, \texttt{TeXstudio} is coupled with upstream analysis tools to evaluate whether the agent can summarize experiment results and compose human-readable scientific narratives. These tasks assess the agent’s ability in multi-step, multi-modal synthesis—bridging data analysis with formal scientific communication.
    To simulate research documentation workflows, we adapt and incorporate \texttt{TeXstudio} to assess the agent’s technical writing capabilities. In standalone tasks, agents are expected to compose well-structured abstracts, generate plots, and produce formal reports based on provided instructions. In cross-application scenarios, \texttt{TeXstudio} is coupled with the aforementioned software to evaluate whether agents can extract meaningful insights from experiments and synthesize them into coherent narratives.
\end{enumerate}

These domains enable evaluating a science agent's capabilities across multiple dimensions, 
including visual / textual reasoning, math, coding, tool use, spatial understanding, domain-specific knowledge, and more.  
Additionally, to explore the potential for end-to-end scientific automation, 
documentation tasks are integrated with other domains to support cross-application workflows—such as automatically generating an experimental report based on completed upstream tasks.
More details about the software platforms used to instantiate and convey the tasks in \ours are provided in Appendix~\ref{app:benchmark_software}.

\begin{figure}[htb]
    \centering
    % \vspace{-0.5em}
    \includegraphics[scale=0.343]{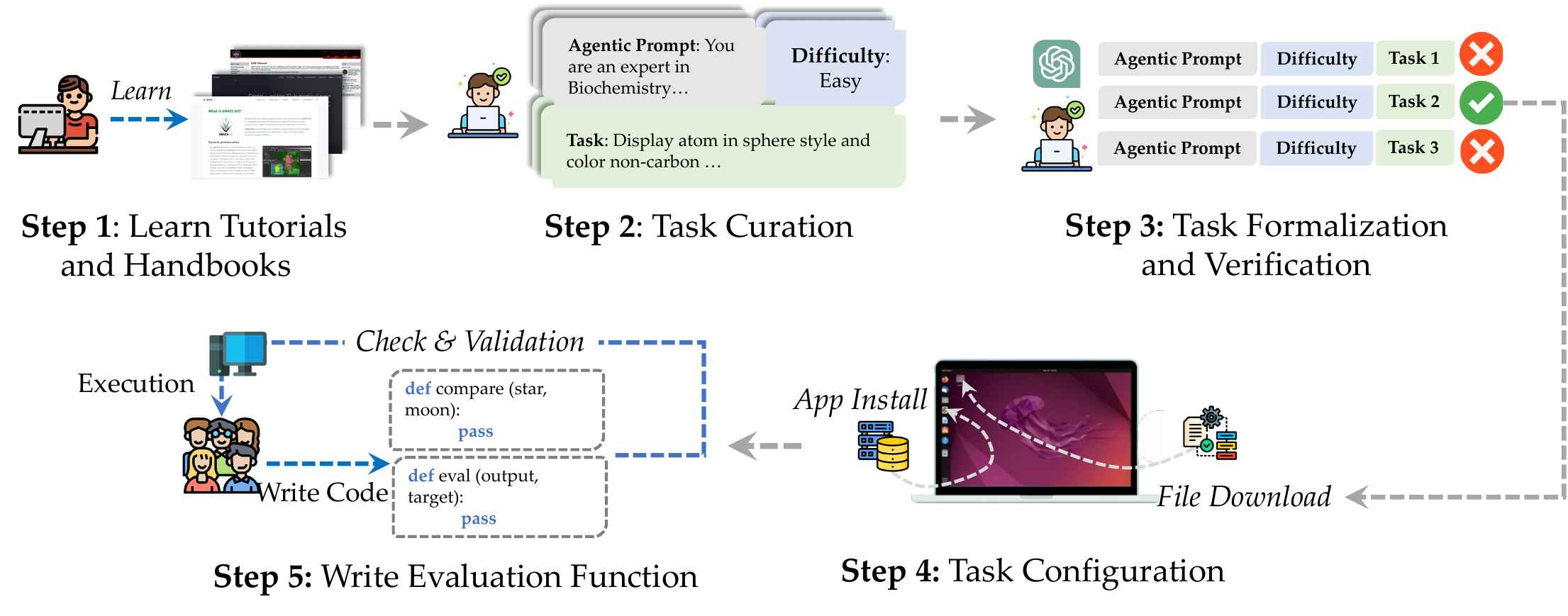}
    % \vspace{-1em}
    \caption{
    The annotation pipeline of the tasks in \ours benchmark. 
    }
    \label{fig:scienceboard_anno}
    \vspace{-1em}
\end{figure}

% \sqs{how about open a new para to emphasize our cross-app?}

% \paragraph{Cross-Application Workflows.}
% Unlike prior benchmarks that confine agent behavior to a single application, SCIENCEBOARD supports multi-application workflows where the agent must autonomously switch between software, transfer intermediate results, and maintain contextual consistency across tasks. For example, an agent may perform molecular visualization in ChimeraX and then summarize findings in TeXStudio. This setting better aligns with real-world scientific workflows and poses greater challenges in planning, memory, and abstraction.

% 对数据库
% cross-app interactions found in real-world computer usage
\vspace{-0.5em}
\subsection{Task Annotation Pipeline}
\label{sec:task_anno}
To effectively construct tasks that are appropriately challenging, diverse, and aligned with the features of scientific software, we leverage a pipeline that spans from training annotators with tutorials and handbooks to conducting execution-based validation, 
as shown in Figure~\ref{fig:scienceboard_anno}. 
% The specific pipeline is as follows:
% \ben{Make the following headings consistent. Make either all "instructions" or all "nouns". E.g., either "Learn from Tutorials and Handbooks" or "Basic Operation Learning". Note that it should be "learn from tutorials" instead of "learn tutorials".}

\begin{enumerate}[leftmargin=*,label=(\arabic*),itemsep=0pt,parsep=1pt,topsep=1pt]
    \item {\bf Tutorial Learning.}
    Five annotators initially collect and learn from tutorials and handbooks related to the software. After that, each annotator studies and explores a software's basic unit operations, \textit{e.g.}, plotting the Bernoulli lemniscate in KAlgebra. Details are in Appendix~\ref{app:benchmark_tasks}. 
    % Details of these tutorials and handbooks are in Appendix~\ref{app:benchmark_tasks}. 
    \item {\bf Task Curation.} Each annotator selects a scientific software, installs it within \ours, and begins drafting task instructions based on its functionalities. 
    Task types include but are not limited to: configuration, simulation, QA, and domain-specific expertise. Each task is tentatively assigned a difficulty. Thereafter, agentic prompts aligned with the drafted tasks will be curated.
    \item {\bf Formalization and Selection.} Different annotators exhibit varying linguistic habits, we employ ChatGPT to standardize the task format. Annotators then conduct a cross-check, excluding those lacking diversity, poor executability, or non-unique answers, to finalize the set of tasks for use.
    \item {\bf Configuration Function Writing.}
    The purpose of this step is to initialize the software and provide specific contexts, \textit{e.g.}, supplying a map for GIS tasks or a protein sequence for biochemistry tasks. Annotators will write a set of functions for each software to modify the VM status,
    \textit{i.e.}, the internal state of the software, along with general configuration functions (\textit{e.g.}, downloading required files). Tasks commence only after all initialization have been successfully executed.
    \item {\bf Evaluation Function Writing and Validation.} 
    Evaluation functions are developed to assess task outcomes rigorously. 
    As described in Section~\ref{sec:env_details},
    evaluations are state-based, with functions derived from a base evaluator template.
    Annotators retrieve the task state from the VM and assess it based on criteria such as 
    % system state, 
    I/O matching and predefined ranges.
    The function returns either ``task complete'' or ``task fail.''
    Cross-validation is performed for consistency, with each task executed by two randomly selected annotators on separate VMs. The results are analyzed to ensure the evaluator’s correctness, even under intentional attempts by annotators to deceive the system.
\end{enumerate}

% Write evaluation functions
\vspace{-1em}
\subsection{Task Statistics}
\label{sec:task_stats}
\begin{figure}[ht]
\vspace{-6pt}
    \centering
    \begin{minipage}{0.48\textwidth}
        \centering
        \captionof{table}{Statistics of \ours. }
        % \small
        \scalebox{0.85}{
        \begin{tabular}{lc}
            \toprule
            \textbf{Task Type} & \textbf{Statistics} \\
            \midrule
            \textbf{Total Tasks} & \textbf{\numtasks (100\%)} \\
            - GUI  & 38 (22.5\%) \\
            - CLI & 33 (19.5\%) \\
            - GUI + CLI & 98 (58.0\%) \\
            \midrule
            \textbf{Difficulty} &  \\
            - Easy & 91 (53.8\%) \\
            - Medium & 48 (28.4\%) \\
            - Hard & 28 (16.6\%) \\
            - Open Problems & 2 (1.2\%) \\
            \midrule
            \textbf{Instructions} &  \\
            Avg. Length of Task Instructions & 20.0\\
            Avg. Length of Agentic Prompt & 374.9\\
            \midrule
            \textbf{Execution} &  \\
            Avg. Steps & 9.0\\
            Avg. Time Consumption & 124(s)\\
            \bottomrule
        \end{tabular}
        }
        \label{tab:scienceboard_statistics}
    \end{minipage}
    \hspace{5mm}
    \begin{minipage}{0.46\textwidth}
        \centering
        \includegraphics[width=0.995\textwidth]{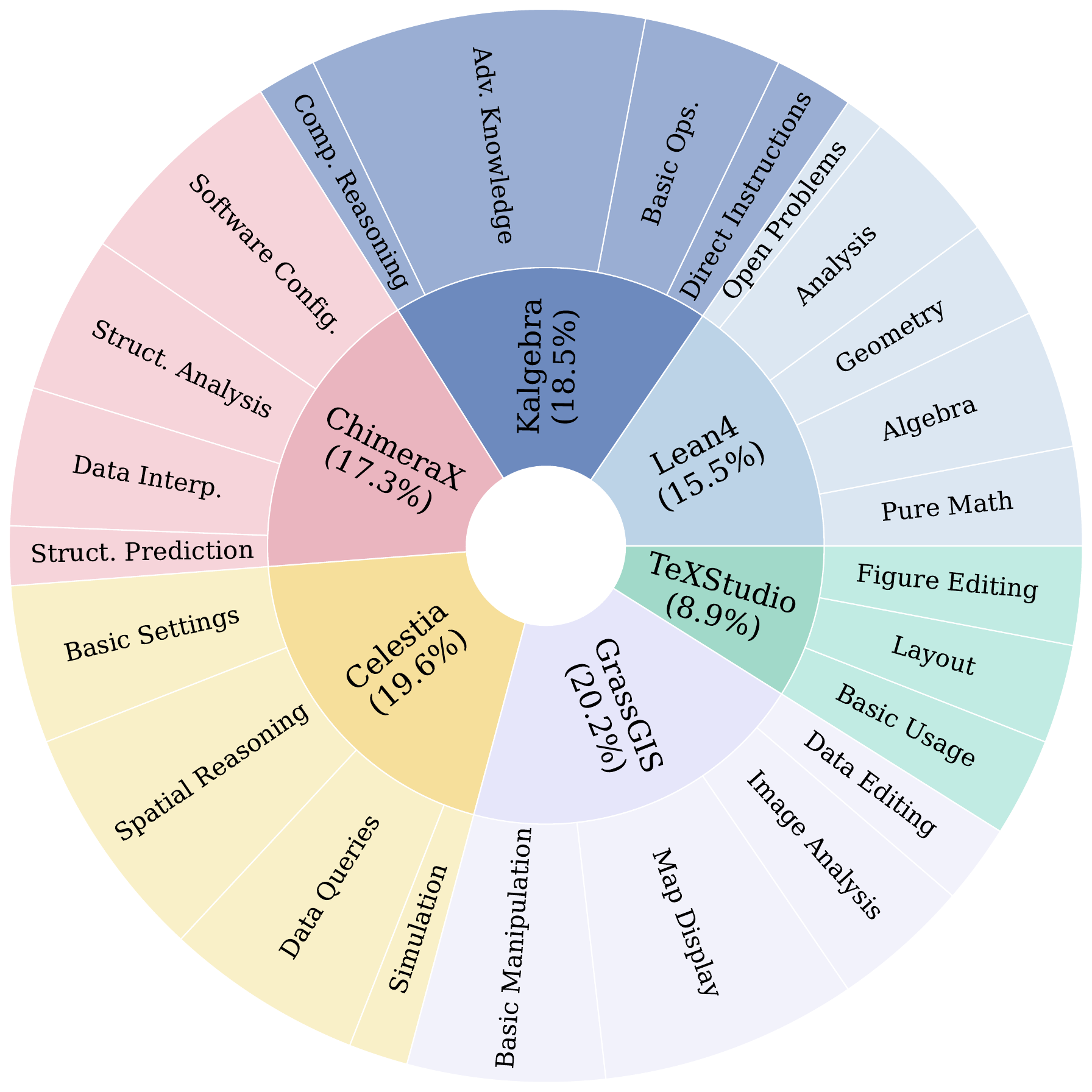}
        \caption{Distribution of tasks in \ours benchmark.}
        \label{fig:domain_pie}
    \end{minipage}
    \vspace{-9.5pt}
\end{figure}

The task statistics of \ours benchmark are presented in Table~\ref{tab:scienceboard_statistics}. Specifically, 
% our benchmark
it comprises \numtasks unique tasks across 6 domains, with task difficulty categorized into three levels. 
We curate a balanced number of tasks that are representative enough while keeping the evaluation cost manageable.
% to assess the agent’s capability in addressing domain-specific scientific challenges, 
% while keeping the evaluation cost manageable.
During annotation, 
we define multiple task types 
% within each domain 
to evaluate agents’ ability to perform diverse operation flows and leverage domain-specific knowledge. 

The distribution of task types is shown in Figure~\ref{fig:domain_pie}.
Beyond the innovation of a realistic environment, \ours benchmark also improves upon prior work in terms of task design and content diversity. 
% A detailed comparison with representative scientific benchmarks is provided in Appendix~\ref{app:benchmark_comparisons}.
More details about task diversity, stability analysis, and comparison with representative scientific benchmarks are provided in Appendix~\ref{app:bench_details}.
\section{Experiments}
\subsection{Experimental Settings} 
\label{sec:exp_setup}

% \paragraph{Backbones.}
% We employ three types of backbone models to construct \cuas for evaluation on \ours. 
% These include \textbf{proprietary models}: \gpt~\citep{hurst2024gpt}, \claude~\citep{anthropic2024claude}, \gemini~\citep{gemini2024introducing}, and \othree~\citep{openai2025o3}; \textbf{open-source models}: Qwen2.5-VL-72B-Instruct~\citep{bai2025qwen25vl}, InternVL3-78B~\citep{chen2024expanding}, and QvQ-72B-Preview~\citep{qvq72bpreview}; 
% and \textbf{GUI action models}:
% % which are VLMs specifically tailored for GUI tasks, 
% % including 
% \atlas~\citep{wu2025osatlas},
% UGround-V1-7B~\citep{gou2025navigating},
% and UI-TARS-72B-DPO~\citep{qin2025uitars}. More details regarding the selected backbones are provided in Appendix~\ref{app:eval_backbones}.

\paragraph{Backbones.}
We employ three types of backbones for agents.
% to construct \cuas. 
These include \textbf{proprietary models}: \gpt~\citep{hurst2024gpt}, \gptfive~\citep{openai2025openaigpt5card}, \claude~\citep{anthropic2024claude}, 
\gemini~\citep{gemini2024introducing}, \geminitwofivepro~\citep{comanici2025gemini} and \othree~\citep{openai2025o3}; \textbf{open-source models}: Qwen2.5-VL-72B-Instruct~\citep{bai2025qwen25vl}, InternVL3-78B~\citep{chen2024expanding}, QvQ-72B-Preview~\citep{qvq72bpreview}, and GPT-oss-120B~\citep{openai2025gptoss}; 
and \textbf{GUI action models}:
% which are VLMs specifically tailored for GUI tasks, 
% including 
\atlas~\citep{wu2025osatlas},
UGround-V1-7B~\citep{gou2025navigating},
UI-TARS-72B-DPO / UI-TARS-1.5-7B~\citep{qin2025uitars},
and GUI-Actor-7B~\citep{wu2025gui}.
% More details are available in Appendix~\ref{app:eval_backbones}.
More details in Appendix~\ref{app:eval_backbones}.

\paragraph{Observation Space.}

We follow established observation settings~\citep{OSWorld,zhou2024webarena}:
(1) full desktop screenshots;
(2) \atree, a structured text-only representation;
(3) Screenshots + \atree; and
(4) Set-of-Marks~\citep{yang2023setofmark}, which partitions images into marked regions to aid grounding.
Further details are in Appendix~\ref{app:obs_space}.

\subsection{Results}
We compare the performance of computer-use agents powered by different LLMs and VLMs on \ours, as presented in Table~\ref{tab:exp_main}.
\begin{table*}[ht]
\centering
% \vspace{-3pt}
\caption{Success rates on \ours.
We present the performance of each agent backbone across different scientific domains under various observation settings.
\raisebox{0pt}[0.8em][0.3em]{\colorbox{boxblue}{Proprietary Models}},
\raisebox{0pt}[0.8em][0.3em]{\colorbox{boxgreen}{Open-Source VLMs / LLMs}},
and 
\raisebox{0pt}[0.8em][0.3em]{\colorbox{boxred}{GUI Action Model}} are distinguished by color.
}
\label{tab:exp_main}
\small
\scalebox{0.95}{
\renewcommand{\arraystretch}{1.2} 
\begin{tabular*}{\linewidth}{@{\extracolsep{\fill}} m{1.6cm} p{2.8125cm} *{7}{S[table-format=2.2]} }
\toprule
\multirow{2}{*}{\centering \textbf{Obs.}} & \multirow{2}{*}{\centering \textbf{Model}} & \multicolumn{7}{c}{\textbf{Success Rate (↑)}} \\ 
% \cline{3-9}
\cmidrule(lr){3-9}
~ & ~ & \multicolumn{1}{c}{\raisebox{-0.5ex}{Algebra}} & \multicolumn{1}{c}{\raisebox{-0.5ex}{Biochem}} & \multicolumn{1}{c}{\raisebox{-0.5ex}{GIS}} & \multicolumn{1}{c}{\raisebox{-0.5ex}{ATP}} & \multicolumn{1}{c}{\raisebox{-0.5ex}{Astron}} & \multicolumn{1}{c}{\raisebox{-0.5ex}{Doc}} & \multicolumn{1}{c}{\raisebox{-0.5ex}{Overall}} \\
\midrule
\multirow{10}{*}{Screenshot} 
& \colorbox{boxblue}{\gptfive}  & 6.45\% & 24.14\% & 0.00\% & 0.00\% & 12.12\% & 12.50\% & 9.20\%\\
& \colorbox{boxblue}{\gpt}  & 3.23\% & 0.00\% & 0.00\% & 0.00\% & 0.00\% & 6.25\% & 1.58\%\\
& \colorbox{boxblue}{\claude}  & 9.67\% & 37.93\% & 2.94\% & 0.00\% & 6.06\% &  6.25\% & 10.48\%\\

& \colorbox{boxblue}{\geminitwofivepro}  & 6.45\% & 31.03\% & 0.00\% & 0.00\% & 0.00\% & 12.50\% & 8.33\%\\

& \colorbox{boxblue}{\gemini}  & 6.45\% & 3.45\% & 2.94\% & 0.00\% & 0.00\% & 6.06\% & 3.15\%\\
& \colorbox{boxgreen}{Qwen2.5-VL-72B}  & 22.58\% & 27.59\% & 5.88\% & 0.00\% & 9.09\% & 12.50\% & 12.94\%\\
& \colorbox{boxgreen}{InternVL3-78B}  & 6.45\% & 3.45\% & 0.00\% & 0.00\% & 0.00\% & 6.25\% & 5.46\%\\
& \colorbox{boxred}{UI-TARS-1.5-7B}  & 12.90\% & 13.79\% & 0.00\% & 0.00\% & 6.06\% & 0.00\% & 2.69\%\\
& \colorbox{boxgreen}{Qwen3-VL-235B}  & 9.68\% & 27.59\% & 0.00\% & 0.00\% & 9.09\% & 12.50\% & 9.81\%\\
& \colorbox{boxgreen}{Kimi-2.5}  & 9.68\% & 27.59\% & 0.00\% & 0.00\% & 3.03\% & 6.25\% & 7.76\%\\
\midrule
\multirow{7}{*}{\atree} 
& \colorbox{boxblue}{\gpt}  & 12.90\% & 20.69\% & 2.94\% & 0.00\% & 6.06\% & 0.00\% & 7.10\%\\
& \colorbox{boxblue}{\claude}  & 19.35\% & 34.48\% & 2.94\% & 3.85\% & 12.12\% & 0.00\% & 12.12\% \\
& \colorbox{boxblue}{\gemini}  & 9.68\% & 17.24\% & 0.00\% & 0.00\% & 0.00\% & 0.00\% & 4.49\%\\
& \colorbox{boxblue}{\othree}  & 16.13\% & 20.69\% & 2.94\% & 3.85\% & 15.15\% & 6.25\% & 10.84\% \\
& \colorbox{boxgreen}{Qwen2.5-VL-72B}  & 9.68\% & 10.34\% & 2.94\% & 0.00\% & 3.03\% & 0.00\% & 4.33\%\\
& \colorbox{boxgreen}{InternVL3-78B}  & 3.23\% & 3.45\% & 0.00\% & 0.00\% & 0.00\% & 0.00\% & 1.11\%\\
& \colorbox{boxgreen}{GPT-oss-120B}  & 19.35\% & 13.79\% & 0.00\% & 0.00\% & 9.09\% & 0.00\% & 7.04\%\\
\midrule
\multirow{7}{*}{\parbox{1.8cm}{Screenshot \\ + \atree}} 
& \colorbox{boxblue}{\gptfive}  & 41.93\% & 62.07\% & 5.88\% & 7.69\% & 15.15\% & 12.50\% & 24.20\%\\
& \colorbox{boxblue}{\gpt}  & 22.58\% & 37.93\% & 2.94\% & 7.69\% & 3.03\% & 12.50\% & 14.45\%\\
& \colorbox{boxblue}{\claude}  & 12.90\% & 41.37\% & 8.82\% & 3.85\% & 9.09\% & 18.75\% & 15.79\%\\
& \colorbox{boxblue}{\geminitwofivepro}  & 16.13\% & 55.17\% & 2.94\% & 0.00\% & 15.15\% & 12.50\% & 16.98\%\\
& \colorbox{boxblue}{\gemini}  & 16.13\% & 24.14\% & 2.94\% & 0.00\% & 18.18\% & 12.50\% & 12.32\%\\
& \colorbox{boxgreen}{Qwen2.5-VL-72B}  & 16.13\% & 20.69\% & 2.94\% & 0.00\% & 18.18\% & 12.50\% & 11.74\%\\
& \colorbox{boxgreen}{InternVL3-78B}  & 6.45\% & 3.45\% & 0.00\% & 0.00\% & 3.03\% & 6.25\% & 3.20\%\\
\midrule
\multirow{6}{*}{\centering Set-of-Mark} 
& \colorbox{boxblue}{\gpt}  & 6.45\% & 3.45\% & 0.00\% & 0.00\% & 3.03\% & 12.50\% & 4.24\%\\
& \colorbox{boxblue}{\claude}  & 16.13\% & 31.03\% & 5.88\% &  0.00\% & 6.06\% & 12.50\% & 11.93\% \\
& \colorbox{boxblue}{\gemini}  & 3.23\% & 0.00\% & 0.00\% & 0.00\% & 3.03\% & 6.25\% & 2.09\%\\
& \colorbox{boxgreen}{Qwen2.5-VL-72B}  & 6.45\% & 6.90\% & 2.94\% & 0.00\% & 3.03\% & 12.50\% & 6.36\%\\
& \colorbox{boxgreen}{QvQ-72B-Preview}  & 0.00\% & 0.00\% & 2.94\% & 0.00\% & 3.03\% & 0.00\% &0.49\%\\
& \colorbox{boxgreen}{InternVL3-78B}  & 3.23\% & 6.90\% & 2.94\% & 0.00\% & 0.00\% & 0.00\%  & 2.18\% \\
\midrule
\multicolumn{2}{c}{Human Performance}  & 74.19\% & 68.97\% & 55.88\% & 42.31\% & 51.52\% & 68.75\% & 60.27\% \\
\bottomrule
\end{tabular*}
}
\vspace{-3pt}
\end{table*}
We summarize our key empirical findings as follows:

\paragraph{Performance Hierarchy.} 
Existing agents remain far from being capable of effectively assisting human scientists in completing real-world scientific exploration tasks. Even SOTA models, such as \texttt{GPT-5} and \texttt{Gemini-2.5-Pro}, achieve an average success rate of only 20\%. 
Across various settings, open-source counterparts can partially match proprietary models.
However, they still exhibit markedly lower overall performance, with an average success rate of less than 12\% %,
and approaching nearly 0\% in some task categories.
% This considerable performance gap underscores the limitations of the current status quo in agent capabilities, demanding further research.
The gap between agent and human performance underscores the limitations of the status quo and necessitates further research.

\paragraph{Domain-Specific Performance Insights.}
Across domains, we observe clear performance imbalances: models perform moderately well on Algebra and Biochemistry but degrade notably on GIS and Astronomy. We attribute this to:
(1) Interfaces: Algebra and Biochemistry tasks often support both CLI and GUI execution, whereas GIS and Astronomy rely mainly on GUI interactions. Agents generally handle CLI commands more reliably than fine-grained GUI grounding, which demands precise visual localization.
(2) Task emphasis: Geographical and astronomical tasks involve dense visual elements (e.g., maps, star charts), making it difficult for agents to identify and reason over relevant information. This also indicates the limited 3D spatial reasoning ability of current VLMs.

\paragraph{Impact of Different Observations.}
Different observation modalities have a significant impact. 
Overall, \atree + screenshots setting yields the best performance. 
In other settings, 
Qwen2.5-VL performs exceptionally well under screenshot setting,
which we attribute to its advanced GUI ability. 
Under \atree, the attribute information of elements allows LLMs to complete certain tasks by relying solely on textual observations.
Meanwhile, 
we observe that the SoM sometimes introduces negative effects. 
It is likely that although SoM provides bounding boxes to ease grounding, 
scientific software often contains massive elements on screen (\textit{e.g.}, dense celestial objects and complex cosmic backgrounds), 
which introduces substantial noise and increases the difficulty of visual reasoning.
% \vspace{-0.5em}

\paragraph{Impact of Compute Scaling and Native Reasoning.}
To determine if test-time compute can mitigate these limitations, we further evaluate frontier models with extended generation limits and native reasoning modes in Appendix~\ref{app:compute_scaling}.
Our findings indicate that scaling inference-time compute yields tangible but bounded improvements. For instance, allocating high reasoning effort increases success rates in the Biochemistry and GIS domains. However, this scaling does not fundamentally resolve the overall performance deficit.
This suggests that the primary bottleneck for current agents in scientific workflows is not merely the depth of cognitive reasoning, but rather general computer-using and agentic capabilities, which require accurately perceiving dense, domain-specific UI elements and translating high-level scientific plans into precise, executable actions.

% \paragraph{Transfer from General Computer-Use to Scientific Workflows is 
% Weak.}

% \paragraph{From Computer-using Agents to AI Co-scientist.} 
% Agent backbones excelling on general computer-using benchmarks such as 
% OSWorld~\citep{xie2024osworld} does not straightforwardly dominate scientific tasks, despite the largely shared action space.
% This gap indicates that building agents for science requires a capability 
% axis beyond planning and action: agents additionally need {domain knowledge} (\textit{e.g.}, knowing to invoke AlphaFold in ChimeraX or the right 
% raster module in GrassGIS) and \emph{flexible GUI + CLI hybrid execution}, 
% since scientific software routinely exposes identical functionality through both modalities and choosing which to use when is itself a 
% learned skill absent from GUI-only pretraining.

\paragraph{From Computer-using Agents to AI Co-scientist.} 
Agent backbones excelling on general computer-using benchmarks such as 
OSWorld~\citep{OSWorld} do not straightforwardly dominate scientific 
tasks, despite the largely shared action space. This gap indicates that 
building agents for science requires a capability axis beyond planning and 
action: agents additionally need {domain knowledge}, spanning both 
scientific reasoning (\textit{e.g.}, interpreting a protein's spatial conformation 
or a hydrological raster) and software-specific expertise (\textit{e.g.}, knowing the right module in GrassGIS).
Also, it requires flexible GUI\,+\,CLI hybrid execution, since some scientific software exposes overlapping but non-identical functionality across the two modalities, with CLI frequently offering a more concise path for complex operations, provided the agent knows how to use it. Choosing which interface to leverage is a learned skill largely absent from GUI-only capabilities.
\section{Analysis}
\label{sec:analysis}

% \vspace{-0.75em}
To further investigate the factors influencing agents’ capabilities,
we conduct additional analysis to understand the underlying causes and the behavioral differences among heterogeneous models.

% \vspace{-1.05em}
\paragraph{GUI vs. Hybrid.}
% Some tasks inherently support both GUI and CLI as interchangeable means.
% For instance, 
% ChimeraX provides nearly full functional coverage through both its GUI and CLI for biochemistry tasks.
% To examine how current \cuas interact with such hybrid interface software, 
% we modify ChimeraX to disable CLI access, thereby enforcing GUI-only execution (under \atree + screenshot setting).
% As shown in Figure~\ref{fig:chimerax_gui_cli}, 
% \gpt and InternVL3 exhibit performance drops when CLI access is removed.
% In contrast,
\begin{wrapfigure}{r}{5.75cm}
\vspace{-0.5em}
    \centering
    \includegraphics[width=0.95\linewidth]{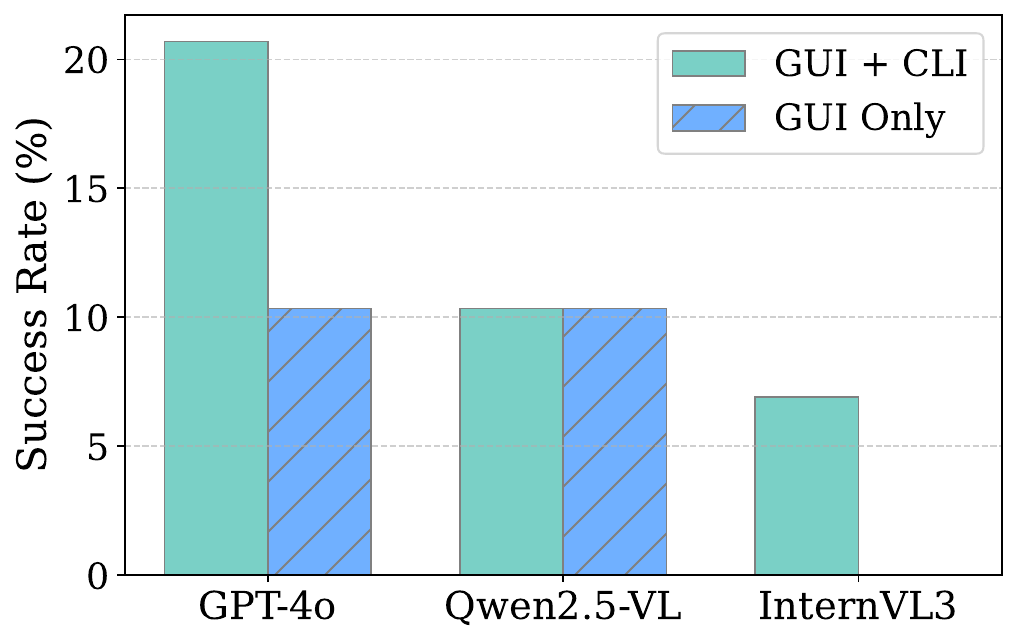}
    \vspace{-0.75em}
    \caption{GUI + CLI v.s. GUI Only.}
    \vspace{-1.75em}
    \label{fig:chimerax_gui_cli}
\end{wrapfigure}

Some tasks support both GUI and CLI as interchangeable interfaces. 
For example, ChimeraX offers nearly full functional coverage through both modes for biochemistry tasks. 
To test how \cuas handle such hybrid software, we disable ChimeraX’s CLI, enforcing GUI-only execution (\atree + screenshot).
As shown in Figure~\ref{fig:chimerax_gui_cli}, \gpt and InternVL3 suffer clear drops in performance, whereas Qwen2.5-VL remains largely unaffected, indicating better adaptation to GUI execution.

% Qwen2.5-VL remains largely unaffected, suggesting that it is well adapted to accomplishing tasks through GUI.

% Experiments are conducted under the \atree + screenshot setting, 
% and further analyses under different observations are provided in Appendix~\ref{app:analysis}.
% Further analyses under different observations are provided in Appendix~\ref{app:analysis}.

% These findings suggest that future agent designs should be more adaptable and equipped with stronger GUI capabilities to ensure robustness across both hybrid and vision-only interfaces.
% Extended analysis on other aspects and observations is presented in Appendix~\ref{app: analysis}.

These results suggest that future agents should be more adaptable and equipped with stronger GUI capabilities to remain robust across hybrid and vision-only settings. Extended analyses are provided in Appendix~\ref{app:analysis}.

% \vspace{-1.05em}
\paragraph{Disentangled Planning and Action.}
Observations from failure cases 
% and results across different settings
indicate that some models, such as \gpt, can effectively plan tasks but lack sufficient grounding capabilities.
% leading to inferior performance on \ours. 
Therefore, we explore separating planning and action.
Following existing practices~\citep{wu2025osatlas}, we configure \gpt as the planner and utilize various VLMs and GUI action models as the grounding models. 

\begin{table*}[htb]
\centering
% \vspace{-1pt}
% \caption{Success rates of LLM and VLM agents on \ours. \colorbox{boxblue}{Proprietary Model} \colorbox{boxred}{Agentic Models}}
\caption{Success rates of different VLM agent combinations under the planner + grounding model setting on \ours.
The observation setting used in this experiment is screenshot.
Colors denote
% \colorbox{boxblue}{Proprietary Models} and \colorbox{boxred}{GUI Action Models.} \sqs{try to complete this for nips submission}}
\raisebox{0pt}[0.8em][0.3em]{\colorbox{boxblue}{Proprietary Models}}, \raisebox{0pt}[0.8em][0.3em]{\colorbox{boxgreen}{Open-Source VLMs}} and \raisebox{0pt}[0.8em][0.3em]{\colorbox{boxred}{GUI Action Models.}} }
\vspace{-0.5em}
\label{tab:exp_main_action}
\small
\scalebox{0.8}{
\renewcommand{\arraystretch}{1.1} 
\begin{tabularx}{\linewidth}{m{2.6cm} p{2.3cm} *{5}{X}}
\toprule
% \multirow{2}{*}{\centering \textbf{Planner}} & \multirow{2}{*}{\centering \textbf{Action Models}} & \multicolumn{5}{c}{\textbf{Success Rate (↑)}} \\ 
\multirow{2}{*}{\centering \textbf{Planner}} & \multirow{2}{*}{\centering \textbf{Grounding Model}} & \multicolumn{5}{c}{\textbf{Success Rate (↑)}} \\ 
\cline{3-7}
~ & ~ & {\raisebox{-0.5ex}{Algebra}} & {\raisebox{-0.5ex}{Biochem}} & \multicolumn{1}{c}{\raisebox{-0.5ex}{GIS}} & {\raisebox{-0.5ex}{Astron}} & {\raisebox{-0.5ex}{Overall}} \\
\midrule
\multirow{5}{*}{\colorbox{boxblue}{\gpt}} 
& \colorbox{boxred}{\atlas}  & 6.25\% & 10.34\% & 0.00\% & 3.03\% & 4.92\% \\
& \colorbox{boxred}{UGround-V1-7B}  & 0.00\% & 3.45\% & 0.00\% & 3.03\% & 1.62\% \\
& \colorbox{boxgreen}{Qwen2.5-VL-72B}  & 12.50\% & 34.48\% & 11.76\% & 9.09\% & 16.96\% \\
& \colorbox{boxred}{UI-TARS-72B}  & 3.23\% & 10.34\% & 5.88\% & 6.06\% & 6.38\% \\
& \colorbox{boxred}{GUI-Actor-7B}  & 21.88\% & 44.83\% & 2.94\% & 12.12\% & 20.44\% \\
\midrule
\multicolumn{2}{c}{{\colorbox{boxblue}{\gpt}}}  & 3.23\% & 0.00\% & 0.00\% & 0.00\% & 0.81\% \\
\midrule
% \multicolumn{2}{c}{Human Performance}  & - & - & - & - & - \\
% \bottomrule
\end{tabularx}
}
% \vspace{-2pt}
\end{table*}

The results in Table~\ref{tab:exp_main_action} show that modular approaches yield significant improvements and are promising for tackling complex and visually demanding tasks in scientific software workflows.

% in \ours

% Interface

% Some tasks support both GUI and CLI as interchangeable interfaces. For example, ChimeraX offers nearly full functional coverage through both modes for biochemistry tasks. To test how \cuas handle such hybrid software, we disable ChimeraX’s CLI, enforcing GUI-only execution (\atree + screenshot). As shown in Figure~\ref{fig:chimerax_gui_cli}, \gpt and InternVL3 suffer clear drops in performance, whereas Qwen2.5-VL remains largely unaffected, indicating better adaptation to GUI-based execution.

% \vspace{-0.75em}
\section{Conclusion}
% We propose \ours, a first-of-its-kind realistic environment designed for scientific exploration, 
% enabling agents to interact with integrated scientific software through both GUI and CLI.
We propose \ours, a first-of-its-kind realistic environment designed to empower autonomous agents in scientific exploration with rigorous validation.
Building upon our infrastructure, 
we curate a highly challenging benchmark of diverse scientific tasks meticulously crafted by human experts.
Through extensive experiments and analysis, 
we found that even state-of-the-art \cuas perform significantly below human-level proficiency.
% Although the realization of autonomous research assistants remains a distant goal, 
% we believe that this work constitutes a vital contribution to the advancement of AI-powered scientific discovery.
Although the realization of autonomous agents for scientific discovery remains a distant goal, this work offers actionable insights for future development, and we believe it constitutes advancing AI-powered scientific discovery.

% \newpage
% \input{Appendices/reproducibility}
\section*{Ethics statement} 
Computer-using agents operating in live OS environments could potentially affect the normal functioning of the system.
This is non-negligible in scientific workflows, where a poorly controlled agent could potentially misconfigure experiments, corrupt sensitive research data, or even lead to irreversible data loss.
However, considering that all settings in this work are conducted within isolated virtual environments, we do not view this as a concern.
% Ethics statement

\section*{Acknowledgement}
\label{sec:ack}
We would like to express our sincere gratitude to the anonymous reviewers from ICLR and the WCUA@ICML workshop for their insightful comments and valuable suggestions, which have significantly helped improve the quality of this work.
This research is supported by the WYNG Foundation (HKU 25AG100407). We gratefully acknowledge their support.

\bibliography{iclr2026_conference}
\bibliographystyle{iclr2026_conference}

\appendix
\section*{Large Language Model Usage}

In this submission, we employed LLMs to aid and polish writing, including grammar and typo checking, as well as for identifying related works.
\section*{Limitations and Broader Impacts}
\label{app:lim}

% \paragraph{Limitations.}
As a pioneering effort marking the early stages of integrating \cuas into scientific workflows, it is important to acknowledge certain limitations.
While our current evaluation, 
based on both VM states and key I/O correctness, provides robust validation, 
its reliance on a binary success flag may not fully capture process correctness or partial task completion (\textit{e.g.}, an agent succeeding in most steps but failing at a final one).
Introducing a ``partial credit'' could offer more granular evaluation, but accurately defining and implementing such a system for open-ended, OS-level tasks within diverse scientific software presents significant challenges due to vast state / action spaces. One potential direction for improvement is to introduce VLMs to serve as judges capable of assigning partial credit and providing richer feedback.
We leave this as future work.

% \section*{Author Contributions}
% \label{app:contributions}

% The authors contributed to this work in the following ways:
% \paragraph{Project Leadership.} Qiushi Sun, Chang Ma, Zhiyong Wu.

% \paragraph{OS-Genesis Concept.} Qiushi Sun, Zhiyong Wu.

% \paragraph{Data Curation.} Kanzhi Cheng, Qiushi Sun, Fangzhi Xu, Yian Wang, Zichen Ding, Liheng Chen, Chengyou Jia, Zhoumianze Liu, Chuanyang Jin.

% % \paragraph{Model and Training.} Zichen Ding, Kanzhi Cheng, Qiushi Sun, Zhenyu Wu.

% \paragraph{Experiments and Analysis.} Qiushi Sun, Zichen Ding, Yian Wang, Kanzhi Cheng, Chuanyang Jin, Zhenyu Wu, Junxian He.

% \paragraph{Paper Writing.} Qiushi Sun, Kanzhi Cheng, Zhiyong Wu, Junxian He, Chuanyang Jin, Zichen Ding.

% \paragraph{Demos and Websites.} Qiushi Sun, Zhoumianze Liu.

% \paragraph{Discussions.} All authors participate in research discussions and provide insightful technical advice.

% \paragraph{Strategic Advice.} Ben Kao, Yu Qiao.

% \sqs{mz, zhenyu please neglect this part, I will finish this by myself.}
% real deploying的问题，system level，data protection...
\section{Discussion and Future Directions}

% \ours represents a significant step forward in leveraging autonomous digital agents to assist scientific workflows.
% Based on the findings presented in this paper, we identify the following potential directions for further development: \sqs{squeeze}
\ours represents a significant advance in using autonomous agents for scientific workflows. Our findings suggest several key directions for future research:

% A primary limitation of current agents in scientific exploration is their insufficient domain knowledge. For instance, the GUI action models we evaluated, while effective at automation, lack the specialized understanding required for complex scientific tasks. Future work should therefore focus on embedding deeper domain-specific abilities, such as improved scientific comprehension~\citep{li2024mmarxiv}, learning from technical resources like manuals, and enabling on-demand knowledge retrieval~\citep{lala2024paperqa}. A key challenge will be to effectively harmonize this specialized knowledge with general agentic capabilities~\citep{xu2024lemur}.

\paragraph{Harmonized Domain Knowledge and Agentic Capability.}
Our evaluations suggest that one contributing factor to current agents’ limitations in scientific exploration is their insufficient domain knowledge.
For instance, the GUI action models we evaluated, while effective at automation, lack the specialized understanding required for complex scientific tasks.
Therefore, future advancements may focus on enhancing domain-specific abilities, such as enhancing scientific comprehension~\citep{li2024mmarxiv}, 
learning from highly relevant resources such as manuals and tutorials, 
and enabling on-demand knowledge retrieval~\citep{lala2024paperqa}. 
A key challenge will be to effectively harmonize this specialized knowledge with general agentic capabilities~\citep{xu2024lemur}.

\vspace{-0.5em}
\paragraph{Collaborative and Specialized Agents as a Solution.}
Analysis in Table~\ref{tab:exp_main_action} indicates that even a basic modular approach of separating planning and action to different agents can yield significant performance improvements in complex scientific software workflows.
This points toward developing sophisticated multi-agent systems composed of specialized, heterogeneous agents~\citep{jia2024agentstore,ghafarollahi2024sciagents,AgentS2}.
For example, 
responsibilities could be disentangled by assigning
planning to agents capable of deep reasoning~\citep{li202512surveyreasoning}, 
action execution to specialized GUI action models~\citep{wu2025osatlas,xu2024aguvis},
and domain-specific capability to models in particular disciplines~\citep{ai4science2023impact,xia2025naturelm}. 
These agents could be plug-and-play, allowing flexible application across broader aspects of the scientific lifecycle, such as data analysis~\citep{chen2025scienceagentbench}, scientific plotting~\citep{jia2024chatgen}, and paper revision~\citep{yu2024sea}.
While promising, it also demands more sophisticated multi-agent designs to manage and coordinate the intricate and multifaceted nature of scientific tasks.

% \vspace{-0.5em}
% \paragraph{Collaborative and Specialized Agents as a Solution.}
% Analysis in Table~\ref{tab:exp_main_action} indicates that even a basic modular approach of separating planning and action to different agents can yield significant performance improvements in complex scientific software workflows.
% This finding points to a compelling direction: the development of multi-agent systems where heterogeneous agents with specialized capabilities are cohesively integrated~\citep{jia2024agentstore,ghafarollahi2024sciagents,AgentS2}.
% For example, 
% responsibilities could be disentangled by assigning
% planning to agents capable of deep reasoning~\citep{li202512surveyreasoning}, 
% action execution to specialized GUI action models~\citep{wu2025osatlas,xu2024aguvis},
% and domain-specific capability to models in particular disciplines~\citep{ai4science2023impact,xia2025naturelm}. 
% These agents could be plug-and-play, allowing flexible application across broader aspects of the scientific lifecycle, such as data analysis~\citep{chen2025scienceagentbench}, scientific plotting~\citep{jia2024chatgen}, and paper revision~\citep{yu2024sea}.
% While promising, it also demands more sophisticated multi-agent designs to manage and coordinate the intricate and multifaceted nature of scientific tasks.

\vspace{-0.5em}
\paragraph{Extending Digital Agents to Physical Laboratory.}
Current AI-assisted scientific workflows are primarily at the digital level,
focusing on tasks such as data analysis, simulation, and software control.
A natural and impactful next step is to extend the capabilities of such autonomous agents, as fostered and benchmarked in \ours, into physical laboratory environments.
This transition involves interfacing agents with robotic systems~\citep{burger2020mobile,angelos2024transform}, applying principles of embodied AI to perceive and interact with the physical world.
Agents would manipulate laboratory instruments and samples, carry out experimental protocols, and monitor physical processes in real time, 
thereby fostering a ``lab-in-the-loop''~\citep{frey2025lab} future where experimentation and AI-driven methods are mutually reinforcing.

\section{Details of \ours Environment}

% refer osworld

\subsection{Environment Setup}
\label{app:env_setup}
Virtual machines can operate their own kernel and system, enabling compatibility with a wide variety of operating systems. For experiments covered in this paper,
we utilize a Linux environment (Ubuntu 22.04.1 LTS with kernel 6.8.0-57-generic) running on x64 personal computers.

% 1. lean4+vscode 的流程基本上是用 vscode+LSP，预期做 ① 检测 theorem 行改动；② 检测 lake 编译结果、③ 检测 remaining goals 数量

% - App info
%   - version：KAlgebra、Celestia、GrassGIS 都修改过，记录的是从哪个版本开始修改
%     - Lean：version 4.14.0, x86_64-unknown-linux-gnu, commit 410fab728470, Release
%     - ChimeraX：1.9 (2024-12-11)
%     - KAlgebra：24.11.70
%     - Celestia：1.7.0 (Development snapshot, commit ea09cdca.)
%     - GrassGIS：8.4.1
%     - TeXstudio：4.8.6

%   - license
%     - Lean：Apache
%     - ChimeraX：（非）商业软件许可协议均有（他们自己编写的协议）
%     - KAlgebra/Celestia/GrassGIS/TeXstudio：GPL（2 或 3）
% 一句话介绍（从 prompt 里取来）
% LEAN_IS = "an interactive theorem prover"
% CHIMERAX_IS = "a molecular visualization software"
% KALGEBRA_IS = "a mathematical graph calculator"
% CELESTIA_IS = "a three-dimension space simulator"
% GRASSGIS_IS = "a GIS software suite used for geospatial data management and analysis, etc."
% TEXSTUDIO_IS = "an integrated writing environment for creating LaTeX documents"

\subsection{Evaluation Criteria}
\label{app:env_eval}

% As stated in Section~\ref{sec:env_details}
% \sqs{We achieve this by adapting the software accordingly
% After the agent indicates completion of the current task,
% we retrieve the task-relevant part of the software’s internal state through a task-specific evaluation function and compare it against the gold standard.}

As stated in Section~\ref{sec:env_details}, we employ a fine-grained evaluation methodology based on:
\begin{itemize}[leftmargin=*,itemsep=2pt]
    \item The final state of the VM (Determinant)
    \item I/O states and intermediate steps (Non-Determinant)
\end{itemize}
While the final state of the VM often provides a determinant measure of overall task completion, the diverse nature of I/O and intermediate steps necessitates a varied set of criteria.
The following outlines the primary principles applied for I/O correctness:

\begin{itemize}[leftmargin=*,itemsep=2pt]
\item \textbf{Exact Match:}
\begin{itemize}[leftmargin=*,itemsep=2pt]
\item Strict equality: The output or relevant state must be exactly identical to the gold standard (e.g., for specific textual outputs or numerical values).
\item Set equality of lines: For multi-line textual outputs, the content of all lines must match the gold standard, but their order may not be strictly enforced.
\item Question-answering: The agent's provided answer to a question is compared against a correct answer or set of acceptable answers.
\end{itemize}
\item \textbf{Predicate Satisfaction:} Verifying if specific information and generated outputs satisfy predefined logical conditions or predicates. This includes:
\begin{itemize}[leftmargin=*,itemsep=2pt]
\item Value Existence: A required value, file, or UI element is present as expected.
\item Value Non-Existence: A specified value, file, or UI element is correctly absent.
\item Range Check: A numerical output or parameter falls within a predefined acceptable range (often with a specified tolerance).
\end{itemize}
\item \textbf{Correct Task Failure (FAIL):} The agent correctly identifies a task as infeasible or terminates appropriately when unable to complete the objective, outputting a designated \texttt{FAIL} signal.
\item \textbf{Domain-Specific Success Markers:} For certain domains, unique success criteria are employed:
\begin{itemize}[leftmargin=*,itemsep=2pt]
\item Lean Tasks: Successful compilation of the generated Lean proof code is considered a primary indicator.
\end{itemize}
\end{itemize}

% - exact match
%   - 严格相等
%   - 多行输出，各行乱序相等
%   - 回答问题
% - 信息满足某个谓词
%   - 值存在
%   - 值不存在
%   - 落在某个范围内
% - FAIL
% - Lean：编译成功

\subsection{Selection and Modification of Scientific Software}
\label{app:benchmark_software}
% \sqs{@zhenyu selection criteria}
To ensure both technical feasibility and representative task diversity, we selected software tools based on the following criteria:

\begin{enumerate}[leftmargin=*,itemsep=2pt]
\item {\bf Accessibility.} The software must be open-source or freely available, allowing transparent integration and reproducibility of experiments.
\item {\bf GUI Compatibility.} The software must expose a usable accessibility tree (a11y tree) to support fine-grained GUI grounding and interaction.
\item {\bf Domain Representativeness.} The software should be representative of key scientific and technical domains, enabling meaningful assessment of multimodal agent capabilities across different types of tasks.
\end{enumerate}

Based on these principles, we selected the following software for each target domain:

% \begin{itemize}[leftmargin=*,itemsep=2pt]
% \item {\bf Biochemistry:} \texttt{UCSF ChimeraX}, a molecular modeling tool supporting biomolecular structure analysis.
% \item {\bf Algebra:} \texttt{KAlgebra}, a symbolic mathematics tool for algebraic computation and visualization.
% \item {\bf Theorem Proving:} \texttt{Lean 4}, a formal proof assistant for logical reasoning and verification.
% \item {\bf Geographic Information System:} \texttt{GrassGIS}, a geospatial analysis engine for raster, vector, and environmental modeling.
% \item {\bf Astronomy:} \texttt{Celestia}, a real-time space simulation platform for celestial navigation and planetary system modeling.
% \item {\bf Scientific Documentation:} \texttt{TeXstudio}, a LaTeX editor for generating structured scientific documents and experiment reports.
% \end{itemize}

\begin{itemize}[leftmargin=*,itemsep=2pt]
    \item \textbf{Lean.} A functional programming language and interactive theorem prover grounded in dependent type theory~(specifically Martin-Löf Type Theory). Lean enables formal verification of mathematical theorems and software correctness through rigorous type checking and logical inference, supporting robust development of maintainable and accurate code.
    \item \textbf{ChimeraX.} A next-generation molecular visualization software developed by UCSF, designed for detailed interactive exploration, visualization, and analysis of protein and biomolecular structures. ChimeraX enhances performance and user experience compared to its predecessor, UCSF Chimera, offering improved graphics rendering, extensibility via plugins, and streamlined workflows for structural biology research.

    \item \textbf{KAlgebra.} An educational calculator and graphical plotting application within the KDE Education Project. It supports a wide range of numerical, logical, symbolic, and analytical computations, enabling users to visualize mathematical functions interactively in both two-dimensional~(2D) and three-dimensional~(3D) environments, thus effectively bridging computational mathematics and educational usability.

    \item \textbf{Celestia.} A cross-platform, interactive real-time 3D astronomical simulation software that allows users to explore the universe through detailed, dynamic visualizations. Celestia is highly extensible via scripting, empowering educational and professional users to model and visualize celestial phenomena and space missions with precision and customization.

    \item \textbf{GrassGIS.} An advanced Geographic Information System (GIS) supporting both raster and vector geospatial data, along with powerful analytical capabilities for spatial modeling, hydrological analysis, and environmental simulations. GrassGIS includes a comprehensive Python API for automation and custom analysis, enabling complex geospatial and temporal analyses tailored to diverse research and application scenarios.

    \item \textbf{TeXstudio.} An integrated \LaTeX \;editor that provides a writing environment tailored specifically for creating and managing complex technical and scientific documents. TeXstudio enhances productivity through features such as syntax highlighting, real-time document preview, automatic reference checking, and intuitive assistance tools, greatly simplifying the process of technical writing and document preparation.
\end{itemize}

\subsection{Details of Action Space}
\label{app:action_space}

% \sqs{@zichen mianze, add a table, with pyqutogui details, and CLI within ChimeraX, Lean4, and KA, remember we have ans as an action}
The action space employed in \ours is shown in Table~\ref{tab:action_space}.
We combine standard interaction primitives (such as GUI operations) with the flexibility of system-level and application-specific Command-Line Interfaces (CLIs), and has been further expanded with several augmented actions tailored for scientific workflows.

\begin{table}[ht]
\centering
\caption{Action space of \ours environment.}  
\resizebox{0.95\textwidth}{!}{%
\begin{tabular}{@{}ll@{}}
\toprule
Action & Description \\
\midrule
\texttt{moveTo(x, y)} & Moves the mouse to the target coordinate. \\
\texttt{moveRel(x, y)} & Moves the mouse by an offset from current position. \\
\texttt{dragTo(x, y)} & Drags the mouse to the target coordinate. \\
\texttt{dragRel(x, y)} & Drags the mouse by an offset from current position. \\
\texttt{click(x, y)} & Clicks at the target coordinate. \\
\texttt{rightClick(x, y)} & Performs a right click at the target coordinate. \\
\texttt{middleClick(x, y)} & Performs a middle click at the target coordinate. \\
\texttt{doubleClick(x, y)} & Performs double clicks at the target coordinate. \\
\texttt{tripleClick(x, y)} & Performs triple clicks at the target coordinate. \\
\texttt{mouseDown(x, y, button)} & Presses a mouse button down. \\
\texttt{mouseUp(x, y, button)} & Releases a mouse button up. \\
\midrule
\texttt{DONE} & Agent decides the task is finished. \\
\texttt{FAIL} & Agent decides the task is infeasible. \\
\texttt{WAIT [n]} & Agent decides it should wait, `n' defaults to 5(s). \\
\texttt{ANS [s]} & Agent decides it should submit an answer, `s' denotes the answer. \\
\texttt{API [name, args]} & Invokes a registered API call with name and arguments. \\
\midrule
\texttt{CODE} & Run a generated code script (for in-app / system-level tasks, or custom functions). \\
% \multirow{2}{*}{Code}
% \texttt{Code} & \begin{tabular}[t]{@{}l@{}}Executes a generated code script \\ (for in-application automation, system-level tasks, \\ \quad or custom functions).\end{tabular} \\
\bottomrule
\end{tabular}
}
\label{tab:action_space}
\end{table}

% \centering
% \resizebox{1.0\linewidth}{!}{%
% \begin{tabular}{@{}ll@{}}
% \toprule
% Function & Description \\
% \midrule
% \texttt{moveTo(x, y)} & Moves the mouse to the specified coordinates. \\
% \texttt{click(x, y)} & Clicks at the specified coordinates. \\
% \texttt{write(`text')} & Types the specified text at the current cursor location. \\
% \texttt{press(`enter')} & Presses the Enter key. \\
% \texttt{hotkey(`ctrl', `c')} & Performs the Ctrl+C hotkey combination (copy). \\
% \texttt{scroll(200)} & Scrolls up by 200 units. \\
% \texttt{scroll(-200)} & Scrolls down by 200 units. \\
% \texttt{dragTo(x, y)} & Drags the mouse to the specified coordinates. \\
% \texttt{keyDown(`shift')} & Holds down the Shift key. \\
% \texttt{keyUp(`shift')} & Releases the Shift key. \\
% \texttt{WAIT} & Agent decides it should wait. \\
% \texttt{FAIL} & Agent decides the task is infeasible. \\
% \texttt{DONE} & Agent decides the task is finished. \\
% \bottomrule
% \end{tabular}
% \caption{Some examples of the mouse and keyboard actions $\mathcal{A}$ in \ours.}  
% }
% \vspace{-0.1in}
% \label{tab:action_space}
% \end{wraptable}

\subsection{Details of Observation Space}
\label{app:obs_space}
% screenshot

% screenshot + a11ytree 比如谁放在前面谁放在后之类的

% \sqs{resolution, celestia a11ytree parsing @ mianze, zhenyu}

We primarily adhere to well-established settings~\citep{OSWorld,zhou2024webarena} for observation space, encompassing:
(1) Screenshots, which consist of a full desktop screenshot as observed by human users; 
(2) \atree, a structured text-only representation without visual information, applicable for agents that take pure text input;
(3) Screenshots + \atree, a hybrid approach that combines and complements both textual and visual modalities; and 
(4) Set-of-Marks~\citep{yang2023setofmark}, a visual prompting method aimed at enhancing the visual grounding capabilities by partitioning an image into marked regions. Details are as follows:
% Details are in Appendix~\ref{app:obs_space}.

\paragraph{Screenshot.}
We capture a screenshot of the entire computer screen. For screen resolution, we set a default value of 1920×1080, and it also offers a 16:9 aspect ratio. 
Following OSWorld~\citep{OSWorld}, our environment also supports modifying the resolution of virtual machines to avoid potential memorization of absolute pixel values and to assist studies on topics like generalization across different resolutions.

\paragraph{A11ytree.}
An \atree refers to an intricate structure generated by the browser or OS accessibility APIs that renders a representative model of the content, providing a means of interaction for assistive technologies. Each node within the accessibility tree hosts important information about a UI element. In \ours, which utilizes an Ubuntu-based GNOME desktop environment, we employ the Assistive Technology Service Provider Interface~\footnote{\url{https://docs.gtk.org/atspi2/}}. Specifically, we adopt \texttt{pyatspi} to programmatically retrieve the accessibility tree on Ubuntu.

% To make complex \atree tractable, we filter out non-essential elements by their tag, visibility, and availability. Only the tag, name, text, position, and size of the remaining elements are kept and concatenated in the input.

To make complex \atree tractable, and critically, to ensure they fit within the context length of open-source models, we filter out non-essential elements. 
This filtering is performed based on element attributes such as their tag, visibility, and availability. For the elements that remain after filtering, only key information—specifically their tag, name, text, position, and size—is retained and subsequently concatenated to form the input representation for the agent.

\paragraph{Screenshot + a11ytree.}
To further enhance the action execution capabilities of \cuas, especially for models with weaker grounding abilities, we utilize a combined input of screenshots and \atree.

\paragraph{Set-of-Mark.}
% \sqs{Details of Set-of-Mark Implementation.}
We follow the official implementation of Set-of-Mark~\citep{yang2023setofmark}.
We leverage the information from the filtered \atree and mark the elements on the screenshot with a numbered bounding box. 
Following VisualWebArena~\citep{koh2024visualwebarena} and UFO~\citep{zhang2024ufo}, we further combine the annotated screenshot with the text metadata from \atree.

\section{Accessing \ours Environment}
\label{app:access}

To facilitate broader adoption and reproducibility, we provide several methods for accessing \ours environment. Researchers can choose the most suitable option based on their technical requirements and resources:

\paragraph{Direct Deployment.}
The entire framework, including all scientific software and evaluation scripts, is available for direct deployment on a native Ubuntu system. Full setup instructions and dependency lists are provided in our repository.

\paragraph{Docker Container.}
We also provide a Docker image that encapsulates the environment, making it easy to run \ours across different machines and operating systems,
which is available at \href{https://anonymous.4open.science/r/Science_Board/}{https://anonymous.4open.science/r/ScienceBoard/}.

\paragraph{Cloud Platforms.}
For scalability and powerful computational resources, \ours can be deployed on cloud platforms like Amazon Web Services (AWS). 
We will provide guidelines upon acceptance.

% We provide guidelines and AWS support at \href{https://anonymous.4open.science/r/Science_Board/}{https://anonymous.4open.science/r/ScienceBoard/}.
\section{Details of \ours Benchmark}
\label{app:bench_details}

% \subsection{Details of tasks}
% \label{app:task_details}

% % 一些原则：开源，可用性强，能获得高质量a11ytree
% % 不需要authentic user accounts

% \paragraph{Algebra.}
% % \paragraph{Kalgebra}

% \paragraph{Biochemistry.}

% \paragraph{Geographic Information System.}
% % \paragraph{Grass GIS}

% \paragraph{Astronomy.}

% \paragraph{ATP.}
% % formal methods?

% \paragraph{Documentation.}

% \sqs{@mianze, add a license table of the software}

\subsection{Task Annotation}
\label{app:benchmark_tasks}

During the task annotation process, we primarily utilize the tutorials and handbooks listed in Table~\ref{tab:tutorial_handbooks} to guide annotators in exploring the relevant domain and corresponding software and tools.
All app data collection and task creation are completed by the authors.

% \url{https://docs.kde.org/stable5/en/kalgebra/kalgebra/index.html}

% \begin{table}[htbp]
% \centering
% \renewcommand{\arraystretch}{1.5}
% \caption{Sources of the tutorials and handbooks employed in the task annotation process.}
% \begin{tabular}{cm{28em}}
% \toprule
% \textbf{Software}          & \textbf{Tutorial \& Handbook Sources}              \\ \midrule
% Kalgebra        & \url{https://docs.kde.org/stable5/en/kalgebra/kalgebra/index.html}          \\ \midrule
% \multirow{2}{*}{ChimeraX} 
% & \url{https://www.cgl.ucsf.edu/chimerax/tutorials.html}             \\ 
% & \url{https://kpwulab.com/wp-content/uploads/2022/04/chimerax-tutorial-kpwulab-2022-0429.pdf}             \\ 
% \multirow{4}{*}{Lean 4} \\ \midrule
% & \url{https://lean-lang.org/theorem_proving_in_lean4/}             \\ 
% & \url{https://leanprover-community.github.io/mathematics_in_lean/index.html}             \\ 
% & \url{https://lean-lang.org/doc/reference/latest/}                             \\ 
% \multirow{4}{*}{Grass GIS}  & \multirow{1}{*}{\url{https://docs.airbyte.com/}}                                      
% \\ \cline{2-2}
%                           & \url{https://airbyte.com/tutorials/}          
% \\ \cline{2-2}
%                           & \url{https://airbyte-public-api-docs.s3.us-east-2.amazonaws.com/rapidoc-api-docs.html}               \\ \midrule
% \multirow{4}{*}{Celestia} 
% & \url{https://celestiaproject.space/guides.html}             \\ 
% & \url{https://en.wikibooks.org/wiki/Celestia}             \\ 
% & \url{https://celestiaproject.space/docs/CELScriptingGuide/Cel_Script_Guide_v1_0g.htm}                             \\ 
% \midrule
% \multirow{2}{*}{TeXStudio} 
% & \url{https://texstudio-org.github.io/getting_started.html}             \\ 
% & \url{https://latex-tutorial.com/tutorials/}             \\ 
% \bottomrule
% % \hline
% \end{tabular}
% \label{tab:tutorial_handbooks}
% \end{table}

\begin{table}[htbp]
\centering
\renewcommand{\arraystretch}{1.5}
\caption{Sources of the tutorials and handbooks employed in the task annotation process.}
\begin{tabular}{@{}l m{29em}@{}}
\toprule
\textbf{Software} & \textbf{Tutorial \& Handbook Sources} \\
\midrule
Kalgebra & 
\url{https://docs.kde.org/stable5/en/kalgebra/kalgebra/index.html} \\
\midrule
\multirow{3}{*}{ChimeraX} & 
\url{https://www.cgl.ucsf.edu/chimerax/tutorials.html} \\
& \url{https://kpwulab.com/wp-content/uploads/2022/04/chimerax-tutorial-kpwulab-2022-0429.pdf} \\
\midrule
\multirow{4}{*}{Lean 4} & 
\url{https://lean-lang.org/theorem_proving_in_lean4/} \\
& \url{https://leanprover-community.github.io/mathematics_in_lean/index.html} \\
& \url{https://lean-lang.org/doc/reference/latest/} \\
\midrule
\multirow{2}{*}{Grass GIS} & 
\url{https://grass.osgeo.org/grass84/manuals/index.html} \\
& \url{https://neteler.gitlab.io/grass-gis-analysis/} \\
\midrule
\multirow{4}{*}{Celestia} & 
\url{https://celestiaproject.space/guides.html} \\
& \url{https://en.wikibooks.org/wiki/Celestia} \\
& \url{https://celestiaproject.space/docs/CELScriptingGuide/Cel_Script_Guide_v1_0g.htm} \\
\midrule
\multirow{2}{*}{TeXStudio} & 
\url{https://texstudio-org.github.io/getting_started.html} \\
& \url{https://latex-tutorial.com/tutorials/} \\
\bottomrule
\end{tabular}
\label{tab:tutorial_handbooks}
\end{table}
% \cline{2-2}

% ChimeraX
% 1. https://www.cgl.ucsf.edu/chimerax/tutorials.html
% 2. https://kpwulab.com/wp-content/uploads/2022/04/chimerax-tutorial-kpwulab-2022-0429.pdf

% Kalgebra
% 1. https://docs.kde.org/stable5/en/kalgebra/kalgebra/index.html

% Lean4
% 1. https://lean-lang.org/theorem_proving_in_lean4/
% 2. https://leanprover-community.github.io/mathematics_in_lean/index.html
% 3. https://lean-lang.org/doc/reference/latest/

% GrassGIS
% 1. https://grass.osgeo.org/grass84/manuals/index.html
% 2. https://neteler.gitlab.io/grass-gis-analysis/

% Celestia
% 1. https://celestiaproject.space/docs/CELScriptingGuide/Cel_Script_Guide_v1_0g.htm
% 2. https://en.wikibooks.org/wiki/Celestia

% TeXStudio
% 1. https://texstudio-org.github.io/getting_started.html
% 2. LaTeX 本身的教程一例：https://latex-tutorial.com/tutorials/

% \subsection{Task Examples}

% \sqs{@mianze, more task descriotions.}

\subsection{Task Diversity}
\label{app:benchmark_diversity}

\begin{figure}[htbp]
    \centering
    \includegraphics[width=0.32\linewidth]{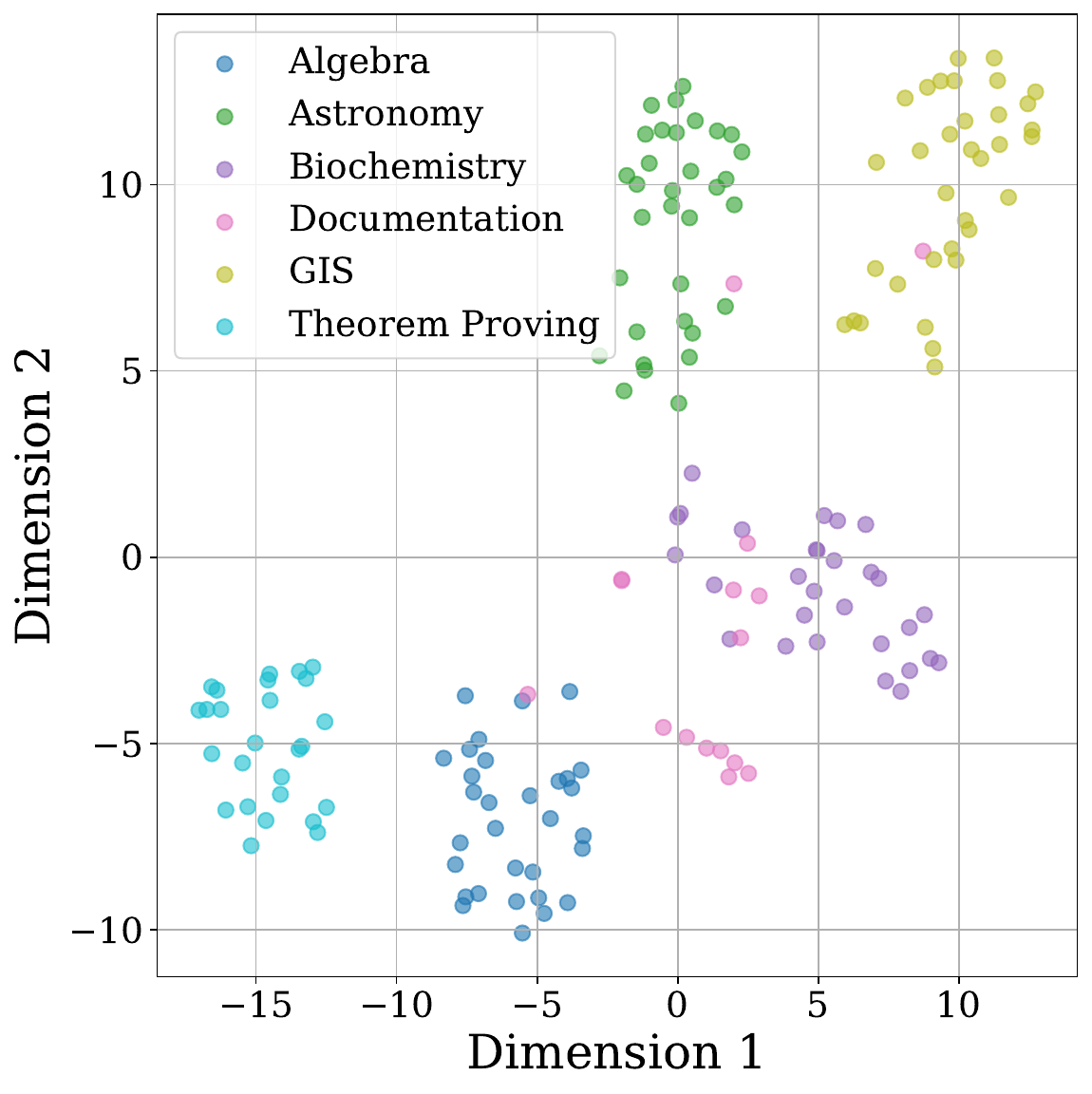}
    \includegraphics[width=0.32\linewidth]{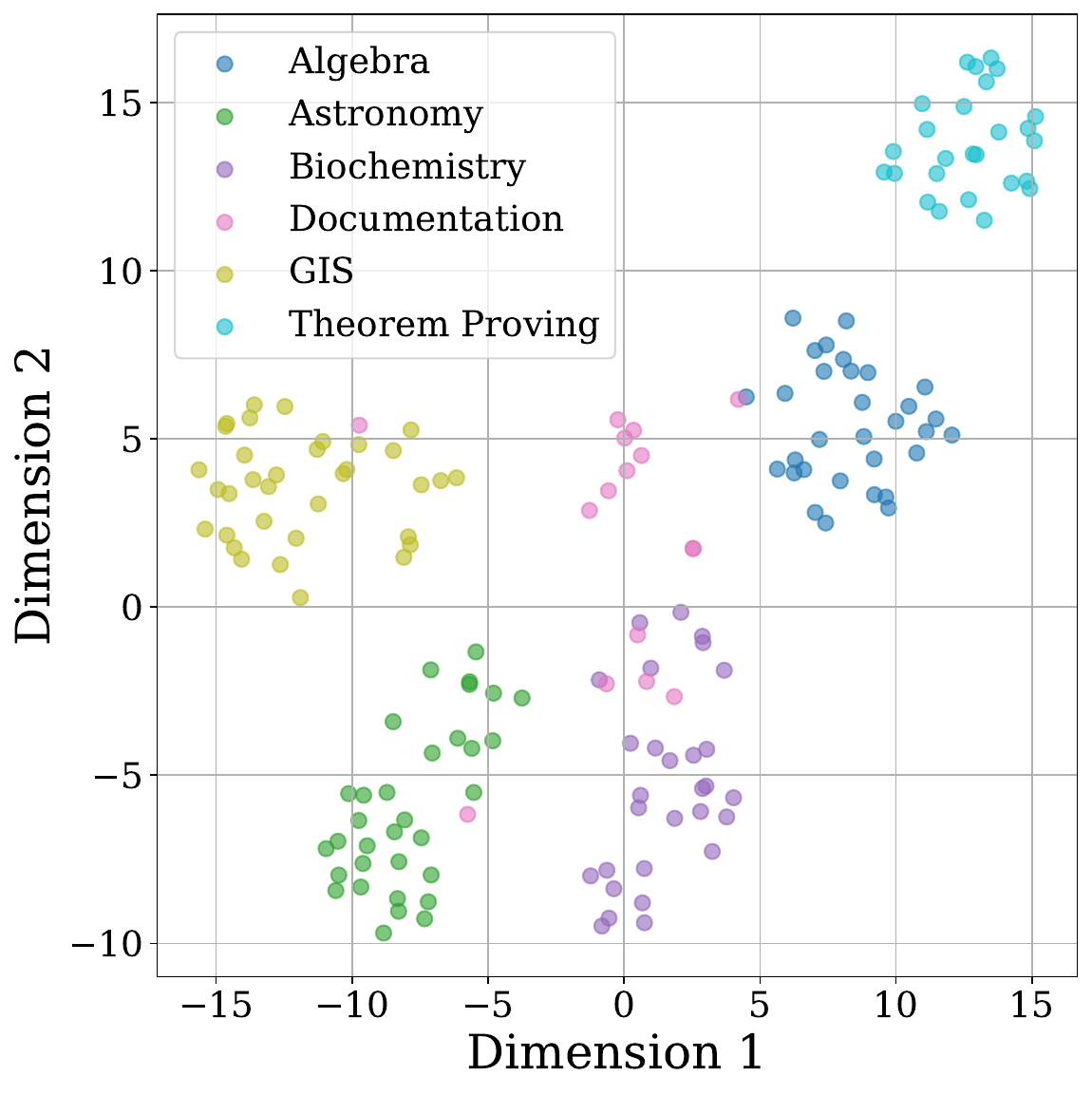}
    \includegraphics[width=0.32\linewidth]{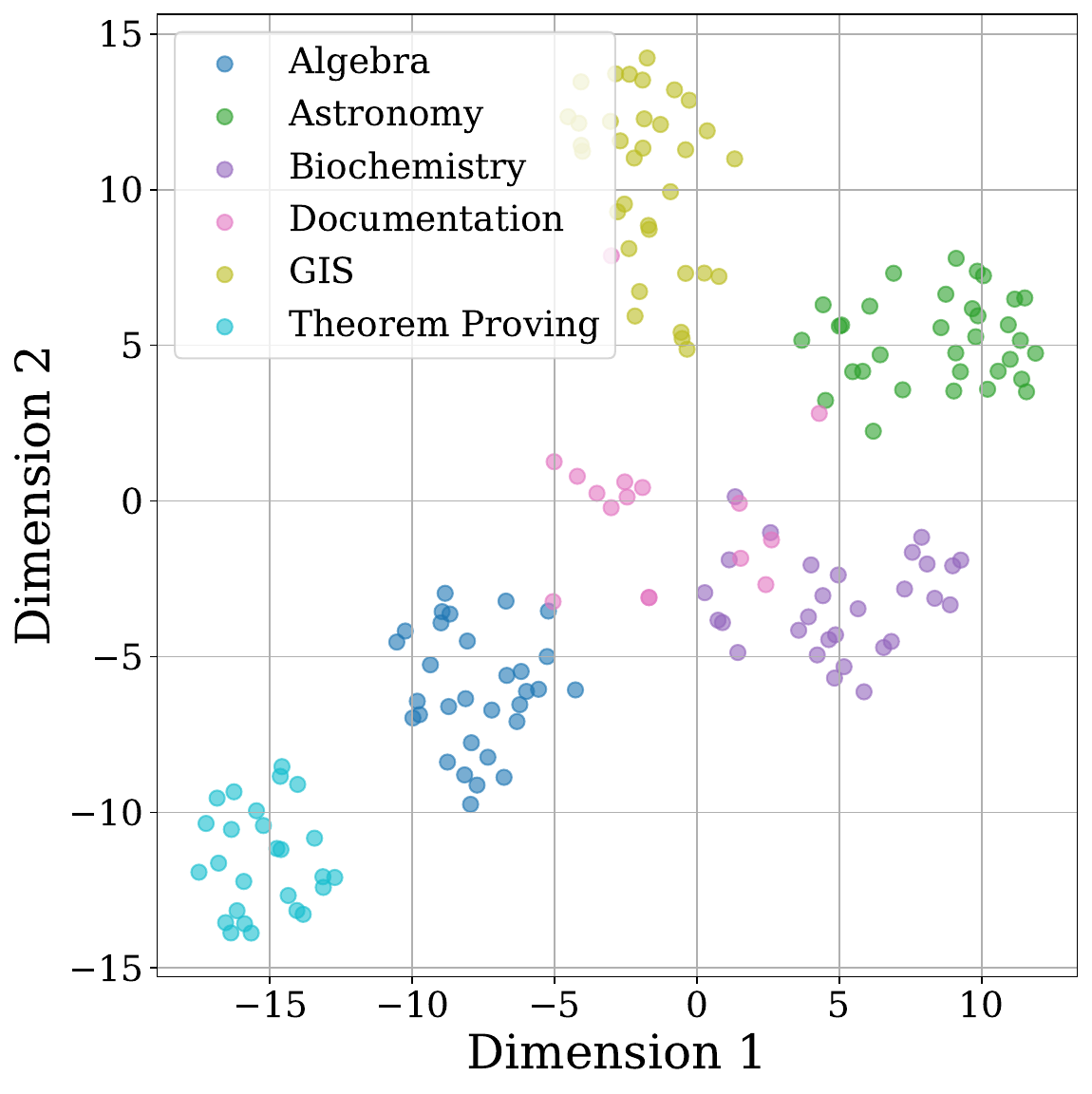}
    \caption{t-SNE visualization of task instructions distribution. The seeds of t-SNE are randomly sampled for each plot.}
    \label{fig:t_sne_visualization}
\end{figure}

To explore the diversity of tasks in \ours, we perform a t-SNE~\citep{tsne2008} visualization, 
as shown in Figure~\ref{fig:t_sne_visualization}. 
We obtain embeddings for all task instructions using \tembd and then apply t-SNE to reduce their dimensionality to two for visualization.
The semantic distribution of instructions clearly distinguishes tasks across different domains, while also revealing considerable diversity within each individual domain. Furthermore, we can observe some intersections between Scientific Documentation tasks and tasks from other domains, which reflects the presence of cross-application workflows in our benchmark.

\subsection{Comparison with Existing Benchmarks}
\label{app:benchmark_comparisons}
We compare \ours with existing well-established benchmarks for scientific tasks, as shown in Table~\ref{tab:dataset_comparison}.

\begin{table*}[ht]
    \centering
    \resizebox{0.95\linewidth}{!}{
    \begin{tabular}{lcccc}
        \toprule
        \textbf{Feature} & \makecell{\textbf{\ours}\\(our work)}& 
        \makecell{ \textbf{ScienceQA}~\citep{lu2022learn} } & 
        \makecell{ \textbf{SciCode}~\citep{tian2024scicode} } & 
        \makecell{ \textbf{ScienceAgentBench}~\citep{chen2025scienceagentbench} } 
        \\
         \cmidrule(lr){1-1}  \cmidrule(lr){2-2}  \cmidrule(lr){3-3}  \cmidrule(lr){4-4}  \cmidrule(lr){5-5} 
         \rowcolor[gray]{0.9} 
        \multicolumn{5}{c}{\textit{I/O Formats}} \\ 
          Code / Structured Input   & \cmark & \xmark & \cmark & \cmark    \\ 
          Visual Information & \cmark & \cmark & \xmark & \xmark  \\ 

          \rowcolor[gray]{0.9} 
        \multicolumn{5}{c}{\textit{Task Type}} \\ 
          Question-Answering   & \cmark & \cmark & \xmark & \xmark \\ 
          Scientific Computing   & \cmark & \xmark & \cmark & \cmark   \\ 
          GUI Automation   & \cmark & \xmark & \xmark & \xmark  \\ 
         \bottomrule
    \end{tabular}
    }
    \caption{
    {A comparison of \ours to notable and recent AI4Science benchmarks.}
    }
    \label{tab:dataset_comparison}
\end{table*}

\ours is the first to offer a realistic environment for evaluating scientific tasks. In terms of I/O, it incorporates structured code input and visual information, which are critical for simulating scientific experiment workflows. 
It also supports GUI automation, making it well-suited for visual agents to fulfill tasks like humans do.
Additionally, \ours covers a broader range of task types compared to existing works, including but not limited to question-answering and scientific computing.
These unique features make \ours both a versatile playground and an expandable framework for evaluating agents’ scientific capabilities.

\subsection{More Evaluation Script Examples}
\label{app:benchmark_eval_cases}

Beyond the evaluation cases listed in Section~\ref{sec:env_details}, Table~\ref{tab:ext_evaluation_examples} showcases a broader variety of evaluation pipelines created using our templates.

\begin{table}[htb]
\vspace{-5pt}
\caption{
More evaluation cases of \ours include exact matching, range-based assessment, and numerical tasks with tolerance.
}
\resizebox{\textwidth}{!}{
\begin{tabular}{lp{.35\textwidth}l}
  \toprule
  \textbf{Initial State} &   \textbf{Instruction}  & \textbf{Evaluation Script (Simplified)}  \\
  \midrule
  \multirow{3}{*}{\raisebox{-3.15cm}{\includegraphics[width=5.5cm]{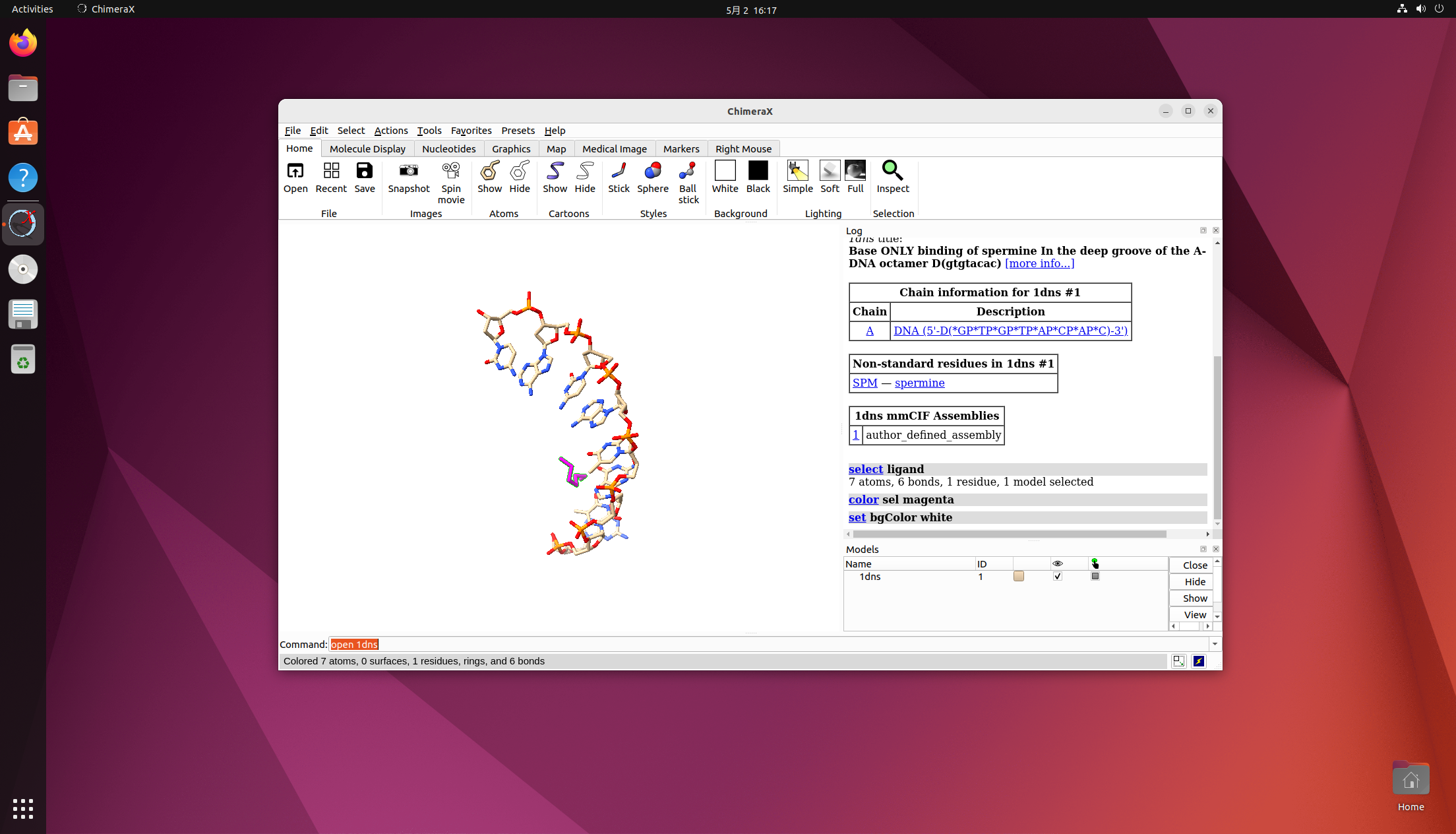}}}  & \multirow{3}{*}{\begin{minipage}{.35\textwidth}
  \textit{Select all ligand(s) and color them into magenta in ChimeraX.}
  \end{minipage}} 
    & \footnotesize\texttt{\ \ \{ } \\
    &   & \footnotesize\texttt{\ \ \ \ "type": "info",} \\
    &   & \footnotesize\texttt{\ \ \ \ "key": "sel",} \\
    &   & \footnotesize\texttt{\ \ \ \ "value": ["atom id /A:9@N1 idatm\_type N3+",} \\
    &   & \footnotesize\texttt{\ \ \ \ \ \ ...} \\
    &   & \footnotesize\texttt{\ \ \ \ ]} \\
    &   & \footnotesize\texttt{\ \ \},\{~} \\
    &   & \footnotesize\texttt{\ \ \ \ "type": "info",} \\
    &   & \footnotesize\texttt{\ \ \ \ "key": "rescolor /A",} \\
    &   & \footnotesize\texttt{\ \ \ \ "value": ["\#1/A:1 color \#d2b48c",} \\
    &   & \footnotesize\texttt{\ \ \ \ \ \ ...} \\
    &   & \footnotesize\texttt{\ \ \ \ ]} \\
    &   & \footnotesize\texttt{\ \ \} } \\
  \midrule
  \multirow{4}{*}{\raisebox{-3cm}{\includegraphics[width=5.5cm]{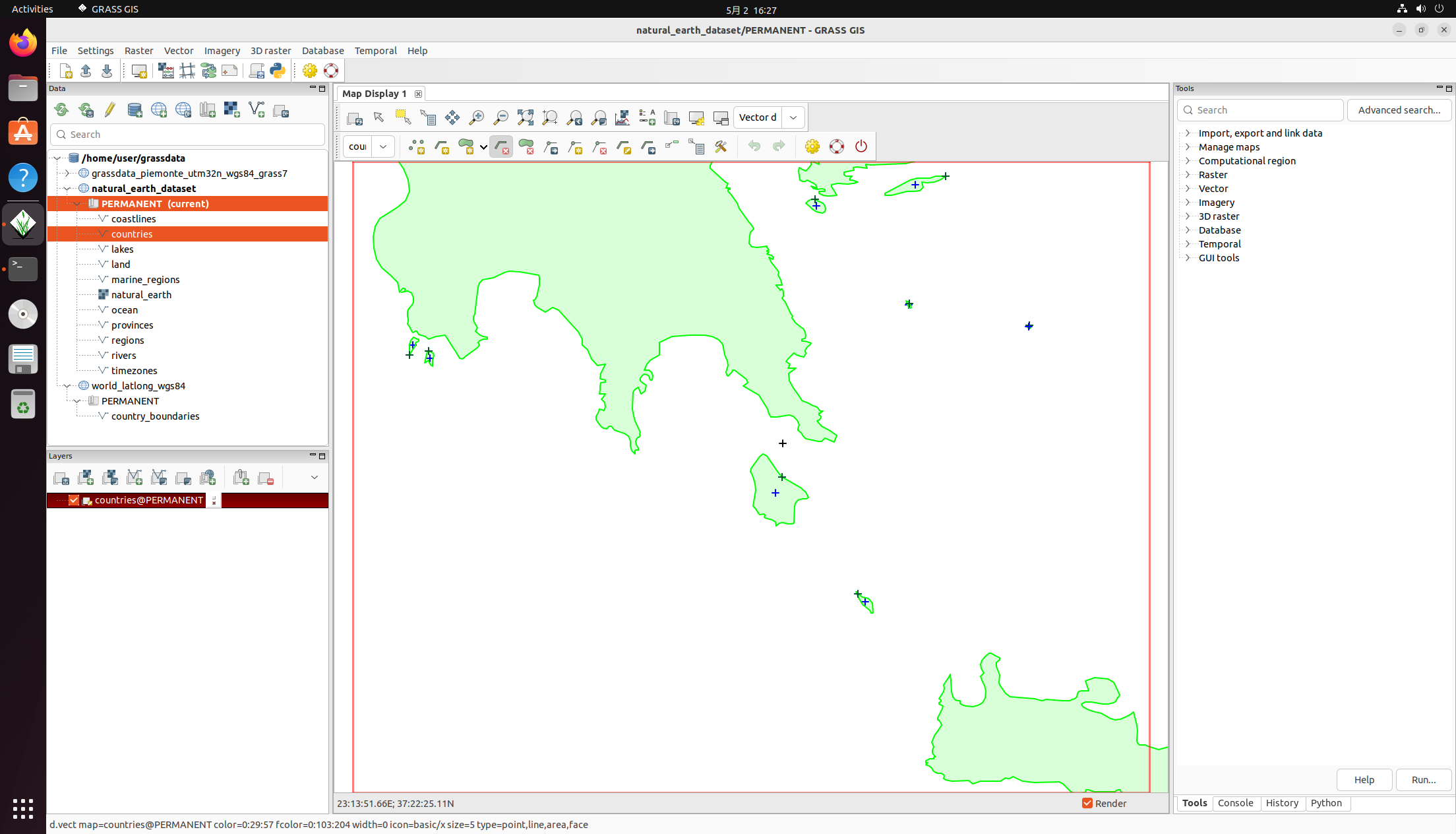}}} &  
   \multirow{4}{*}{
   \begin{minipage}{.35\textwidth}
   \textit{There is a point located in the Mediterranean Sea. Please find and delete it.}
   \end{minipage}} 
    & \footnotesize\texttt{\ \ \{ } \\
    &   & \footnotesize\texttt{\ \ \ \ "type": "db",} \\
    &   & \footnotesize\texttt{\ \ \ \ "cmd": "v.to.db",} \\
    &   & \footnotesize\texttt{\ \ \ \ "kwargs": \{} \\
    &   & \footnotesize\texttt{\ \ \ \ \ \ "flags": "p",} \\
    &   & \footnotesize\texttt{\ \ \ \ \ \ "map": "countries@PERMANENT",} \\
    &   & \footnotesize\texttt{\ \ \ \ \ \ "type": "point",} \\
    &   & \footnotesize\texttt{\ \ \ \ \ \ "option": "coor"} \\
    &   & \footnotesize\texttt{\ \ \ \ \},} \\
    &   & \footnotesize\texttt{\ \ \ \ "key": "lambda out: out.strip()",} \\
    &   & \footnotesize\texttt{\ \ \ \ "value": "cat|x|y|z\textbackslash n...|8.348947891274|0",} \\
    &   & \footnotesize\texttt{\ \ \ \ "pred": "lambda key, value: key == value"} \\
    &   & \footnotesize\texttt{\ \ \} } \\
    \midrule
    \multirow{4}{*}{\raisebox{-3cm}{\includegraphics[width=5.5cm]{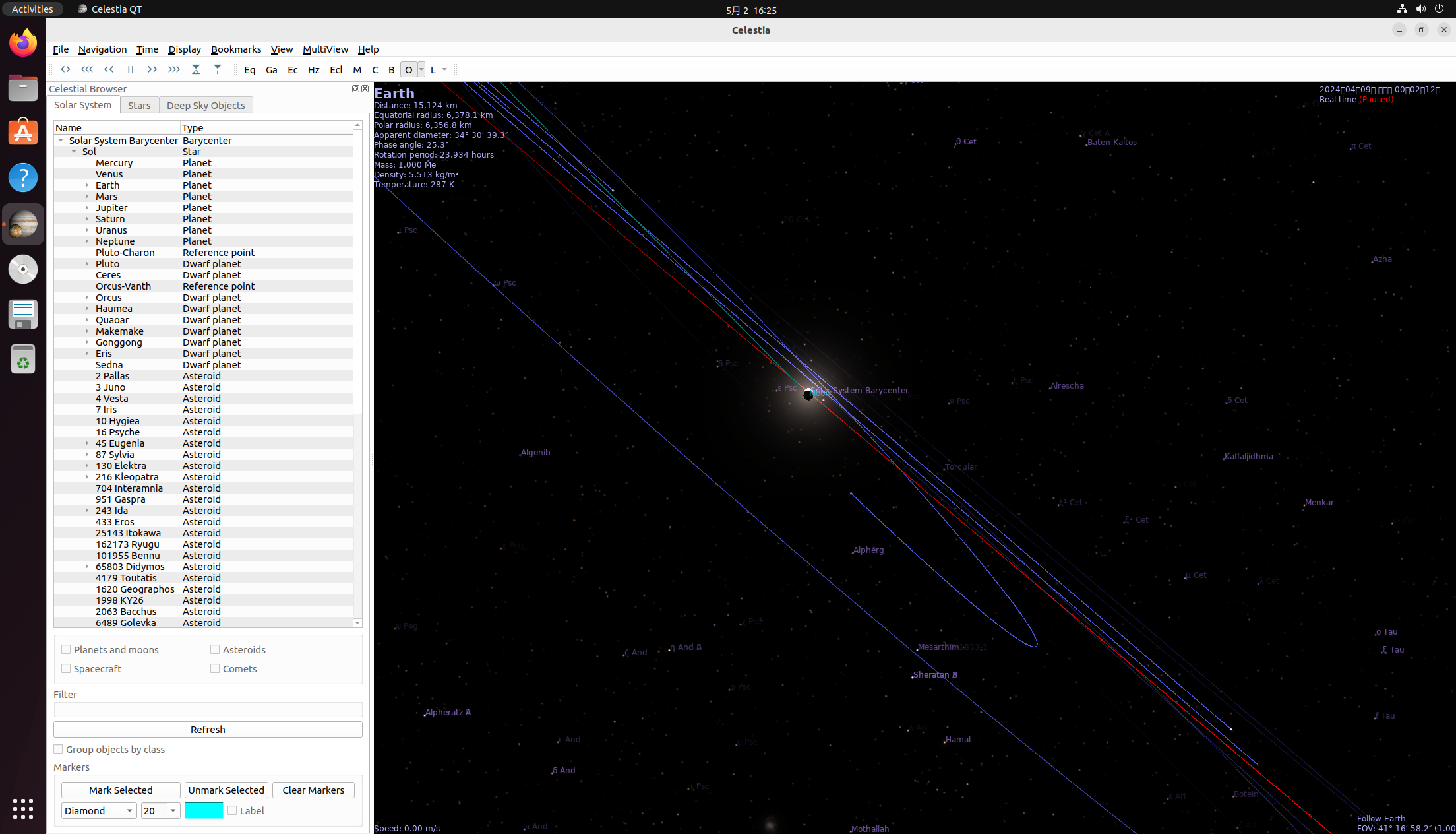}}} &  \multirow{4}{*}{
   \begin{minipage}{.35\textwidth}
   \textit{Approach to the Earth and display a solar eclipse in Celestia.}
   \end{minipage}} 
    & \footnotesize\texttt{\ \ \{ } \\
    &   & \footnotesize\texttt{\ \ \ \ "type": "info",} \\
    &   & \footnotesize\texttt{\ \ \ \ "key": "lambda ...['Earth']['distance']",} \\
    &   & \footnotesize\texttt{\ \ \ \ "value": 0,} \\
    &   & \footnotesize\texttt{\ \ \ \ "pred": "lambda k, v: abs(k - v) < 450000"} \\
    &   & \footnotesize\texttt{\ \ \},\{~} \\
    &   & \footnotesize\texttt{\ \ \ \ "type": "info",} \\
    &   & \footnotesize\texttt{\ \ \ \ "key": "lambda ...['Sol']['visible']",} \\
    &   & \footnotesize\texttt{\ \ \ \ "value": false} \\
    &   & \footnotesize\texttt{\ \ \},\{~} \\
    &   & \footnotesize\texttt{\ \ \ \ "type": "info",} \\
    &   & \footnotesize\texttt{\ \ \ \ "key": "lambda ...['Moon']['visible']",} \\
    &   & \footnotesize\texttt{\ \ \ \ "value": true} \\
    &   & \footnotesize\texttt{\ \ \},\{~} \\
    &   & \footnotesize\texttt{\ \ \ \ "type": "info",} \\
    &   & \footnotesize\texttt{\ \ \ \ "key": "lambda ...",} \\
    &   & \footnotesize\texttt{\ \ \ \ "value": 0.99,} \\
    &   & \footnotesize\texttt{\ \ \ \ "pred": "lambda key, value: key > value"} \\
    &   & \footnotesize\texttt{\ \ \} } \\
    \midrule
    \multirow{4}{*}{\raisebox{-3.25cm}{\includegraphics[width=5.5cm]{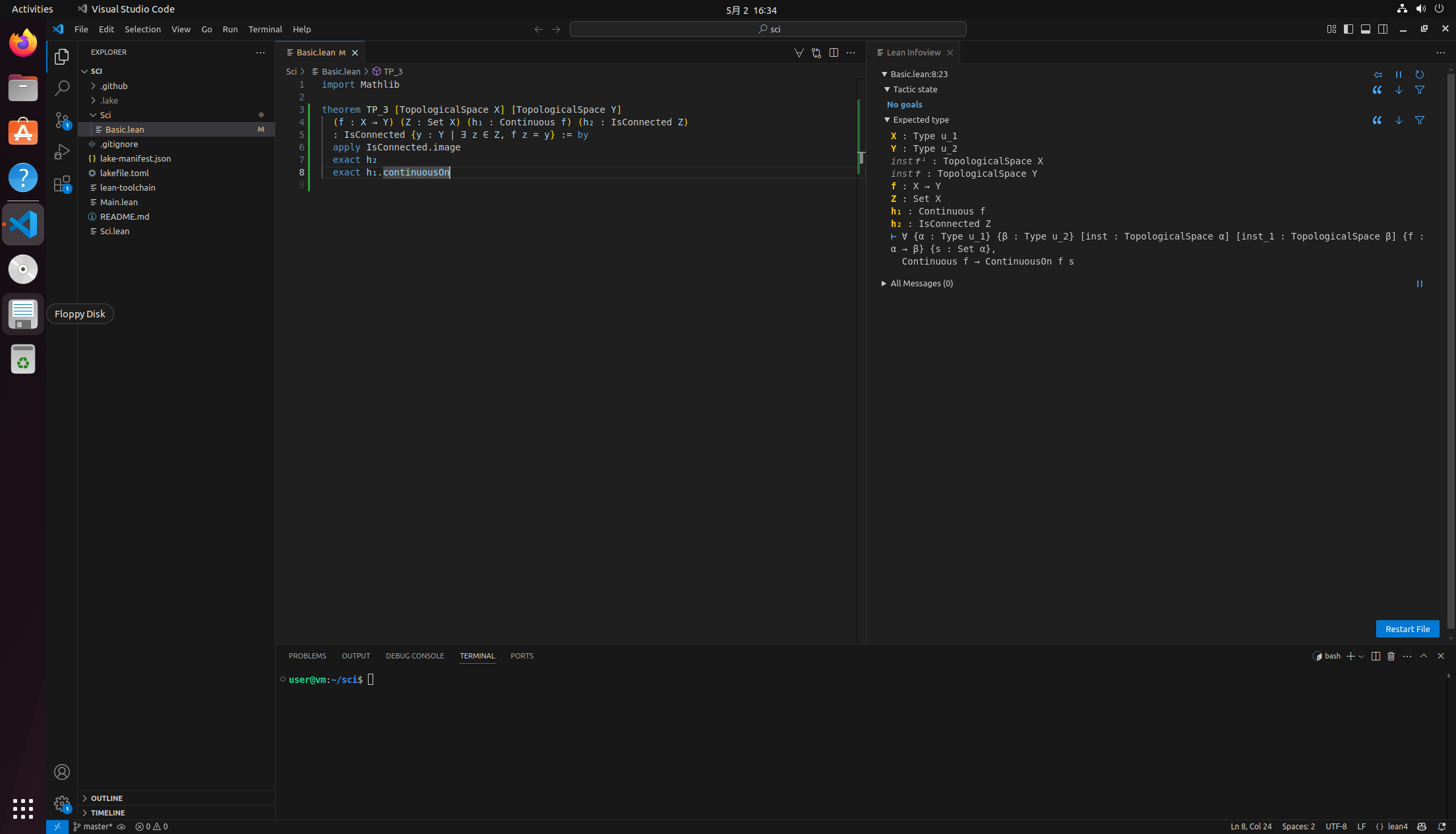}}} &  \multirow{4}{*}{
   \begin{minipage}{.35\textwidth}
   {\ttfamily
    theorem TP\_3 \\
    \hspace*{0.5em} [TopologicalSpace X] \\
    \hspace*{0.5em} [TopologicalSpace Y] \\
    \hspace*{0.5em} (f : X -> Y) \\
    \hspace*{0.5em} (Z : Set X) \\
    \hspace*{0.5em} (h\textsubscript{1} : Continuous f) \\
    \hspace*{0.5em} (h\textsubscript{2} : IsConnected Z) \\
    \hspace*{0.5em} : IsConnected \{y : Y | \\
    \hspace*{1.5em} \ttfamily$\exists$ z \ttfamily$\in$ Z,\ f z = y\} \\
    \hspace*{0.5em} := by sorry
    }
   \end{minipage}} 
    & \footnotesize\texttt{\ \ \{ }\\ 
    &   & \footnotesize\texttt{\ \ \ \ "type": "placeholder"}\\ 
    &   & \footnotesize\texttt{\ \ \} }\\ 
    &   &   \\
    &   &   \\
    &   &   \\
    &   &   \\
    &   &   \\
    &   &   \\
    &   &   \\
  \bottomrule
\end{tabular}
}
% \vspace{-1em}
\vspace{-5pt}
\label{tab:ext_evaluation_examples}
\end{table}

\subsection{Human Performance}
\label{app:task_human_results}

In our main experiments, as reflected in Table~\ref{tab:exp_main}, we recruit college-level students to establish normal human performance on \ours benchmark.
Before attempting the tasks, participants are required to familiarize themselves with foundational knowledge of the relevant scientific disciplines and study the provided operational manuals. They were then given instructions, as shown in Instruction~\ref{fig:prompt_humans}, to complete the assigned tasks. Participants were compensated at a rate of \$10 per hour for their involvement.

The \ours environment and scientific software used do not record any personal information, and all participants provide informed consent. The experiment does not involve surveys, interviews, or any behavioral tracking.

\subsection{Stability Analysis}
\label{app:task_tability}

\begin{figure}[htbp]
    \centering
    \includegraphics[scale=0.312]{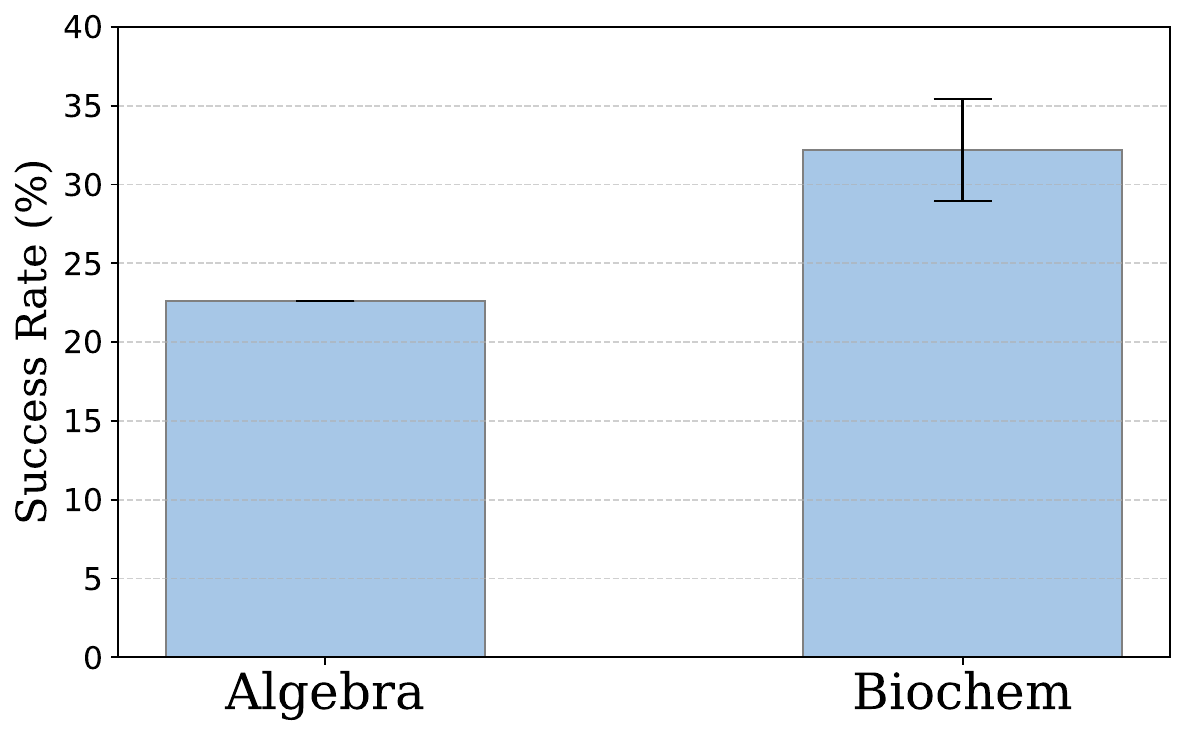}
    \caption{Stability analysis.}
    \label{fig:bench_stability}
\end{figure}

Considering that dynamic environments could potentially lead to experimental instability, 
we conduct an additional set of experiments focusing on consistency. For these, we utilize \gpt under the \atree + screenshot setting, with results and error bars reported in Figure~\ref{fig:bench_stability}.

Across three independent runs,
performance on Algebra tasks remains stable. However, Biochemistry tasks exhibited minor fluctuations in success rates. Upon closer inspection of individual cases, we hypothesize that these variations likely stem from network connectivity issues or transient system lag encountered during task execution.

\subsection{Evaluation Cost}
\label{app:eval_cost}

We use API keys to access proprietary models. 
On average,
a single run on all \ours tasks costs \$64 using \gpt, \$86 using \claude, and \$45 using \gemini.

\section{Details of Experiments}
\label{app:eval}

\subsection{Backbone Models}
\label{app:eval_backbones}

We briefly discuss the backbones we used to build our \cuas.

% \sqs{add size and model version, e.g., GPT-4o-20240806}
% \sqs{@zichen and zhenyu, add a table / itemize describing the evaluated models' features, \textit{e.g.}, evaluated GUI agent models are based on qwenvl chat model, qwen2.5vl with adv grounding capability}

\paragraph{Proprietary Models.} 
% In recent advancements, proprietary models have demonstrated exceptional proficiency in complex task reasoning, significantly pushing the boundaries of artificial intelligence. 
% Furthermore, these models are increasingly exhibiting nascent agentic capabilities, enabling them to interact with and manipulate real-world environments in a more autonomous fashion.
Proprietary models now demonstrate striking capabilities in complex reasoning and are increasingly exhibiting agentic potential for dynamic real-world interaction, prompting a closer look at their diverse forms.
In the experimental section, we accessed the following proprietary models via API keys:

% \begin{itemize}
%     \item{\gpt~\citep{hurst2024gpt}}
%     \item{\claude~\citep{anthropic2024claude}}
%     \item{\gemini~\citep{gemini2024introducing}}
%     \item{\othree~\citep{openai2025o3}}
% \end{itemize}

\begin{itemize}[leftmargin=*,itemsep=2pt]
\item \gpt~\citep{hurst2024gpt}.
\item \gptfive~\citep{openai2025openaigpt5card}.
\item \claude~\citep{anthropic2024claude}.
\item \gemini~\citep{gemini2024introducing}.
\item \geminitwofivepro~\citep{comanici2025gemini}
\item \othree~\citep{openai2025o3}.
\end{itemize}

\paragraph{Open-source Models.} Open-source models are demonstrating remarkable advancements, steadily narrowing the performance gap with proprietary models. 
Crucially, the open-source community recognized the significance of agentic capabilities early on, fostering development in this direction. 
This foresight has translated into exceptional performance, particularly within GUI scenarios where these models now excel on various challenging benchmarks. 
Our evaluation is based on the following open-source models, which are characterized by their advanced grounding capabilities:
\begin{itemize}[leftmargin=*,itemsep=2pt]
    \item{Qwen2.5-VL-72B-Instruct~\citep{bai2025qwen25vl}}: The latest evolution in the Qwen vision-language model family, primarily distinguished by its robust agentic capabilities. 
    It operates directly as a visual agent, proficient in reasoning, dynamically utilizing tools, and executing tasks for computer and phone operation. 
    Complementing its agentic prowess, Qwen2.5-VL-72B-Instruct demonstrates advanced proficiency in detailed visual analysis~(including texts, charts, icons, and layouts within images), comprehension of videos exceeding one hour with event pinpointing, precise object localization with structured coordinate output, and the generation of structured data from documents such as invoices and forms.
    In our experiments, this model is deployed using interconnected clusters of 8 $\times$ A100 80GB GPUs with vLLM~\citep{kwon2023efficient}.
    
    \item{InternVL3-78B~\citep{chen2024expanding}}: An advanced MLLM recognized for its superior overall performance and significantly enhanced multimodal perception and reasoning. 
    A key advancement is its robust agentic functionality, demonstrated through proficient tool usage and GUI agent operations, alongside extended capabilities in areas like industrial image analysis and 3D vision perception. 
    These comprehensive abilities are underpinned by innovations such as a native multimodal pre-training approach, supervised fine-tuning with diverse, high-quality data tailored to these advanced tasks, and mixed preference optimization for refined reasoning. 
    In our experiments, this model is deployed using interconnected clusters of 8 $\times$ A100 80GB GPUs with vLLM.
    
    \item{QvQ-72B-Preview~\citep{qvq72bpreview}}: An experimental research model focused on advancing visual reasoning capabilities. 
    It has achieved compelling performance in complex multidisciplinary understanding and problem-solving, highlighting its specialized strength in sophisticated visual cognitive tasks.
    However, it exhibits some limitations in instruction following, appearing less adept in agent scenarios that require precise action outputs. 
    In our experiments, this model is deployed using interconnected clusters of 8 $\times$ A100 80GB GPUs with vLLM.
\end{itemize}

\paragraph{GUI Action Models.} While foundational models provide impressive general-purpose intelligence, their intrinsic agentic capabilities for nuanced GUI manipulation are still under active exploration, often requiring further specialization. 
Consequently, a prominent line of research involves adapting open-source VLMs by fine-tuning them on extensive, GUI-specific datasets. 
This targeted training methodology yields dedicated action models equipped with significantly enhanced proficiencies for understanding and interacting with GUIs. 
The GUI action models adopted in this paper are as follows:

\begin{itemize}[leftmargin=*,itemsep=2pt]
    \item{OS-Atlas-Pro-7B~\citep{wu2025osatlas}}: A foundational GUI action model that significantly advances open-source VLMs for agentic tasks, excelling in GUI grounding and out-of-distribution scenarios through innovations in modeling and the creation of the largest open-source, cross-platform GUI grounding corpus with over 13 million elements.
    It demonstrates state-of-the-art performance across six diverse benchmarks~(mobile, desktop, web) and verifies the existence of model scaling laws in GUI scenarios.
    In our experiments, this model is deployed using a single A100 80GB GPU with vLLM~\citep{kwon2023efficient}.
    \item{UGround-V1-7B~\citep{gou2025navigating}}: A universal visual grounding model that identifies GUI action elements by pixel coordinates.
    It powers the SeeAct-V framework~\citep{zheng2024seeact}, which enables purely visual GUI perception and pixel-level operations.
    Agents using SeeAct-V with UGround have achieved SOTA results across five distinct benchmarks spanning web, mobile, and desktop evaluations.
    In our experiments, this model is deployed on a single A100 80GB GPU with vLLM.
    \item{UI-TARS-72B-DPO~\citep{qin2025uitars}}: An end-to-end native GUI agent that uniquely perceives screenshots as its sole input to perform human-like keyboard and mouse interactions, outperforming prevailing agent frameworks that depend on heavily wrapped commercial models with expert-crafted prompts. 
    It has established state-of-the-art performance across more than ten GUI agent benchmarks. 
    This advanced capability stems from key innovations including enhanced perception, unified action modeling, System-2 reasoning, iterative training with reflective online traces, and a final Direct Preference Optimization~(DPO) phase, which refines its ability to make precise, context-aware decisions.
    In our experiments, UI-TARS-72B-DPO utilizes vLLM for inference and is deployed on interconnected clusters of 8 $\times$ A100 80GB GPUs.
    \item{GUI-Actor-7B~\citep{wu2025gui}}: A recently proposed GUI grounding model that introduces a novel coordinate-free visual grounding approach. It utilizes an action head to direct the special token <ACTOR> to the target screenshot patches for localization.
    It claims to surpass the text-based coordinate prediction baseline and demonstrates better generalization in out-of-distribution (OOD) scenarios.
    In our experiments, we used the 7B version of GUI-Actor based on the Qwen2.5-VL backbone.
\end{itemize}

% \sqs{interconnected clusters of 8 $\times$ A100 80GB GPUs}.

% \sqs{@zichen, how about using para instead of subsub? I put some model names with citations below}
% \sqs{GPU nodes and inference details, temperature, top k}
% Qwen2.5-VL-72B QvQ-72B-Preview, and InternVL3-78B 
% which are VLMs specifically tailored for GUI tasks, 
% including 

\subsection{Evaluation Settings - Main Experiments}
\label{app:eval_settings}

We adhered to common prompt engineering strategies from previous works~\citep{sun2024osgenesis,zhou2024webarena,zhang2024autogui} for the agents under evaluation. For each domain, the agent interacts with the environment under the guidance of a meta-prompt, which includes information about the software being operated, executable special actions, and related details. When taking actions, the agent generates outputs in the ReAct style~\citep{yao2023react}, with its step-by-step thoughts recorded in the interaction history.
% Prompts we employed are listed in Appendix~\ref{app:prompts}.

Throughout the evaluation,
we set the \texttt{temperature parameter} to 0.5, \texttt{top\_p} to 0.9, and \texttt{max\_tokens} to 1500.
We list some prompt examples in Prompt~\ref{fig:prompt_chimerax_ss}, Prompt~\ref{fig:prompt_celestia_ss}, Prompt~\ref{fig:prompt_chimerax_som} and Prompt~\ref{fig:prompt_celestia_som}.

\subsection{Evaluation Settings - Analysis}
\label{app:eval_settings_seperate}

In experiments with interleaved planning and action,
we first address inconsistencies in coordinate outputs from different GUI action models. While InternVL3-78B~\citep{chen2024expanding} outputs coordinates on a \texttt{[0, 1]} scale,
models such as OS-Atlas, UI-TARS, and UGround use a \texttt{[0, 1000]} scale. 
To ensure uniformity, we normalized all coordinate outputs to a \texttt{[0, 1]} scale prior to execution.

This part of the experiments employs a two-stage process:
First, the planner model receives the current observation (obs) and task instruction to generate a high-level plan or a specific action.
If the planner outputted a directly executable primitive action (\textit{e.g.}, a non-GUI system-level command or a special control token like \texttt{DONE}),
that action will be performed immediately, and the action model was not invoked for that step.
Otherwise, the grounding model received the current observation and the plan (or sub-task) from the planner. Its role was to output low-level executable instructions. 
If the grounding model generate \texttt{pyautogui} actions directly, these commands were executed. For models outputting in their specific native formats, we implement custom parsers to translate these into \texttt{pyautogui} actions: 
for UGround and UI-TARS, all coordinate-based outputs were interpreted as \texttt{click}, whereas for OS-Atlas, its outputs were parsed to differentiate between \texttt{click}, \texttt{type}, and \texttt{scroll} based on its defined schema.

We list some prompt examples in Prompt~\ref{fig:prompt_atlas}, Prompt~\ref{fig:prompt_uground}, Prompt~\ref{fig:prompt_qwen} and Prompt~\ref{fig:prompt_tars}.

% \paragraph{Backbones.}

% 讲一下测试的时候怎么处理
% internvl3: screenshot 通常用相对坐标，a11y_tree 基本用绝对坐标，screenshot + a11y_tree 随机使用坐标，更偏好相对坐标

% 在interleaved planning and action的实验中，
% 不同GUI action model的坐标输出形式不同，InternVL-3的的scale为[0 - 1] OS-Atlas、UI-TARS、UGround的scale为[0 - 1000]。我们统一将scale转化为[0-1]，

% 说一下两个模型是怎么结合的，以及answer这个动作，哪个模型是谁做的
% 向第一个模型提供 obs，要求输出一个计划
% - 第一个模型输出 primitves，则直接执行，该步不调用其他模型
% 向第二个模型提供 obs 和计划，要求输出特定指令
% - 直接输出 pyautogui，直接执行
% - 输出 grounding model 的特殊输出，人工解析转化为 pyautogui 代码
%   - uground 和 utars：所有坐标解释为 click
%   - atlas：解析 click、type 和 scroll

% \sqs{add more details about seeact setting experiments, planner and grounding model}

% 包括各种坐标问题
\section{Extended Analysis}
\label{app:analysis}

% As shown in Figure~\ref{fig:extended_analysis_of_gui_cli}.

\subsection{Interfaces}
\label{app:analysis_interface}

In Section~\ref{sec:analysis}, we analyze the performance difference between Vision-Only and Hybrid Interface settings under the \atree + screenshot.
Here, we present empirical results under the other three observation settings.

\begin{figure}[ht]
    \centering
    % \vspace{-0.8cm}
    \includegraphics[scale=0.312]{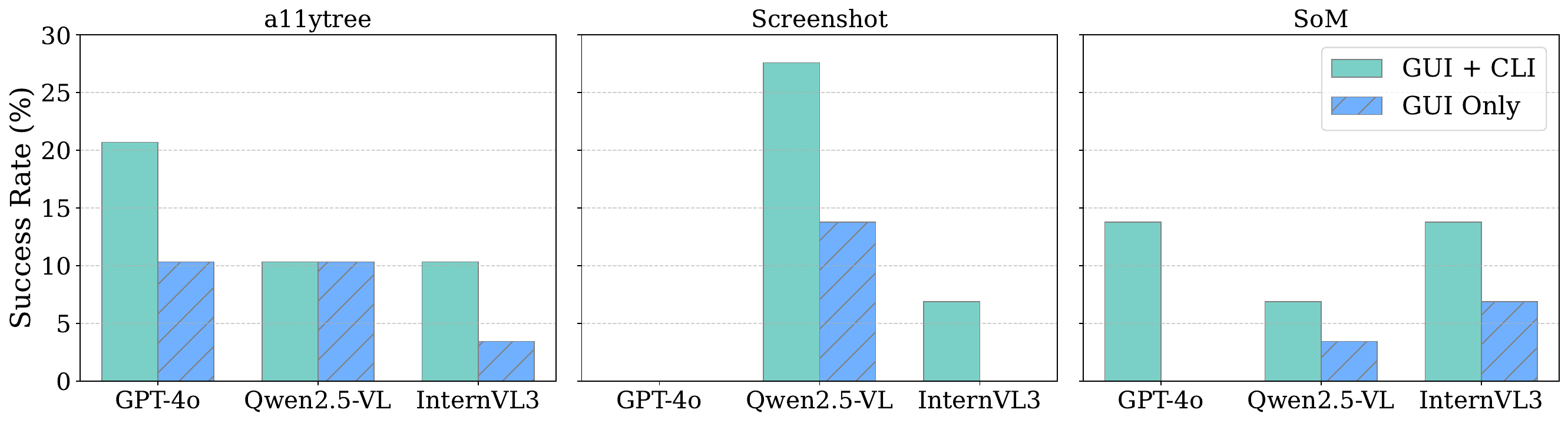}
    \caption{Extended analysis of Vision-Only vs. Hybrid Interface.}
    \label{fig:extended_analysis_of_gui_cli}
\end{figure}

As shown in Figure~\ref{fig:extended_analysis_of_gui_cli}, 
the hybrid GUI + CLI setting consistently achieves performance that is comparable to or better than the GUI-Only setting across all scenarios.
Interestingly, 
while \gpt achieves state-of-the-art performance under other observation settings,
it exhibits very weak action capabilities when using screenshot setting, indicating the reliance on structured observations for effective reasoning and planning.

%%%%%%%%%%%%%%%%%%%%%%%%%%%
\subsection{Interactive Environments}
% \subsection{Textual vs. Interactive Environments.}
ATP represents one of the most logic-intensive tasks for agents and has been traditionally studied in textual settings in prior works (\textit{e.g.}, plain text or bash terminal).
% We extend ATP to live OS environments in \ours and further compare agents’ performance under textual and interactive settings, the latter providing features such as syntax highlighting, autocompletion, type inference, and more in a live VSCode environment.

\begin{figure}[ht]
    \centering
    % \vspace{-0.8cm}
    \includegraphics[width=0.45\linewidth]{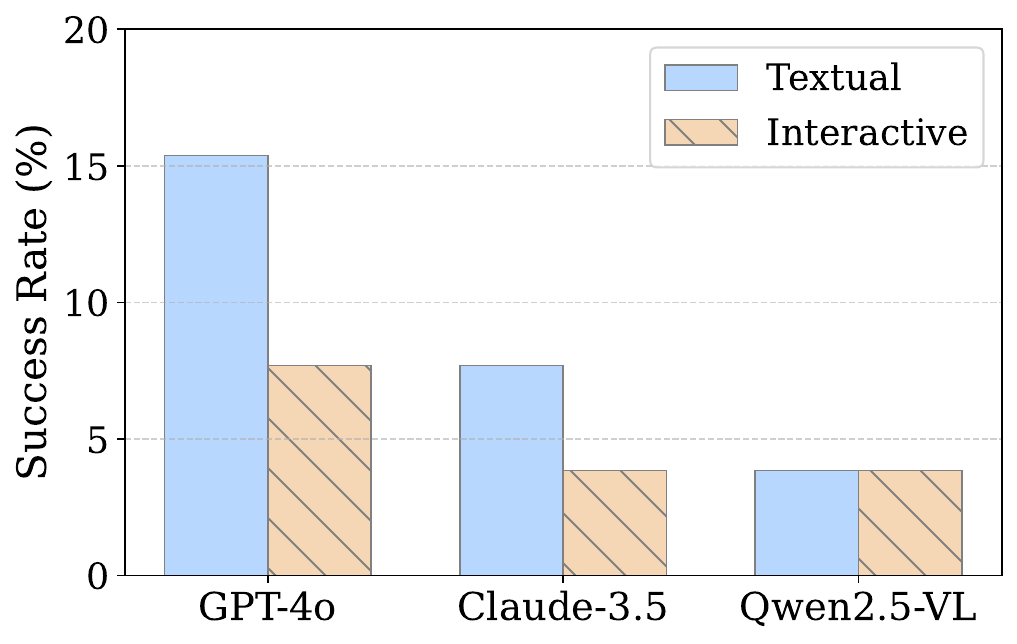}
    \caption{Textual v.s. Interactive }
    \label{fig:atp_agentic_textual}
\end{figure}

% \begin{wrapfigure}{r}{6cm}
% \vspace{-0.75em}
%     \centering
%     \includegraphics[width=0.95\linewidth]{Figures/analysis/scienceboard_atp.pdf}
%     \vspace{-0.75em}
%     \caption{Textual v.s. Interactive }
%     % by each method.}
%     \vspace{-1em}
%     \label{fig:atp_agentic_textual}
% \end{wrapfigure}
We extend ATP to live OS in \ours and further compare agents’ performance under textual and interactive settings. 
The latter, similar to environments commonly used by humans, provides a live VSCode interface with features such as syntax highlighting, autocompletion, type inference, and other functionalities.
As shown in Figure~\ref{fig:atp_agentic_textual}, in the textual setting, the agent applies heuristic strategies (\textit{e.g.}, Monte Carlo search) to make predictions over the proof tree without interacting with the environment.
In contrast, in the interactive setting, the agent must autonomously decide which \textsc{ProofState} to proceed with. 
Moreover, the agent is also required to localize the relevant code segments within the interface.
Completing formal methods tasks becomes substantially more challenging in realistic environments, which significantly increases the cognitive complexity.
%%%%%%%%%%%%%%%%%%%%%%%%%%%

\subsection{Difficulty Analysis}
\label{app:analysis_difficulty}

\begin{figure}[ht]
    \centering
    % \vspace{-0.8cm}
    \includegraphics[scale=0.312]{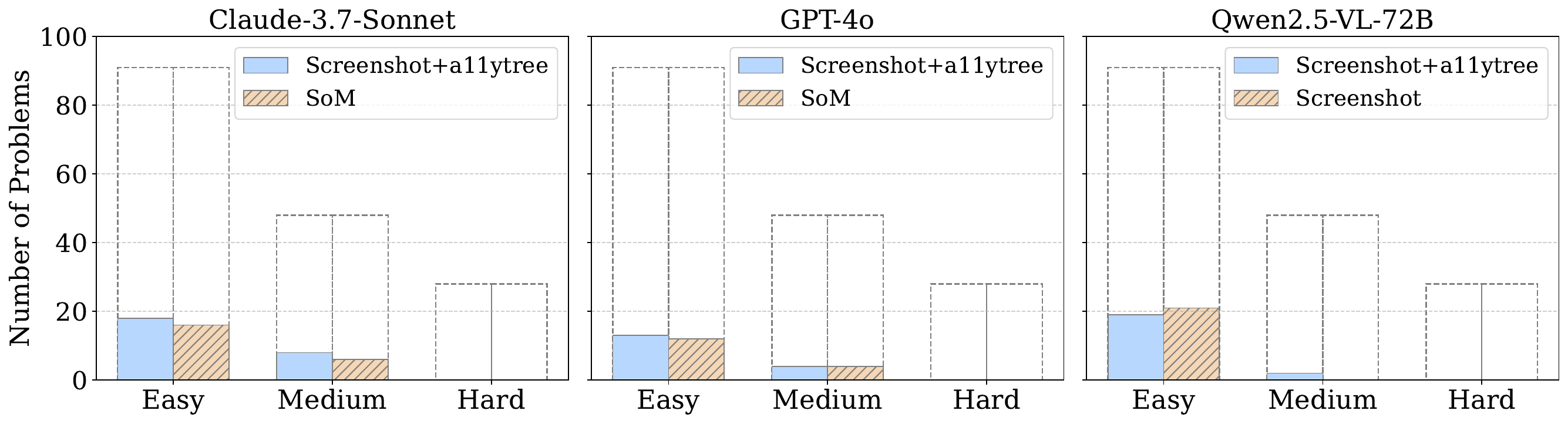}
    \caption{Comparative analysis of task difficulty solve rates.}
    \label{fig:extended_analysis_of_difficulty}
\end{figure}

We further analyze the success rates of \cuas on the \ours benchmark across different task difficulty levels. 
We employ \claude, \gpt, and Qwen2.5-VL, with results presented in Figure~\ref{fig:extended_analysis_of_difficulty}. 

% as shown in Figure~\ref{fig:extended_analysis_of_difficulty}.

The findings indicate that solvable tasks are primarily concentrated among a subset of ``Easy'' problems and a few ``Medium'' tasks.
All ``hard'' tasks, which involve complex computations, cross-application workflows, or long-horizon planning, could not be completed by any of the evaluated agents.

\subsection{Compute Scaling}
\label{app:compute_scaling}

During the evaluation of frontier models, a natural question arises: can scaling inference-time compute (test-time scaling) resolve the performance bottlenecks observed in complex scientific workflows? To investigate whether extended generation limits and native reasoning modes enable agents to better navigate these dynamic environments, we conducted an extended analysis focusing on three domains.

\begin{table}[h]
\centering
\caption{Success rates of varying \texttt{max\_tokens} under the Screenshot observation setting.}
\label{tab:max_tokens_ablation}
\resizebox{0.55\textwidth}{!}{
\begin{tabular}{lccc}
\toprule
{\texttt{max\_tokens}} & {Algebra} & {Biochemistry} & {GIS} \\
\midrule
1500 & 41.90\% & 62.10\% & 11.80\% \\
2000 & 41.90\% & 58.62\% & 11.80\% \\
2500 & 41.90\% & 65.52\% & 14.70\% \\
\bottomrule
\end{tabular}
}
\end{table}

Further, we evaluated the performance of \texttt{GPT-5} utilizing its native reasoning mode with varying levels of test-time reasoning effort (\texttt{medium} and \texttt{high}). As detailed in Table~\ref{tab:reasoning_effort}, elevating the reasoning effort improves performance across several settings. Notably, under the \textit{Screenshot + \atree} setting, shifting from medium to high reasoning effort slightly increases the success rate in Biochemistry from 68.96\% to 72.41\%, and in GIS from 14.70\% to 17.64\%.

\begin{table}[h]
\centering
\caption{Performance of \texttt{GPT-5} under different native reasoning efforts across two observation settings.}
\label{tab:reasoning_effort}
\resizebox{0.85\textwidth}{!}{
\begin{tabular}{llccc}
\toprule
\textbf{Observation} & \textbf{Reasoning Effort} & \textbf{Algebra} & \textbf{Biochemistry} & \textbf{GIS} \\
\midrule
\multirow{2}{*}{Screenshot} 
& Medium & 41.90\% & 65.52\% & 14.70\% \\
& High & 45.16\% & 65.52\% & 14.70\% \\
\midrule
\multirow{2}{*}{Screenshot + \atree} 
& Medium & 48.39\% & 68.96\% & 14.70\% \\
& High & 51.61\% & 72.41\% & 17.64\% \\
\bottomrule
\end{tabular}
}
\end{table}

These results empirically demonstrate that while scaling test-time compute and leveraging native reasoning models provide measurable benefits, they do not fundamentally overcome the core challenges of our benchmark. The bottlenecks likely stem from the agent's general agentic capabilities, \textit{i.e.}, the ability to accurately perceive dense, domain-specific UI elements and translate high-level scientific plans into precise, executable actions, rather than solely a deficiency in deep cognitive reasoning.

\subsection{Failure Analysis}
\label{app:analysis_failure}

To further investigate the reasons why \cuas fail when planning or taking actions on scientific tasks, here we include and discuss several typical examples of such errors.

\begin{figure}[ht]
    \centering
    % \vspace{-0.8cm}
    \includegraphics[width=0.98\textwidth]{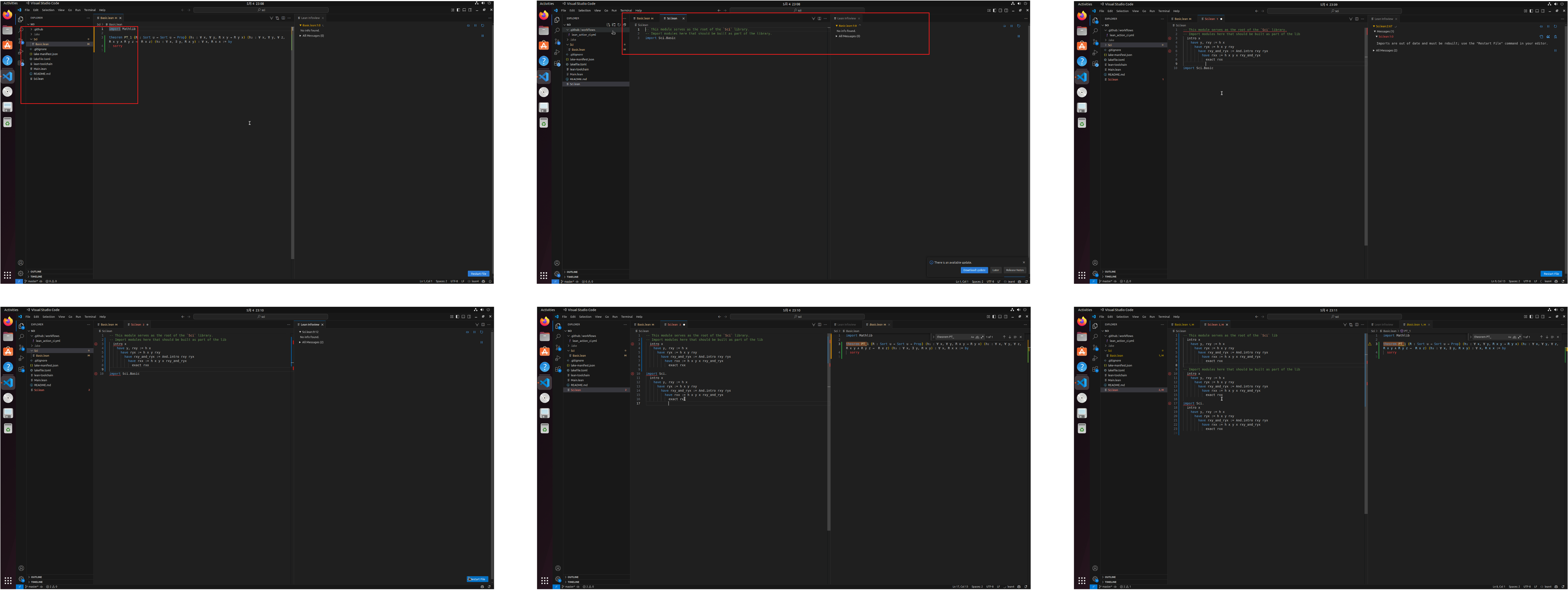}
    \caption{Use wrong file.}
    \label{fig:failure_analysis_wrong_file}
\end{figure}

\paragraph{Opening the Wrong File.}
This error is frequently caused by grounding issues. The agent initially clicks on an incorrect file and then attempts to perform subsequent actions, such as inputting data, within that wrong file. This often leads to the agent repeatedly making the same mistake or getting stuck in an unproductive loop. A typical case is shown in Figure~\ref{fig:failure_analysis_wrong_file}.

% 一开始点到了另一个文件，然后在错误的文件里输入, 错误之后还一直重复之前的错误

\begin{figure}[ht]
    \centering
    % \vspace{-0.8cm}
    \includegraphics[width=0.98\textwidth]{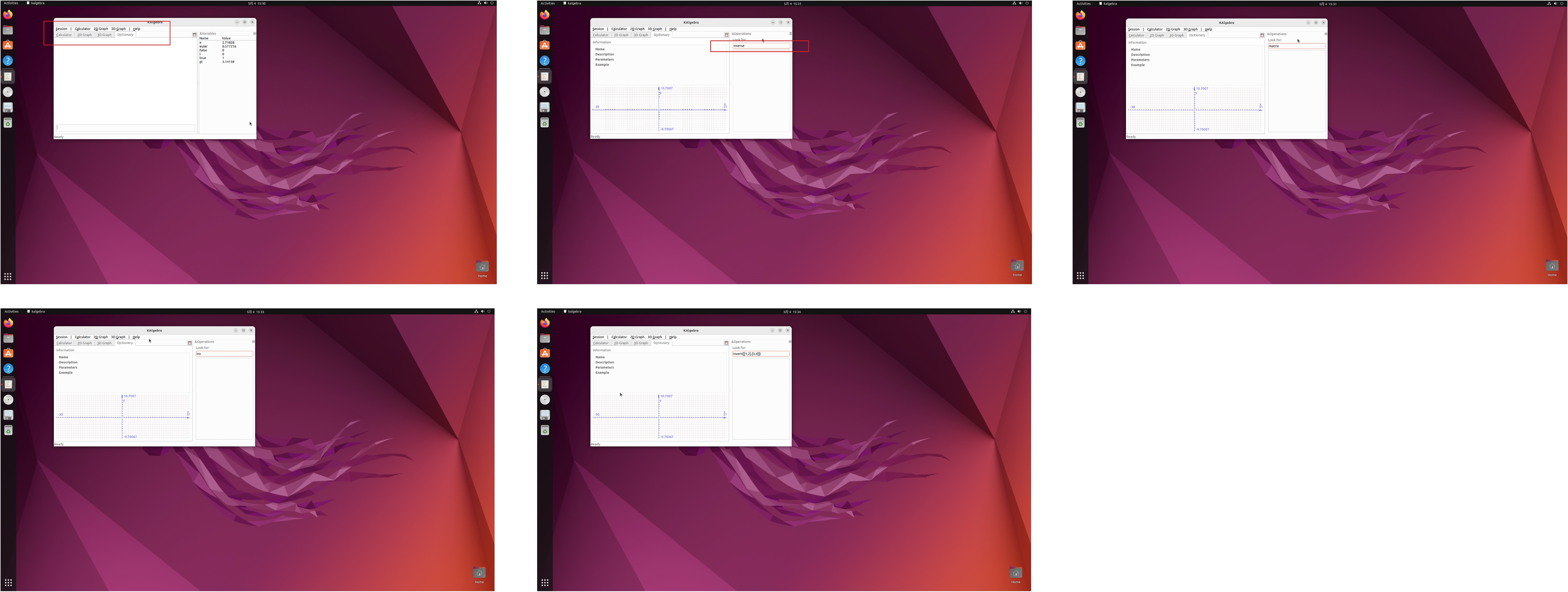}
    \caption{Function invocation error.}
    \label{fig:failure_analysis_wrong_invoke}
\end{figure}

\paragraph{Inability to Invoke the Correct Function.}
In some instances, agents need to identify and use a specific function within a software application but attempt to do so by directly typing an assumed function name into a search bar or command input.
If the exact function name is unknown or guessed incorrectly, a more robust strategy would be to browse available menus or function lists. Instead, agents may incorrectly assume knowledge of the function name and attempt to look up its usage, leading to failure. 
A typical example of this behavior is presented in Figure~\ref{fig:failure_analysis_wrong_invoke}.

% 应该要找出想要的函数，但是直接在搜索栏里输入。如果不知道函数名，应该在列表里找，而不是假装自己已经知道了，然后把它搜出来查用法，KAlg 的大部分题目都没有有效操作

\begin{figure}[ht]
    \centering
    % \vspace{-0.8cm}
    \includegraphics[width=0.98\textwidth]{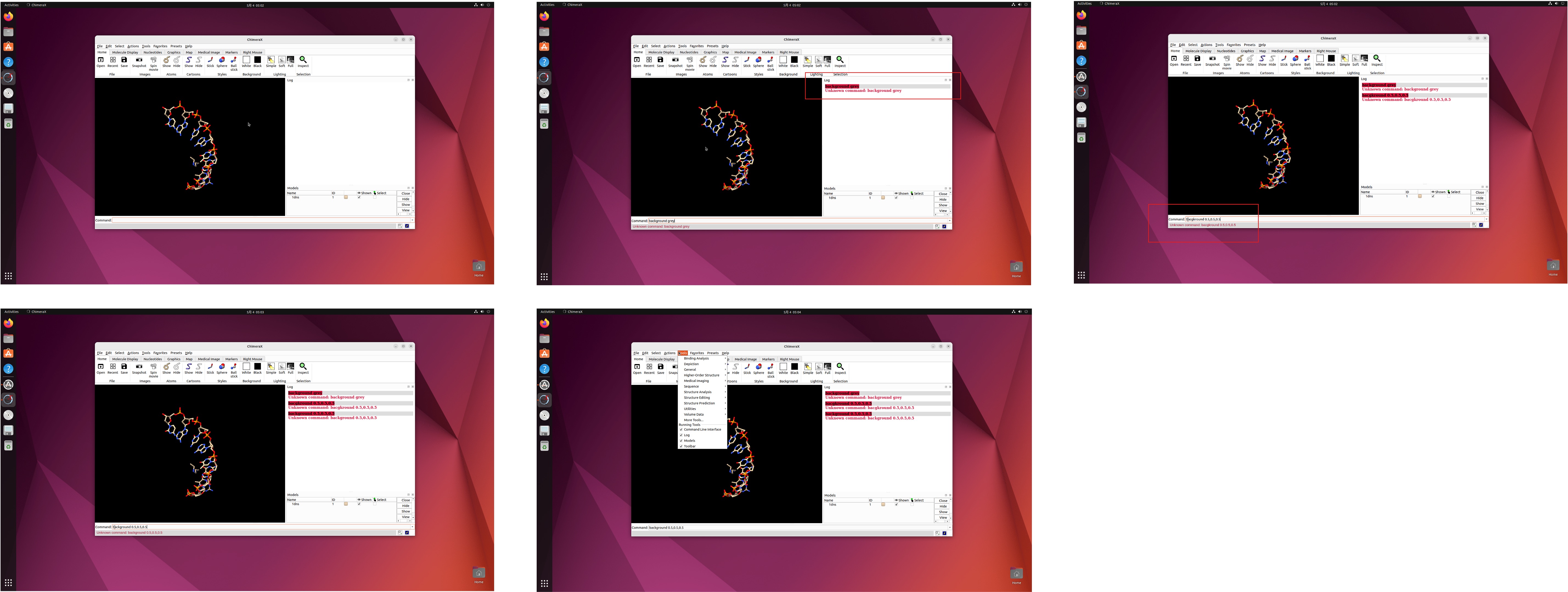}
    \caption{CLI code error.}
    \label{fig:failure_analysis_wrong_code}
\end{figure}

\paragraph{Incorrect CLI Code.}
Failures also occur when agents formulate CLI commands incorrectly. This can involve syntax errors, wrong command names, or incorrect parameters. Notably, in some of these failed CLI attempts, the intended task could have been accomplished more straightforwardly by interacting with a corresponding button or element in the GUI.
A typical example is shown in Figure~\ref{fig:failure_analysis_wrong_code}.

% 命令写错了，应该是 set bgColor white；也可以直接点一下上面的按钮，可惜非要用那个 cli

% \sqs{slow thinking llm problems}

% \sqs{difficulty analysis}

% claude ss+a11y 13+5+0
% som 15+7+0

% Easy (steps=5)	91
% Medium (steps=10)	48
% Hard (steps=15)	28

% \sqs{failure analysis}
\section{Prompts}
\label{app:prompts}

The prompt examples we used in \ours are listed below. 
\newpage

\begin{figure*}
    \centering
    \setlength{\fboxrule}{0.85pt}
    \fbox{ \footnotesize
        \parbox{\textwidth}{\texttt{\textbf{Agentic Prompt - ChimeraX with screenshot}\\
        \\
You are an agent which follow my instruction and perform desktop computer tasks as instructed. \\
You have good knowledge of ChimeraX, a molecular visualization software; and assume your code will run on a computer controlling the mouse and keyboard. \\
For each step, you will get an observation of the desktop by an accessibility tree, which is based on AT-SPI library, and you will predict actions of the next step based on that. \\
 \\
You are required to use `pyautogui` to perform the action grounded to the observation, but DO NOT use the `pyautogui.locateCenterOnScreen` function to locate the element you want to operate with since we have no image of the element you want to operate with. DO NOT USE `pyautogui.screenshot()` to make screenshot. \\
You ONLY need to return the code inside a code block, like this: \\
``` \\
\# your code here \\
``` \\
Return one line or multiple lines of python code to perform the action each time, and be time efficient. When predicting multiple lines of code, make some small sleep like `time.sleep(0.5);` interval so that the machine could take breaks. Each time you need to predict a complete code, and no variables or function can be shared from history. \\
 \\
Specially, it is also allowed to return the following special code: \\
When you think the task is done, return ```DONE```; \\
When you think the task can not be done, return ```FAIL```. Don't easily say ```FAIL```; try your best to do the task; \\
When you think you have to wait for some time, return ```WAIT``` or ```WAIT n```, in which n defaults to 5(s); \\
When you are asked to submit an answer, return ```ANS s``` without quotation marks surrounding s, and use `FAIL` if there is no answer to the question. \\
 \\
My computer's password is 'password', feel free to use it when you need sudo rights. \\
DO NOT introduce any unrelated models or easily close existing models, otherwise the task might be evaluated as FAILED. \\
DO NOT close the current ChimeraX session, or every effort you made will be in vain. \\
NEVER try to reopen the command line interface in ChimeraX if it is hidden, because it has been deactivated and cannot do anything. But you are welcome to use it once it is presented. \\
 \\
First give the current observation and previous things we did a short reflection, then RETURN ME THE CODE OR SPECIAL CODE I ASKED FOR. NEVER EVER RETURN ME ANYTHING ELSE. \\
You are asked to complete the following task: Fetch 2OLX from PDB in ChimeraX.
        }
    }}
    \captionsetup{labelformat=default, name=Prompt}
    \caption{Prompts for ChimeraX with screenshot}
    \label{fig:prompt_chimerax_ss}
\end{figure*}
\begin{figure*}
    \centering
    \setlength{\fboxrule}{0.85pt}
    \fbox{ \footnotesize
        \parbox{\textwidth}{\texttt{\textbf{Agentic Prompt - Celestia with screenshot}\\
        \\
You are an agent which follow my instruction and perform desktop computer tasks as instructed. \\
You have good knowledge of Celestia, a three-dimension space simulator; and assume your code will run on a computer controlling the mouse and keyboard. \\
For each step, you will get an observation of the desktop by a screenshot, and you will predict actions of the next step based on that. \\
 \\
You are required to use `pyautogui` to perform the action grounded to the observation, but DO NOT use the `pyautogui.locateCenterOnScreen` function to locate the element you want to operate with since we have no image of the element you want to operate with. DO NOT USE `pyautogui.screenshot()` to make screenshot. \\
You ONLY need to return the code inside a code block, like this: \\
``` \\
\# your code here \\
``` \\
Return one line or multiple lines of python code to perform the action each time, and be time efficient. When predicting multiple lines of code, make some small sleep like `time.sleep(0.5);` interval so that the machine could take breaks. Each time you need to predict a complete code, and no variables or function can be shared from history. \\
 \\
Specially, it is also allowed to return the following special code: \\
When you think the task is done, return ```DONE```; \\
When you think the task can not be done, return ```FAIL```. Don't easily say ```FAIL```; try your best to do the task; \\
When you think you have to wait for some time, return ```WAIT``` or ```WAIT n```, in which n defaults to 5(s); \\
When you are asked to submit an answer, return ```ANS s``` without quotation marks surrounding s, and use `FAIL` if there is no answer to the question. \\
 \\
My computer's password is 'password', feel free to use it when you need sudo rights. \\
The criterion for a celestial body to be displayed on the screen is that the object's center is within the window range and is not blocked by others. \\
 \\
First give the current observation and previous things we did a short reflection, then RETURN ME THE CODE OR SPECIAL CODE I ASKED FOR. NEVER EVER RETURN ME ANYTHING ELSE. \\
You are asked to complete the following task: Set the Julian date to 2400000 in Celestia.
        }
    }}
    \captionsetup{labelformat=default, name=Prompt}
    \caption{Prompts for Celestia with screenshot}
    \label{fig:prompt_celestia_ss}
\end{figure*}
\begin{figure*}
    \centering
    \setlength{\fboxrule}{0.85pt}
    \fbox{ \footnotesize
        \parbox{\textwidth}{\texttt{\textbf{Agentic Prompt - ChimeraX with set-of-marks}\\
        \\
You are an agent which follow my instruction and perform desktop computer tasks as instructed. \\
You have good knowledge of ChimeraX, a molecular visualization software; and assume your code will run on a computer controlling the mouse and keyboard. \\
For each step, you will get an observation of the desktop by 1) an accessibility tree, which is based on AT-SPI library; and 2) a screenshot with interact-able elements marked with numerical tags, and you will predict actions of the next step based on that. \\
 \\
You are required to use `pyautogui` to perform the action grounded to the observation, but DO NOT use the `pyautogui.locateCenterOnScreen` function to locate the element you want to operate with since we have no image of the element you want to operate with. DO NOT USE `pyautogui.screenshot()` to make screenshot. \\
You ONLY need to return the code inside a code block, like this: \\
``` \\
\# your code here \\
``` \\
Return one line or multiple lines of python code to perform the action each time, and be time efficient. When predicting multiple lines of code, make some small sleep like `time.sleep(0.5);` interval so that the machine could take breaks. Each time you need to predict a complete code, and no variables or function can be shared from history. \\
 \\
You can replace x, y in the code with the tag of elements you want to operate with, such as: \\
``` \\
pyautogui.moveTo(tag\_3) \\
pyautogui.click(tag\_2) \\
pyautogui.dragTo(tag\_1, button='left') \\
``` \\
When you think you can directly output precise x and y coordinates or there is no tag on which you want to interact, you can also use them directly; but you should be careful to ensure the correct of coordinates. \\
 \\
Specially, it is also allowed to return the following special code: \\
When you think the task is done, return ```DONE```; \\
When you think the task can not be done, return ```FAIL```. Don't easily say ```FAIL```; try your best to do the task; \\
When you think you have to wait for some time, return ```WAIT``` or ```WAIT n```, in which n defaults to 5(s); \\
When you are asked to submit an answer, return ```ANS s``` without quotation marks surrounding s, and use `FAIL` if there is no answer to the question. \\
 \\
My computer's password is 'password', feel free to use it when you need sudo rights. \\
DO NOT introduce any unrelated models or easily close existing models, otherwise the task might be evaluated as FAILED. \\
DO NOT close the current ChimeraX session, or every effort you made will be in vain. \\
NEVER try to reopen the command line interface in ChimeraX if it is hidden, because it has been deactivated and cannot do anything. But you are welcome to use it once it is presented. \\
 \\
First give the current observation and previous things we did a short reflection, then RETURN ME THE CODE OR SPECIAL CODE I ASKED FOR. NEVER EVER RETURN ME ANYTHING ELSE. \\
You are asked to complete the following task: Fetch 2OLX from PDB in ChimeraX. \\
        }
    }}
    \captionsetup{labelformat=default, name=Prompt}
    \caption{Prompts for ChimeraX with Set-of-Marks}
    \label{fig:prompt_chimerax_som}
\end{figure*}
\begin{figure*}
    \centering
    \setlength{\fboxrule}{0.85pt}
    \fbox{ \footnotesize
        \parbox{\textwidth}{\texttt{\textbf{Agentic Prompt - Celestia with set-of-marks}\\
        \\
You are an agent which follow my instruction and perform desktop computer tasks as instructed. \\
You have good knowledge of Celestia, a three-dimension space simulator; and assume your code will run on a computer controlling the mouse and keyboard. \\
For each step, you will get an observation of the desktop by 1) an accessibility tree, which is based on AT-SPI library; and 2) a screenshot with interact-able elements marked with numerical tags, and you will predict actions of the next step based on that. \\
 \\
You are required to use `pyautogui` to perform the action grounded to the observation, but DO NOT use the `pyautogui.locateCenterOnScreen` function to locate the element you want to operate with since we have no image of the element you want to operate with. DO NOT USE `pyautogui.screenshot()` to make screenshot. \\
You ONLY need to return the code inside a code block, like this: \\
``` \\
\# your code here \\
``` \\
Return one line or multiple lines of python code to perform the action each time, and be time efficient. When predicting multiple lines of code, make some small sleep like `time.sleep(0.5);` interval so that the machine could take breaks. Each time you need to predict a complete code, and no variables or function can be shared from history. \\
 \\
You can replace x, y in the code with the tag of elements you want to operate with, such as: \\
``` \\
pyautogui.moveTo(tag\_3) \\
pyautogui.click(tag\_2) \\
pyautogui.dragTo(tag\_1, button='left') \\
``` \\
When you think you can directly output precise x and y coordinates or there is no tag on which you want to interact, you can also use them directly; but you should be careful to ensure the correct of coordinates. \\
 \\
Specially, it is also allowed to return the following special code: \\
When you think the task is done, return ```DONE```; \\
When you think the task can not be done, return ```FAIL```. Don't easily say ```FAIL```; try your best to do the task; \\
When you think you have to wait for some time, return ```WAIT``` or ```WAIT n```, in which n defaults to 5(s); \\
When you are asked to submit an answer, return ```ANS s``` without quotation marks surrounding s, and use `FAIL` if there is no answer to the question. \\
 \\
My computer's password is 'password', feel free to use it when you need sudo rights. \\
The criterion for a celestial body to be displayed on the screen is that the object's center is within the window range and is not blocked by others. \\
 \\
First give the current observation and previous things we did a short reflection, then RETURN ME THE CODE OR SPECIAL CODE I ASKED FOR. NEVER EVER RETURN ME ANYTHING ELSE. \\
You are asked to complete the following task: Set the Julian date to 2400000 in Celestia.
        }
    }}
    \captionsetup{labelformat=default, name=Prompt}
    \caption{Prompts for Celestia with Set-of-Marks}
    \label{fig:prompt_celestia_som}
\end{figure*}
% \begin{figure*}
%     \centering
%     \setlength{\fboxrule}{0.85pt}
%     \fbox{ \footnotesize
%         \parbox{\textwidth}{\texttt{\textbf{Human Instructions}\\
%         \\
% You are required to finish the given tasks manually to provide sample data of human accuracy.  \\
% First, please start up the evaluation script with debug option ON and headless option OFF. Then, wait for the environment to be initialized and perform your actions when you receive corresponding logs from stdout. Press ENTER after you finish operating and the script will evaluate your result submitted automatically.  \\
% Attention:  \\
% 1. If you need to finish the task with primitives other than TIMEOUT, please input directly into stdin;  \\
% 2. You can search for documents or manuals if you encounter domain-specidfic knowledge you are not familiar with;  \\
% 3. Make sure that the number of your steps is less than expected. To be more precise, a pop-up without possibility to predict its position should be split into different steps.  \\
%         }
%     }}
%     % \captionsetup{labelformat=default, name=Prompt}
%     % \caption{Instruction for humans.}
%     \caption{{Instruction 1:} Instruction for humans.}
%     \label{fig:prompt_humans}
% \end{figure*}

\begin{figure*}
    \centering
    \setlength{\fboxrule}{0.85pt}
    \fbox{ \footnotesize
        \parbox{\textwidth}{\texttt{\textbf{Human Instructions}\\
        \\
You are required to finish the given tasks manually to provide sample data of human accuracy.  \\
First, please start up the evaluation script with debug option ON and headless option OFF. Then, wait for the environment to be initialized and perform your actions when you receive corresponding logs from stdout. Press ENTER after you finish operating and the script will evaluate your result submitted automatically.  \\
Attention:  \\
1. If you need to finish the task with primitives other than TIMEOUT, please input directly into stdin;  \\
2. You can search for documents or manuals if you encounter domain-specific knowledge you are not familiar with;  \\
3. Make sure that the number of your steps is less than expected. To be more precise, a popup without possibility to predict its position should be split into different steps.  \\
        }
    }}
    \refstepcounter{instruction}
    \caption*{{Instruction \theinstruction:} Instruction for humans.}
    \label{fig:prompt_humans}
\end{figure*}

\begin{figure*}
    \centering
    \setlength{\fboxrule}{0.85pt}
    \fbox{ \footnotesize
        \parbox{\textwidth}{\texttt{\textbf{Agentic Prompt - OS-Atlas}\\
        \\
You are an agent which follow my instruction and perform desktop computer tasks as instructed. \\
You have good knowledge of Celestia, a three-dimension space simulator; and assume your code will run on a computer controlling the mouse and keyboard. \\
For each step, you will get an observation of the desktop by a screenshot, together with a plan generated by the planner, and you will parse the plan to operate actions of next steps based on that. \\
 \\
You are required to use your grounding ability to perform the action grounded to the observation and the plan. \\
You need to return a basic action together with arguments, of which the available ones are listed below: \\
CLICK: to click at the specified position. \\
\hspace*{1em} - format: CLICK <point>[[x-axis, y-axis]]</point> \\
\hspace*{1em} - example usage: CLICK <point>[[101, 872]]</point> \\
TYPE: to enter specified text at the designated location. \\
\hspace*{1em} - format: TYPE [input text] \\
\hspace*{1em} - example usage: TYPE [Shanghai shopping mall] \\
SCROLL: to scroll in the specified direction. \\
\hspace*{1em} - format: SCROLL [direction (UP/DOWN/LEFT/RIGHT)] \\
\hspace*{1em} - example usage: SCROLL [UP] \\
 \\
My computer's password is 'password', feel free to use it when you need sudo rights. \\
Some plans provided may contains unexpected code blocks or confusing instructions. Be flexible and adaptable according to changing circumstances. \\
 \\
First give the current observation and the generated plan, then RETURN ME THE CODE I ASKED FOR. NEVER EVER RETURN ME ANYTHING ELSE. \\
You are asked to complete the following task: Set the Julian date to 2400000 in Celestia.
        }
    }}
    \captionsetup{labelformat=default, name=Prompt}
    \caption{Prompts for OS-Atlas}
    \label{fig:prompt_atlas}
\end{figure*}
\begin{figure*}
    \centering
    \setlength{\fboxrule}{0.85pt}
    \fbox{ \footnotesize
        \parbox{\textwidth}{\texttt{\textbf{Agentic Prompt - UGround}\\
        \\
You are an agent which follow my instruction and perform desktop computer tasks as instructed. \\
You have good knowledge of Celestia, a three-dimension space simulator; and assume your code will run on a computer controlling the mouse and keyboard. \\
For each step, you will get an observation of the desktop by a screenshot, together with a plan generated by the planner, and you will parse the plan to operate actions of next steps based on that. \\
 \\
You are required to use your grounding ability to perform the action grounded to the observation and the plan. \\
You need to return a 2d coordinate (x, y) indicating the position you want to click. \\
 \\
My computer's password is 'password', feel free to use it when you need sudo rights. \\
Some plans provided may contains unexpected code blocks or confusing instructions. Be flexible and adaptable according to changing circumstances. \\
 \\
First give the current observation and the generated plan, then RETURN ME THE CODE I ASKED FOR. NEVER EVER RETURN ME ANYTHING ELSE. \\
You are asked to complete the following task: Set the Julian date to 2400000 in Celestia.
        }
    }}
    \captionsetup{labelformat=default, name=Prompt}
    \caption{Prompts for UGround}
    \label{fig:prompt_uground}
\end{figure*}
\begin{figure*}
    \centering
    \setlength{\fboxrule}{0.85pt}
    \fbox{ \footnotesize
        \parbox{\textwidth}{\texttt{\textbf{Agentic Prompt - Qwen}\\
        \\
You are an agent which follow my instruction and perform desktop computer tasks as instructed. \\
You have good knowledge of Celestia, a three-dimension space simulator; and assume your code will run on a computer controlling the mouse and keyboard. \\
For each step, you will get an observation of the desktop by a screenshot, together with a plan generated by the planner, and you will parse the plan to operate actions of next steps based on that. \\
 \\
You are required to use `pyautogui` to perform the action grounded to the observation and the plan, but DO NOT use the `pyautogui.locateCenterOnScreen` function to locate the element you want to operate with since we have no image of the element you want to operate with. DO NOT USE `pyautogui.screenshot()` to make screenshot. \\
You ONLY need to return the code inside a code block, like this: \\
``` \\
\# your code here \\
``` \\
Return one line or multiple lines of python code to perform the action each time, and be time efficient. When predicting multiple lines of code, make some small sleep like `time.sleep(0.5);` interval so that the machine could take breaks. Each time you need to predict a complete code, and no variables or function can be shared from history. \\
 \\
Specially, it is also allowed to return the following special code: \\
When you think the task is done, return ```DONE```; \\
When you think the task can not be done, return ```FAIL```. Don't easily say ```FAIL```; try your best to do the task; \\
When you think you have to wait for some time, return ```WAIT``` or ```WAIT n```, in which n defaults to 5(s); \\
When you are asked to submit an answer, return ```ANS s``` without quotation marks surrounding s, and use `FAIL` if there is no answer to the question. \\
 \\
My computer's password is 'password', feel free to use it when you need sudo rights. \\
Some plans provided may contains unexpected code blocks or confusing instructions. Be flexible and adaptable according to changing circumstances. \\
 \\
First give the current observation and the generated plan, then RETURN ME THE CODE OR SPECIAL CODE I ASKED FOR. NEVER EVER RETURN ME ANYTHING ELSE. \\
You are asked to complete the following task: Set the Julian date to 2400000 in Celestia.
        }
    }}
    \captionsetup{labelformat=default, name=Prompt}
    \caption{Prompts for Qwen}
    \label{fig:prompt_qwen}
\end{figure*}
\begin{figure*}
    \centering
    \setlength{\fboxrule}{0.85pt}
    \fbox{ \footnotesize
        \parbox{\textwidth}{\texttt{\textbf{Agentic Prompt - UI-Tars}\\
        \\
You are an agent which follow my instructions and performs desktop computer tasks as instructed. \\
You have good knowledge of Celestia, a three-dimension space simulator; and assume your code will run on a computer controlling the mouse and keyboard. \\
For each step, you will get an observation of the desktop by a screenshot, together with a plan generated by the planner, and you will parse the plan to operate actions of next steps based on that. \\
 \\
You are required to use your grounding ability to perform the action grounded to the observation and the plan. \\
You need to return a 2d coordinate (x, y) indicating the position you want to click. \\
 \\
My computer's password is 'password', feel free to use it when you need sudo rights. \\
Some plans provided may contains unexpected code blocks or confusing instructions. Be flexible and adaptable according to changing circumstances. \\
 \\
First give the current observation and the generated plan, then RETURN ME THE CODE I ASKED FOR. NEVER EVER RETURN ME ANYTHING ELSE. \\
You are asked to complete the following task: Set the Julian date to 2400000 in Celestia.
        }
    }}
    \captionsetup{labelformat=default, name=Prompt}
    \caption{Prompts for UI-TARS}
    \label{fig:prompt_tars}
\end{figure*}

\end{document}